\documentclass[preprint,12pt]{elsarticle}

\usepackage{amssymb}

\usepackage{amsmath}
\usepackage{amsthm}

\newtheorem{assumption}{Assumption}
\newtheorem{theorem}{Theorem}
\newtheorem{lemma}{Lemma}

\usepackage{times}
\usepackage[normalem]{ulem}
\usepackage{xcolor}
\usepackage{url}
\usepackage[hidelinks]{hyperref}
\usepackage[utf8]{inputenc}
\usepackage{marvosym}
\usepackage[small]{caption}
\usepackage{graphicx}
\usepackage{booktabs}
\usepackage{algorithm}
\usepackage[switch]{lineno}

\usepackage{balance} 
\usepackage[T1]{fontenc}    
\usepackage{amsfonts}       
\usepackage{nicefrac}       
\usepackage{microtype}      
\usepackage{xcolor}         

\usepackage{algorithm}
\usepackage{algorithmic}
\usepackage{verbatim}
\usepackage{multirow}
\usepackage{capt-of}
\usepackage{subcaption}
\usepackage{caption}

\usepackage{fancyhdr}
\usepackage{diagbox}
\usepackage{makecell}
\usepackage{rotating}
\usepackage{enumitem}
\usepackage{lineno}

\usepackage[table]{xcolor}

\usepackage[markup=default,authormarkup=none,commandnameprefix=always]{changes}
\definechangesauthor[name={Chenran}, color=red]{cr}

\newcommand{\shortdownarrow}{\scalebox{0.65}{$\downarrow$}}

\newtheorem{definition}{Definition}


\journal{Neural Networks}

\begin{document}

\begin{frontmatter}

\title{DECHRL: Empowerment-Driven Delay-Aware Causal Hierarchical Reinforcement Learning}




\author[1]{Chenran Zhao\fnref{equal}}
\author[1,2]{Dianxi Shi\corref{cor1}\fnref{equal}}
\author[1]{Haotian Wang}
\author[4]{Mengzhu Wang}
\author[3]{Yaowen Zhang}
\author[2]{Chunping Qiu}
\author[1]{Shaowu Yang}

\fntext[equal]{These authors contributed equally to this work.}
\cortext[cor1]{Corresponding author.}

\affiliation[1]{organization={College of Computer Science and Technology, National University of Defense Technology},
	city={Changsha},
	country={China}}

\affiliation[2]{organization={Intelligent Game and Decision Lab (IGDL)},
	city={Beijing},
	country={China}}

\affiliation[3]{organization={Institute of Military Transportation},
	city={Tianjin},
	country={China}}

\affiliation[4]{organization={School of Artificial Intelligence, Hebei University of Technology},
	city={Tianjin},
	country={China}}

\begin{abstract}
Many real-world tasks involve delayed effects, where the outcomes of actions emerge after varying time lags. Existing delay-aware reinforcement learning methods often rely on state augmentation, prior knowledge of delay distributions, or access to non-delayed data—limiting their generalization. Hierarchical reinforcement learning, by contrast, inherently offers advantages in handling delays due to its hierarchical structure, yet existing methods are restricted to fixed delays. To address these limitations, we propose Delay-Empowered Causal Hierarchical Reinforcement Learning (DECHRL). DECHRL explicitly models both the causal structure of state transitions and their associated stochastic delay distributions. These are then incorporated into a delay-aware empowerment objective that drives proactive exploration toward highly controllable states, thereby improving performance under temporal uncertainty. We evaluate DECHRL in modified 2D-Minecraft and MiniGrid environments featuring stochastic delays. Experimental results show that DECHRL effectively models temporal delays and significantly outperforms baselines in decision-making under temporal uncertainty.

\end{abstract}

\begin{keyword}
Delay, Empowerment, Causal Hierarchical Reinforcement Learning

\end{keyword}

\end{frontmatter}





\section{Introduction}

In real-world decision-making scenarios, the consequences of a decision often do not manifest immediately but emerge after a period of time in several subsequent steps, i.e., a phenomenon known as \emph{delay effects}. Without considering such delay effects, traditional RL methods typically assume the \emph{immediate} transitions between the state-action pair~\cite{sutton1999reinforcement}. As a result, such an oversimplification may result in incorrect credit assignment in reinforcement learning, thus leading to delay-specific bias and hindering its applicability to real-world, delay-aware settings.

Recent advances in constructing the Delay-Aware Reinforcement Learning (DARL) framework have yielded remarkable progress, broadly falling into three categories. First, \textit{memory-based} methods mitigate the impact of delayed state transitions by augmenting the current state with historical trajectory information~\cite{nath2021revisiting,bouteiller2020reinforcement,DBLP:journals/corr/abs-2106-11854,DBLP:journals/corr/abs-2403-12309,DBLP:journals/prl/AgarwalA21}, but the complexity of the augmented state space grows substantially as the maximum delay in the environment increases. Second, \textit{prior-knowledge-based} approaches leverage known constant delays or delay distributions to identify the effective actions that actually trigger state transitions~\cite{yudelay,schuitema2010control}, yet rely on domain-specific priors that hinder generalization. Third, \textit{oracle-assisted} methods rely on offline-collected non-delayed data, typically in the form of non-delayed expert demonstrations, to provide training guidance~\cite{wang2024addressing,liotet2022delayed}. However, their reliance on environment-specific oracles hinders generalization. Currently, there is no method that directly models the stochastic nature of environmental delays to truly equip the agent with the capacity for exploration in unknown environments.

Hierarchical Reinforcement Learning (HRL) offers inherent robustness to delay effects. \textit{Goal-based HRL}~\cite{levy2017learning} abstracts control into high-level sub-goal selection and low-level execution, thereby mitigating the impact of low-level delays by reducing the decision frequency of the high-level policy, while \textit{option-based HRL}~\cite{bacon2017option,kim2023lesson} handles stochastic delays through learned option termination functions. However, these methods lack explicit modeling of fine-grained dynamics, limiting their ability to reason about delays in an interpretable way. Recent \textit{Causal HRL (CHRL)} approaches~\cite{ZHAO2026108275,hu2022causality} offer a more principled framework by uncovering latent causal hierarchies via structural causal models~\cite{peters2017elements}, enabling robust planning under delayed signals. However, they are constrained to fixed delays, making them less effective in real-world scenarios with stochastic delays.

To tackle this challenges, 
we propose a novel approach named {\bf D}elay-{\bf E}mpowered {\bf C}ausal {\bf H}ierarchical {\bf R}einforcement {\bf L}earning (\textbf{DECHRL}), served as the pioneer framework in CHRL that is aware of stochastic delays. 
To be specific, our DECHRL consists of two key components: (1) \textbf{A Causal Delay Distribution Modeling Module}, which jointly infers the underlying causal structure of the environment and the stochastic delay distribution directly from delayed observational data. This enables the agent to perceive and internalize the unknown delay characteristics of the environment, effectively endowing it with the ability to reason about when and how actions influence future states despite observation or action delays; (2) \textbf{A Delay-Aware Empowerment Objective}, which explicitly maximizes the agent’s causal influence over future states under delayed feedback. By quantifying the agent’s capacity to shape diverse and controllable outcomes in a delay-aware manner, this objective encourages proactive and effective exploration, thereby ensuring robust hierarchical decision-making even when interactions with the environment are temporally misaligned.
Overall, DECHRL provides a principled approach to reinforcement learning under unknown, stochastic delays. It enables agents to learn effective and robust policies in environments where action effects are temporally misaligned and delay characteristics are unknown a priori. By jointly equipping agents with the ability to infer delay structures and to explore purposefully under uncertainty, DECHRL achieves adaptive policy learning in realistic delayed settings.
To further enhance practical applicability, we also propose a \textbf{Simplified-DECHRL} variant that reduces computational overhead by adopting a coarse-grained delay representation. This version significantly improves scalability while incurring only minimal performance degradation.

Overall, our contributions are summarized as follows:
\begin{itemize}
    \item To the best of our knowledge, this is the first attempt to integrate stochastic delay modeling into a CHRL framework, showcasing how causal hierarchies inherently capture delayed temporal dependencies. This enables the agent to reason about when actions take effect under uncertainty and to align high-level decisions with temporally misaligned low-level outcomes, capabilities that are essential for robust policy learning in real-world delayed environments.
    \item 
    We propose DECHRL, a novel Causal Hierarchical Reinforcement Learning (CHRL) framework that explicitly integrates stochastic delay distribution modeling with a delay-aware empowerment objective. This dual design enables effective and robust policy learning in environments with temporally misaligned action effects and a priori unknown stochastic delays.

    \item We empirically evaluate DECHRL on two structured, long-horizon benchmark environments: modified 2D-Minecraft~\cite{sohn2018hierarchical} and MiniGrid~\cite{MinigridMiniworld23}. Both environments feature semantically rich, structured state representations (e.g., object-centric or grid-based symbolic observations) and involve \textbf{discrete stochastic action delays}—the regime for which our method is designed. Under these conditions, DECHRL maintains solid performance with maximum delays $\tau_{\max} \in \{4, 8\}$ and across stochastic delay levels $\sigma_{delay} \in \{0.4, 0.6, 0.8, 1.0\}$. Notably, the simplified variant, Simplified-DECHRL, demonstrates improved scalability: with a delay-modeling granularity of $\kappa=4$, it continues to perform well on the 2D-Minecraft task even when the maximum delay is increased to $\tau_{\max} = 30$. These results confirm that DECHRL is particularly well-suited to hierarchical decision-making problems where (i) delays are discrete and stochastic, and (ii) the environment provides structured, interpretable states that enable causal discovery and meaningful temporal abstraction. In such settings, our framework reliably recovers delay patterns from delayed interaction data and aligns outcomes with their causal antecedents. The resulting delay-aware empowerment objective then enables robust credit assignment under temporal misalignment, facilitating effective policy learning in long-horizon tasks with uncertain timing.
\end{itemize}



\section{Related Work}

\subsection{Delay-Aware Reinforcement Learning}\label{sec:related_work_delayed_rl}
Significant progress has been made in Delay-Aware Reinforcement Learning (DARL), with existing methods typically grouped into three categories:
\textbf{(1) Memory-based methods.} These approaches enhance the current state with historical trajectory information~\cite{nath2021revisiting,bouteiller2020reinforcement,DBLP:journals/corr/abs-2106-11854,DBLP:journals/prl/AgarwalA21} to mitigate the effects of delay. This can be done by either concatenating past $K$ steps of observations and actions, or using a model to infer the current latent state~\cite{liotet2021learning,DBLP:journals/corr/abs-2403-12309}. However, these methods face scalability challenges due to the increasing state complexity caused by longer delays.
\textbf{(2) Prior knowledge-based methods.} These methods leverage known delay information to identify the action responsible for a given outcome without the need to store full interaction histories. Examples include delay-aware Q-learning~\cite{yudelay} and modified TD updates~\cite{schuitema2010control}. While efficient, these methods rely heavily on accurate prior knowledge of delays.
\textbf{(3) Oracle-assisted methods.} These leverage expert demonstrations or non-delayed data to train or initialize policies. Methods like DRC-SAA~\cite{wang2024addressing} and DIDA~\cite{liotet2022delayed} show improved performance, but require access to non-delayed data which may not be available in real-world settings.

Memory-based methods augment the current state with past trajectories, prior knowledge-based methods assume known delays, while oracle-assisted approaches rely on non-delayed data. Our method, in contrast, learns from scratch under unknown stochastic delays, without prior knowledge or non-delayed data, enabling more practical and generalizable learning.

\subsection{Causal Hierarchical Reinforcement Learning}\label{sec:chrl}
Hierarchical Reinforcement Learning (HRL)~\cite{DBLP:journals/nn/LiWT25} decomposes complex tasks into manageable sub-tasks, with higher-level policies selecting subgoals and lower-level policies focusing on their execution. While HRL enhances efficiency through skill abstraction, it remains susceptible to delays in state transitions.

Causal HRL (CHRL) extends this framework by learning hierarchical policies that capture and utilize causal dependencies between actions and states, with potential to handle multi-step causal dependencies.
Most Causal Hierarchical Reinforcement Learning (CHRL) methods adopt a common two-stage framework: (i) \textit{causal discovery}, which identifies causal relationships or chains from interaction data, and (ii) \textit{hierarchical policy training}, which maps the discovered causal structures to a set of skills or sub-goals, assigns a dedicated policy to each, and trains these policies to achieve their respective objectives. These steps are repeated until the full causal structure is uncovered. While sharing this core idea, methods vary in how they implement each stage. We classify them into four categories:
\textbf{(1) One-step causal chain.} CEHRL~\cite{corcoll2022disentangling} and CDHRL~\cite{hu2022causality} learn one-step causal graphs using interventional or counterfactual data, but ignore multi-step delays.
\textbf{(2) Skill libraries from static graphs.} SCALE~\cite{lee2023scale}, COInS~\cite{chuckgranger}, HCPI-HRL~\cite{chen2025hcpi} convert static causal graphs into skill libraries, but rely on external supervision, human priors, or controlled environments.
\textbf{(3) Attention-based key-step graph.} VACERL~\cite{nguyen2024variable} uses attention to identify key observation–action pairs that predict returns, forming a graph over critical steps, though assuming fixed delays.
\textbf{(4) Delay-aware graph.} D3HRL~\cite{ZHAO2026108275} models multi-step causal effects using multi-timescale causal discovery and conditional independence tests, then builds a hierarchical controller on the resulting delay-aware graph. It performs well under fixed delays but assumes they are deterministic.


Previous CHRL methods assume fixed delays, limiting their applicability to real-world stochastic latencies. Our approach models delay distributions for each causal relationship and integrates them into an empowerment-based objective, enabling more effective learning under delay uncertainty.

\subsection{Empowerment-Driven Exploration}
\label{sec:empowerment_summary}

Empowerment is a task-agnostic intrinsic reward that measures an agent’s influence over its future, guiding it toward states with greater control—without external rewards.
Empowerment-based methods can be grouped into the following categories:
\textbf{(1) Controllability-based.} VIMIM~\cite{mohamed2015variational} and VIC~\cite{gregor2016variational} maximize empowerment directly as an intrinsic reward to drive exploration.
\textbf{(2) Diversity-driven.} DIYAN~\cite{eysenbach2018diversity} and DADS~\cite{sharma2019dynamics} discover diverse and predictable skills using mutual information between latent skills and future states.
\textbf{(3) Latent planning and representation.} IPE~\cite{bharadhwaj2022information} and VGCRL~\cite{choi2021variational} shape latent spaces to focus on controllable features.
\textbf{(4) Causal-aware.} ECL~\cite{cao2025towards} incorporates causal modeling into empowerment to improve interpretability.

Most empowerment-based methods rely on explicit modeling of environment dynamics. In contrast, our approach leverages the causal dependencies discovered during the causal discovery phase as an abstract transition model, imposing structural constraints and offering semantic guidance for hierarchical policy design, thereby reducing the need for explicit dynamics modeling.

\section{Preliminaries}


Real-world dynamics often involve state transitions that vary in duration and may temporally overlap, yet remain non-interfering. Crucially, these transitions are not arbitrary; they conform to underlying causal principles. This motivates modeling them as temporally extended causal links spanning multiple timesteps, a formulation that is both principled and expressive. In this section, we first discuss how causal relationships can be represented and learned using neural networks (Section~\ref{sec:scm}), and then we propose a new formalization of the model, tailored to the specific task we aim to solve, based on previous work in this area (Section~\ref{sec:cfsmdp}).



\subsection{Structural Causal Model}\label{sec:scm}

\textbf{Structural Causal Model (SCM)}~\cite{peters2017elements} is a foundational framework for modeling and reasoning about causal relationships among variables. It comprises two components: (1) a causal graph, typically a directed acyclic graph (DAG), that encodes causal edges; if \(C\) causes \(E\), the graph includes a directed edge \(C \to E\), making \(C\) a parent of \(E\), denoted \(\mathcal{PA}(E)\); and (2) a set of generating functions that specify how each variable is produced from its parents together with an exogenous noise term, e.g., \(E = f_E(\mathcal{PA}(E), U_E)\).

\textbf{The implementation of an SCM}~\cite{ke2019learning} with $M$ effect variables $\{S^i\}_{i=1}^{M}$ and $N$ cause variables $\{S^j\}_{j=1}^N$ comprises two components:

\begin{itemize}
    \item \textbf{Causal Graph Adjacency Matrix $\boldsymbol{\eta} \in \mathbb{R}^{M \times M}$:} This matrix contains learnable parameters that govern the edge probabilities in the causal graph. Specifically, for each pair of variables $S^i$ and $S^j$, the value $\sigma(\eta_{ij})$, where $\sigma(\cdot)$ denotes the sigmoid function, represents the probability that $S^i$ is a direct cause of $S^j$.

    The gradient with respect to the edge probability parameters $\boldsymbol{\eta}$ is typically estimated using a REINFORCE-style estimator~\cite{DBLP:conf/iclr/BengioDRKLBGP20}. Specifically, $N_s$ causal graphs $\{G^n\}_{n=1}^{N_s}$ are independently sampled as $G^n \sim \text{Bernoulli}(\sigma(\boldsymbol{\eta}))$, and their fit to the data is evaluated via the negative log-likelihood $\mathcal{L}$ over $K$ batches of interventional data $\{D^k\}_{k=1}^{K}$. The resulting gradient estimator takes the form:
    \begin{equation}
    \begin{aligned}
        &\nabla_{\boldsymbol{\eta}} \mathbb{E}_{G \sim p(G \mid \boldsymbol{\eta})} \left[ \sum_{k=1}^{K} \mathcal{L}(G, D^k) \right] \\
        &\approx \frac{1}{N_s} \sum_{n=1}^{N_s} \left[ \sum_{k=1}^{K} \mathcal{L}(G^n, D^k) \cdot \nabla_{\boldsymbol{\eta}} \log p(G^n \mid \boldsymbol{\eta}) \right],
    \end{aligned}
    \end{equation}
    where $\log p(G^n \mid \boldsymbol{\eta})$ denotes the log-probability of the sampled binary adjacency matrix $G^n$ under the Bernoulli distribution parameterized by $\sigma(\boldsymbol{\eta})$.
    
    \item \textbf{A Set of Generating Functions $\mathbf{F} = \{f^i(\theta)\}_{i=1}^M$}: Each predicts the logits of $S^i$ based on its parent variables in the sampled causal graph $G$:
    \begin{equation}
    \widehat{S}^i = f^i_\theta(S^i \mid X^{\mathcal{PA}(S^i;G)}),
    \end{equation}
    where $\mathcal{PA}(S^i; G)$ denotes the set of parent variables of $S^i$ under the causal graph structure $G$, and $\theta$ represents the network parameters. 
    
    Causal dependencies are typically modeled between successive time steps, under the assumption that the values of variables at time $t-1$ directly influence those at time $t$. D3HRL~\cite{ZHAO2026108275} generalizes this paradigm by adopting a multi-timescale causal discovery approach, learning dependencies from the extended temporal window $\{t-\tau_{\max},\, t-\tau_{\max}+1,\, \dots,\, t-1\}$ to time $t$. The generating function is then trained by optimizing the log-likelihood:
    \begin{equation}
        \max_\theta\mathcal{L}(\theta)=\log\mathrm{~softmax}\left(f_\theta^i\left(S_t^i\mid \mathcal{PA}(S^i;G)\right)\right).
    \end{equation}
    To enable the generating function to better fit the effects under different values of a cause variable, we need to set the cause to different values and observe the corresponding effects. This operation of assigning different values to the cause variable is referred to as an \textbf{\textit{intervention}}.

\end{itemize}


\subsection{Stochastic-Delay Causal Factored-SMDPs}\label{sec:cfsmdp}
Motivated by SCM and Causal Factored-SMDPs~\cite{ZHAO2026108275}, we introduce a new formalization that captures causal structures governing state transitions with random delays within factored state and action spaces.

Before presenting the formal model, we first state the standard causal assumptions upon which our framework is built:
\begin{itemize}
    \item \textbf{Acyclicity}: The task hierarchy induces a forward progression of subgoals, as illustrated by the state transition dynamics in Figure~\ref{fig:task_state_transitions}.
    \item \textbf{Faithfulness}~\cite{DBLP:books/daglib/0023012}: State representations are sparse and functionally modular—each sub-state (e.g., \texttt{has\_wood}) is determined by a small, explicit set of causes. This renders accidental cancellations among multiple causal pathways implausible, satisfying the faithfulness assumption in practice.
    \item \textbf{Independent Sub-State Transitions}~\cite{DBLP:journals/ai/BoutilierDG00}: The global state decomposes into disjoint factors (e.g., agent position, inventory, object states), and each factor evolves independently given the action, consistent with factored MDP formulations.
    \item \textbf{Cause Precedes Effect}~\cite{pearl2000models}: Causal influences are strictly unidirectional in time: for all $i \neq j$, $X_t^j$ does not cause $X_t^i$ (no instantaneous causation), nor does it influence any past state $X_{t-\tau}^i$ for $\tau \geq 1$.
    \item \textbf{Causal Sufficiency}~\cite{DBLP:books/daglib/0023012}: All relevant causal variables are fully observed; there are no hidden confounders. Effects arise only when their explicitly modeled causes are jointly satisfied within a bounded temporal window. Complex dependencies such as path-dependent or cumulative triggers lie beyond the scope of this work.
    \item \textbf{Causal Markov Condition}~\cite{DBLP:books/daglib/0023012}: In the causal graph, each variable is independent of its non-descendants given its direct parents. This condition holds in our setting and aligns with the standard Markov property in reinforcement learning~\cite{sutton1999reinforcement}.
\end{itemize}

Our formal model is then defined as the tuple $\langle \tau_{\max}, \mathbf{C}, \mathbf{E}, \mathbf{P}, \mathbf{F}, \mathbf{N} \rangle$, where:

\begin{itemize}
    \item \textbf{$\tau_{\max}$}: The maximum possible delay in the environment.
    
    \item \textbf{\(\mathbf{C}\) (Cause Space)}: The factored state space $\mathbf{S} = \{S^1\times S^2 \times \dots \times S^M\}$ and the factored action space $\mathbf{A} = \{A^1\times A^2 \times \dots \times A^N\}$ serve as causes.
    \item \textbf{\(\mathbf{E}\) (Effect Space)}: The factored state space $\mathbf{S} = \{S^1\times S^2 \times \dots \times S^M\}$ serves as effects.
     
    \item \textbf{\(\mathbf{P} = \{P_\tau\}_{\tau=1}^{\tau_{\max}}\) (Causal Relationship Matrices)}: Each causal relationship matrix \( P_\tau \in \{0,1\}^{M \times (M+N)} \) is a binary matrix that indicates which causes influence which effects after a delay of \( \tau \) time steps. The entry \( P_\tau[i][j] \) equals $1$ if \( S^j \) is the cause of \( S^i \), and $0$ otherwise.

    \item \textbf{\(\mathbf{F} = \{F_\tau\}_{\tau=1}^{\tau_{\max}}\) (Generating Functions)}: A collection of generating functions that determine the value of each effect variable given its direct causes over delay \(\tau\).

    \item \textbf{$\mathbf{N}=\{\mathcal{N}(\mu_i, \sigma_i^2)\}_{i=1}^M$ (Delay Distributions)}: Each delay distribution \( \mathcal{N}(\mu_i, \sigma_i^2) \) defines the distribution of delays for causal relationships where \( S^i \) is the effect. In practical problem-solving, for simplicity, we discretize the delay into positive integers \( \tau \) that do not exceed the maximum delay \( \tau_{\max} \) of the environment. The distribution \( \mathcal{N}(\mu_i, \sigma_i^2) \) then specifies the probability of each discrete delay value \( \tau \). In our setup, the delays in all causal relationships corresponding to the same effect are consistent.

    
\end{itemize}


\section{Method}
\label{sec:method}

\begin{figure*}[htbp]
    \centering
    \includegraphics[width=1.0\linewidth]{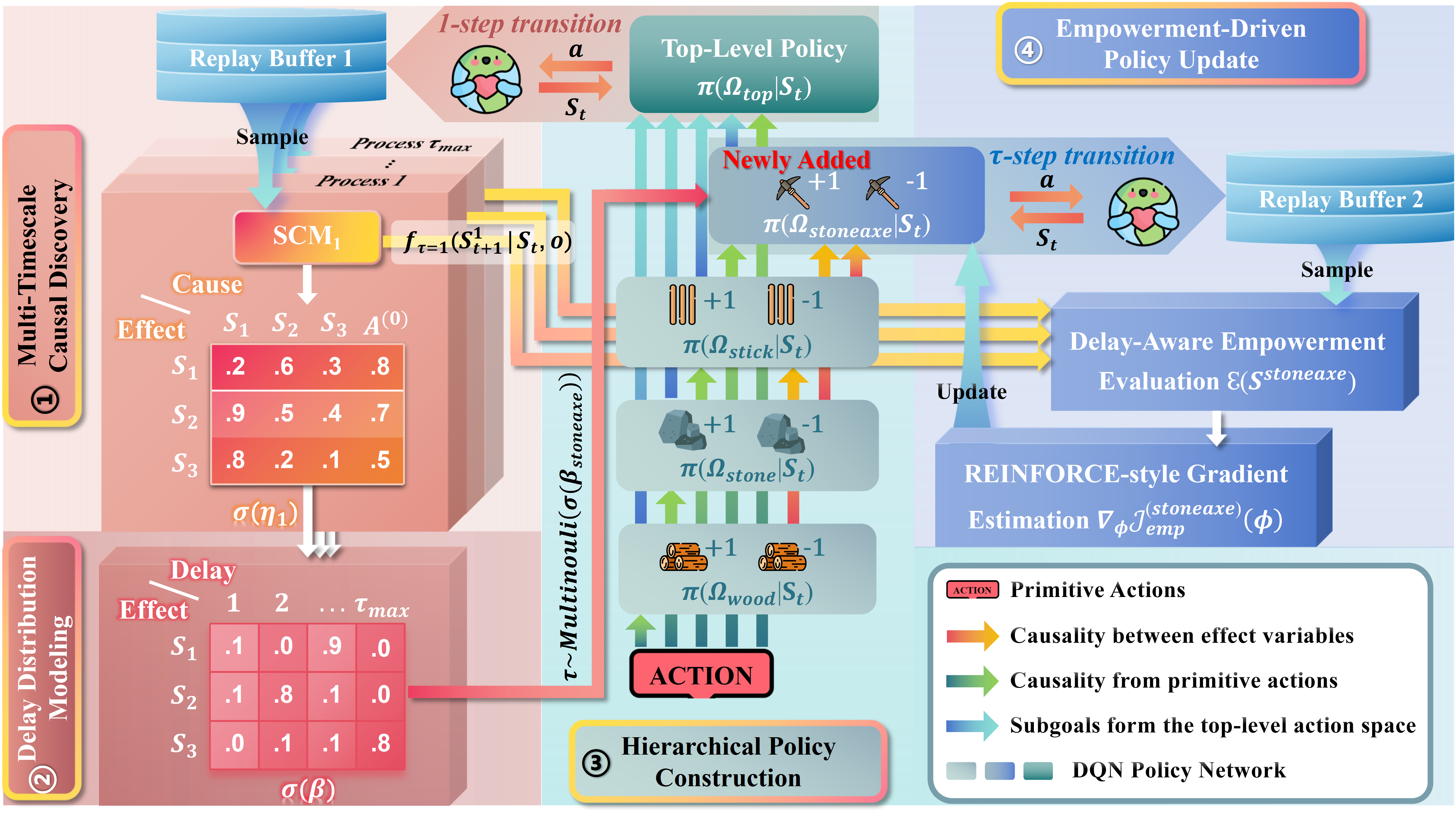}
    \caption{The overall framework of DECHRL.}\label{fig:overall}
\end{figure*}

This section introduces DECHRL, a CHRL framework that infers delay distributions and causal dependencies to enhance delay-aware exploration efficiency in hierarchical policy learning. As illustrated in Figure~\ref{fig:overall}, the overall algorithm proceeds in a progressive, iterative manner. In each round: (1) the agent employs the policies currently under its control to collect interventional data for \textit{multi-timescale causal discovery}, modeling causal dependencies across heterogeneous temporal horizons; (2) it uses the discovered structure to \textit{estimate the corresponding causal delay distributions}, enabling it to perceive and adapt to time-varying state-transition latencies; (3) based on the newly acquired causal knowledge, it \textit{instantiates a hierarchical policy architecture} that reflects the current causal understanding; and (4) it \textit{trains these policies with a delay-aware empowerment objective} to exploit the learned delay structure for decision making.
In the initial rounds, the only subgoals the agent can intervene on, i.e., the only ones under its control, are those in the primitive action space $\mathbf{A}$, which we denote as the set $List_{do}=\{\mathbf{A}\}$. Consequently, the only policy under its control is the bottom-level policy $\pi_{bottom}$, whose action space coincides with the primitive action space $\mathbf{A}$. As iterations proceed, the agent progressively uncovers deeper causal chains and builds higher-level policies, thereby gaining control over an increasingly rich set of subgoals $List_{do}$ and enabling exploration of increasingly deeper causal structures. This cycle continues until the learned causal chain includes subgoals that directly accomplish the target task.


\subsection{Learning Causal Delay Distributions from Delayed Interactions}\label{sec:delay-length}
Without modeling the distribution of delay effects, existing theories have informed the occurrence of bias during the state transition~\citep{wang2024addressing}. To fill this gap, we introduce a multi-timescale causal discovery approach to model the distribution of the delay variable, enabling the agent to capture temporal stochasticity.

\subsubsection{Multi-Timescale Causal Discovery}


Given a structured environment with $M$ effect variables and $N$ cause variables, we adopt the multi-timescale causal discovery framework~\cite{ZHAO2026108275} to model multi-timescale causal relationships among variables. In each iteration, the agent performs the following procedure sequentially:


\paragraph{Interventional Data Collection}
In non-simulatable and fully unknown environments, it is generally infeasible to perform ideal \textit{do}-interventions—such as arbitrarily forcing state variables to desired values—as is commonly done in simulation. Waiting for such interventions to occur naturally (i.e., for an intermediate variable to spontaneously attain a target value) would be prohibitively time-consuming. To accelerate this process, we leverage the current hierarchical policy to actively realize interventions with high success probability, thereby enabling efficient data collection for causal discovery. More specifically, by assigning a currently achievable subgoal $S^i\!\uparrow \in List_{do}$ ($S^i\!\uparrow$ denotes increasing the value of sub-state $S^i$) to the top-level policy $\pi_{top}$, the command is recursively propagated downward until the sub-policy $\pi_i$ responsible for accomplishing that subgoal is activated. Trained hierarchical policies can accomplish subgoals efficiently and with high success rates. Once a subgoal is achieved (interpreted as the intervention of a cause), we treat the post-intervention sub-state changes as candidate effects triggered by that cause. Aggregating such interventional data enables deeper exploring of the environment’s underlying causal structure, such as $S^i\!\uparrow \rightarrow S^j\!\uparrow$, where $S^j\!\uparrow$ denotes the sub-states that increases as a result of the intervention of the cause. \textit{It is important to note that the changes observed in the sub-states following the trigger of a cause are not necessarily, or not entirely, its causal effects; whether a true causal relationship exists must be determined through SCM modeling below.}

\paragraph{Multi-Timescale SCM Modeling} 
The agent uses the collected interventional data to estimate the underlying set of $\{\mathrm{SCM}{\tau}\}_{\tau=1}^{\tau_{\max}}$ across multiple timescales. To be specific, as shown in Figure~\ref{fig:overall}.A, We define a set of processes $\{ P_{\tau} \}_{\tau=1}^{\tau_{\max}}$, each dedicated to learning an $\mathrm{SCM}_\tau$ for a specific delay $\tau$, up to the maximum delay $\tau_{\max}$ (see Section~\ref{sec:scm} for details). For each effect variable $S^{i}$ ($i \in \{1, \dots, M\}$), process $P_{\tau}$ learns the $\tau$-lagged causal relationships $\{\mathcal{PA}(S^{i}) \xrightarrow{\tau} S^{i}\}_{i=1}^M$, where $\mathcal{PA}(S^{i})$ denotes the set of parent variables that directly influence $S^{i}$. Each process $P_{\tau}$ yields a $\mathrm{SCM}_{\tau}$ consisting of a set of generating functions \( F_\tau = \{ f^i_\tau(S^i \mid \mathcal{PA}(S^i)) \}_{i=1}^M \) and a causal graph matrix \( \boldsymbol{\eta}^\tau \in \mathbb{R}^{M \times N}\). The learned causal relationships are leveraged to model delay distributions (Section~\ref{sec:ddm}) and to construct the hierarchical policy framework (Section~\ref{sec:hierarchical_framework_construction}), while the learned generating function serves as a surrogate for the environment dynamics in evaluating the delay-aware empowerment objective (Section~\ref{sec:delay_aware_obj_eva}).



\subsubsection{Delay Distribution Modeling}\label{sec:ddm}

Building on the multi-timescale causal relationships learned above, we next model the corresponding delay distributions. To capture the stochastic temporal dynamics of causal effects, we first introduce a causal delay matrix $\boldsymbol{\beta} \in \mathbb{R}^{M \times \tau_{\max}}$, where $M$ denotes the number of effect variables in the environment and $\tau_{\max}$ represents the maximum delay horizon considered. Each element $\beta_{i,\tau}$ is the logit of the probability that the causal influence on effect variable $S^i$ manifests after a delay of $\tau$ steps. Applying the sigmoid transformation $\sigma(\cdot)$ yields the corresponding probability:
\begin{equation}
\sigma(\beta_{i,\tau}) = p(\mathcal{PA}(S^i) \xrightarrow{\tau} S^i) \in (0, 1), \quad \sum_{\tau} \sigma(\beta_{i,\tau}) = 1.
\end{equation}
Thus, each row $\sigma(\boldsymbol{\beta}_i)$ defines a soft categorical distribution over possible delays for the causal edge $\mathcal{PA}(S^i) \rightarrow S^i$, characterizing its delay-specific strength. 

With the modeling of data-generation process introduced above, we then detail the learning procedure of the delay distributions below:

\noindent{\bf Objective.} During training, we generate \(K\) delay hypotheses $\boldsymbol{H}^{(k)}$ by sampling from a Multinoulli distribution parameterized by softmax-normalized logits \(\sigma(\boldsymbol{\beta})\):
\begin{equation}
    \boldsymbol{H}^{(k)} \in \{1,...,\tau_{\max}\}^{M\times 1}\sim \mathrm{Multinoulli}\left(\sigma(\boldsymbol{\beta})\right), \quad k \in \{1,2,...,K\}.
\end{equation}
Our objective is to maximize the expected log-likelihood of the sampled hypotheses $\{\boldsymbol{H}^{(k)}\}_{k=1}^K$ under the previous learned causal graph matrix $\{\boldsymbol{\eta}^{\tau}\}_{\tau=1}^{\tau_{\max}}$:
\begin{equation}
    \max_{\boldsymbol{\beta}}\ \mathbb{E}_{\boldsymbol{H}, \boldsymbol{\eta}} \Bigg[ \sum_{i} \log \sigma (\boldsymbol{\eta}_{i}^{h_i^{(k)}}) \Bigg], 
\end{equation}
where \(\mathcal{L}_i^{(k)} = \log \sigma(\boldsymbol{\eta}_{i}^{h_i^{(k)}})\) denotes the log-probability that the causal influence \(\mathcal{PA}(S^i)\!\to\! S^i\) occurs with delay \(h_i^{(k)}=\boldsymbol{H}^{(k)}[i]\).

\noindent{\bf Optimization.} To reduce the variance, we employ the REINFORCE-style gradient~\cite{DBLP:conf/iclr/BengioDRKLBGP20} to update the parameters of the causal delay matrix $\boldsymbol{\beta}$:
\begin{equation}
\begin{aligned}
\nabla_{\beta_{i}} \mathbb{E}_{\boldsymbol{H} \sim \mathrm{Multinoulli}(\sigma(\beta_{i}))} \left[ \mathcal{L}_i^{(k)} \right] 
\approx
\sum_k \underbrace{\left(\sigma (\beta_{i}) - \mathrm{OneHot}(h^{(k)}_{i})\right)}_{\text{policy gradient}}
\cdot
\underbrace{w_i^{(k)}}_{\text{likelihood weight}},
\end{aligned}
\end{equation}
where \( w_i^{(k)} \propto \exp(\mathcal{L}_i^{(k)}) \) denotes the softmax-normalized likelihood weight evaluated over \(K\) hypotheses:
\begin{equation}
w_i^{(k)} = \frac{\exp(\mathcal{L}_i^{(k)})}{\sum_{k'} \exp(\mathcal{L}_i^{(k')})}, \quad \sum_k w_i^{(k)} = 1.
\end{equation}
This gradient adjusts the delay distribution to assign higher probabilities to delays that are more consistent with the previous learned causal graph matrix $\{\boldsymbol{\eta}^{\tau}\}_{\tau=1}^{\tau_{\max}}$.
We also explored two alternative gradient estimators, PPO~\cite{DBLP:journals/corr/SchulmanWDRK17} and A2C~\cite{DBLP:journals/corr/MnihBMGLHSK16}, to update the parameter matrices of the delay distribution. We assessed their effect on the overall performance of DECHRL. The full experimental results and analysis can be found in Section~\ref{sec:gradient_exp}.

\noindent{\bf Regularization.} To prevent degenerate solutions, we introduce the following regularization term:
\begin{equation}
\mathcal{L}_{reg} =
-\lambda_{1} \sum_{i,\tau} \sigma(\beta_{i,\tau})(1 - \sigma(\beta_{i,\tau}))
+ \lambda_2 \sum_{i,\tau} \sigma(\beta_{i,\tau}),
\end{equation}
where the first term encourages high entropy to avoid premature convergence, and the second encourages sparsity to highlight dominant delay choices. 
Although these two objectives may appear contradictory, their combination enables a balanced trade-off: with appropriately chosen hyperparameters, the model learns delay distributions that are both \emph{diverse enough to explore multiple temporal scales} and \emph{sparse enough to identify the most relevant causal lags}. We perform sensitivity analyses of $\lambda_{1}$ and $\lambda_{2}$ to characterize the hyperparameter regimes in which the algorithm exhibits robust performance, as detailed in Section~\ref{sec:sensitivity}.

\noindent {\bf Algorithmic Procedure.} As shown in Algorithm~\ref{alg:1}, each row of $\sigma(\boldsymbol{\beta})$ forms a delay distribution for the corresponding causal edge until the convergence. We also provide a detailed analysis of the computational complexity of the delay distribution modeling in \ref{app:delay_complexity}.

\begin{algorithm}
\caption{Delay Distribution Modeling}\label{alg:1}
\begin{algorithmic}[1]
\FOR{each iteration}
    \FOR{$k = 1$ to $K$}
        \STATE Sample delay hypothesis \(\boldsymbol{H}\) 
        \STATE Compute log-likelihoods \(\mathcal{L}_i^{(k)} = \log \boldsymbol{\eta}_{i}^{h_i^{(k)}}\)
    \ENDFOR
    \STATE Compute weights \(w_i^{(k)} \propto \exp(\mathcal{L}_i^{(k)})\)
    \STATE Update \(\beta_{i,\tau} \gets \beta_{i,\tau} + \alpha\big(\nabla_{\beta_{i,\tau}} \mathbb{E}_{h_i^{(k)}}[\,\mathcal{L}_i\,] - \nabla_{\beta_{i,\tau}} \mathcal{L}_{\mathrm{reg}}\big)\)
\ENDFOR
\STATE \textbf{Return} learned delay distribution \(\sigma(\boldsymbol{\beta})\)
\end{algorithmic}
\end{algorithm}

\subsection{Delay-Aware Empowerment Objective for Hierarchical Policy Training}\label{sec:emp_derivation}

Knowledge acquired in the preceding stage cannot be used directly for decision-making. Accordingly, we instantiate a hierarchical policy on top of the learned causal structure and optimize it with a delay-aware empowerment objective constructed from the learned distribution of delay effects, thereby guiding exploration toward actions with long-horizon influence.

\subsubsection{Hierarchical Policy Construction}\label{sec:hierarchical_framework_construction}
In this stage, we first construct the hierarchical framework grounded in the causal relationships learned in the preceding phase. Specifically, we define $\mathcal{V}_{e}$ as the set of sub-state variables that appear as effect nodes in the learned multi-timescale SCMs. For each effect variable $S^i \in \mathcal{V}_{e}$, we treat its increase ($S^i\!\uparrow$) and decrease ($S^i\!\downarrow$) as subgoals, governed by the dedicated policy $\pi_i$. The sub-goals currently under training constitute a set defined as $\mathcal{G}_e=\{S^i\!\uparrow, S^i\!\downarrow \mid S^i \in \mathcal{V}_e\}$. We define a bottom-level policy $\pi_{bottom}$ that operates over primitive actions $\mathbf{A}$ to realize directly executable subgoals. For higher-level variables \(S^j\), the corresponding subgoals (\(S^j\!\uparrow\) and \(S^j\!\downarrow\)) must be achieved by recursively invoking policies across the hierarchy; accordingly, we expand the action space of the policies \(\pi_j\) responsible for realizing these subgoals to:
\begin{equation}
    \Omega_j = \mathbf{A} \cup \left\{ S^k\!\uparrow, S^k\!\downarrow \mid S^k \in \mathcal{PA}(S^j) \right\},
\end{equation}
enabling recursive delegation to parent sub-goals. Additionally, a top-level policy $\pi_{top}$ is introduced, with an action space spanning all trained sub-goals $List_{do}$ and ongoing sub-goals $\mathcal{G}_{e}$:
\begin{equation}
    \Omega_{top} = List_{do} \cup \mathcal{G}_{e}.
\end{equation}
The hierarchy is implemented using Hindsight Experience Replay (HER)~\cite{DBLP:conf/nips/AndrychowiczCRS17}, with each level of the policy sampling transition delays from the learned delay distribution $\sigma(\boldsymbol{\beta})$ during experience collection.

\subsubsection{Delay-Aware Policy Optimization}\label{sec:delay_aware_obj_eva}

To enable the agent to effectively leverage the learned delay distribution for training hierarchical policies, we introduce a delay-aware empowerment objective. This encourages the high-level policy to explicitly account for temporal uncertainty in action effects, selecting sub-goals that exhibit strong causal influence and predictability across plausible delay scenarios.

We next detail the training procedure for the hierarchical policies and begin by defining the notion of delay-aware empowerment as follows:

\begin{definition}[Delay-aware Empowerment]\label{def:emp}
Given the current full state \(\mathbf{S}_t\) and a effect variable \(S^i\), the \emph{delay-aware empowerment} \(\mathcal{E}(S_t^i)\) is defined as the expected conditional mutual information (CMI) between a sub-goal \(o \in \Omega_i\) and the delayed future effect variable \(S^i_{t+\tau}\), under a delay distribution matrix \(\beta_i\):
$$\begin{aligned}
\mathcal{E}(S_t^{i}) &= \mathbb{E}_{\tau \sim \mathrm{Multinoulli}(\sigma(\beta_{i}))} \left[ I( S^{i}_{t+\tau}; o \mid \mathbf{S}_t) \right] \\
&= \sum_{\tau=1}^{\tau_{\max}} \sigma(\beta_{i,\tau}) \cdot \left[ H(S^{i}_{t+\tau} \mid \mathbf{S}_t) - H(S^{i}_{t+\tau} \mid \mathbf{S}_t, o) \right],
\end{aligned}$$
where $I(\cdot\,;\,\cdot \mid \cdot)$ denotes conditional mutual information, \(H(\cdot \mid \cdot)\) denotes conditional entropy.
\end{definition}

\noindent {\bf Empowerment Estimation.} The learned SCM captures causal effects among variables, allowing its generating functions to serve as surrogates for environment dynamics. More specifically, given a policy $\pi_i(\cdot | \mathbf{S}_t)$ over sub-goals $ o \in \Omega_i$, the marginal future state distribution is computed as:
\begin{equation}
p(S^{i}_{t+\tau} \mid \mathbf{S}_t) = \sum_{o \in \Omega_i} \pi_i(o \mid \mathbf{S}_t) \cdot f^i_{\theta}(S^{i}_{t+\tau} \mid \mathbf{S}_t, o).
\end{equation}
Then the first entropy term is computed using the marginal distribution computed above:
\begin{equation}
H(S^{i}_{t+\tau} \mid \mathbf{S}_t) = - \mathbb{E}_{\substack{o \sim \Omega_i \\ S^{i}_{t+\tau} \sim f_{\theta}^i}} \left[ \log p(S^{i}_{t+\tau} \mid \mathbf{S}_t) \right], \label{eq:emp_1}
\end{equation}
whose maximization promotes diverse state transitions and broad exploration of the reachable state space. Meanwhile, the second entropy term is:
\begin{equation}
    H(S^{i}_{t+\tau} \mid \mathbf{S}_t, o) = -\mathbb{E}_{\substack{S^{i}_{t+\tau} \sim f_{\theta}^i}} \left[ \log f_{\theta}^i(S_{t+\tau}^{i} \mid \mathbf{S}_t, o) \right], \label{eq:emp_2}
\end{equation}
and minimizing it encourages the agent to select sub-goals that reduce uncertainty about delayed future outcomes, specifically those exhibiting strong and predictable causal influence despite temporal misalignment.

\noindent {\bf Policy Gradient.} To optimize the high-level policy, the estimated delay-aware empowerment serves an advantage signal to guide a REINFORCE-style gradient estimator~\cite{DBLP:conf/iclr/BengioDRKLBGP20}:
\begin{equation}
\nabla_{\phi}\mathcal{J}^{(i)}_{\text{emp}}(\phi) = \mathbb{E}_{o \sim \pi_i} \left[ \hat{\mathcal{E}}(S_t^{i}) \cdot \nabla_{\phi} \log \pi_i(o \mid \mathbf{S}_t) \right].
\end{equation}
where \(\hat{\mathcal{E}}(S_t^{i})\) denotes the estimated delay-aware empowerment at state \(\mathbf{S}_t\), computed via Monte Carlo estimation based on Eqs.~(\ref{eq:emp_1}) and (\ref{eq:emp_2}), with log-sum-exp smoothing and normalization to ensure training stability. Maximizing this objective encourages the policy to select sub-goals with higher empowerment, thereby enhancing controllability and structured exploration.

\noindent {\bf Algorithmic Procedure.} We leave the full training procedure in Algorithm~\ref{alg:2}. The computational complexity of delay-aware empowerment–driven hierarchical policy training is analyzed in detail in \ref{app:delay_aware_policy_training_complexity}.

\begin{algorithm}[H]
\caption{Delay-Aware Policy Optimization}
\begin{algorithmic}[1]\label{alg:2}
\FOR{each delay $\tau = 1$ to $\tau_{\max}$}
    \STATE Sample sub-goal $o \sim \pi_i(\cdot \mid \mathbf{S}_t)$
    \STATE Predict future effect state $S^i_{t+\tau} \sim f_\theta^i(\cdot \mid \mathbf{S}_t, o)$
    \STATE Estimate marginal and conditional entropies using Eqs.~(\ref{eq:emp_1}) and (\ref{eq:emp_2})
\ENDFOR
\STATE Estimate $\hat{\mathcal{E}}(S_t^{i})$ according to Definition~\ref{def:emp}
\STATE Update $\pi_i$ using REINFORCE with $\hat{\mathcal{E}}(S_t^{i})$ as advantage
\end{algorithmic}
\end{algorithm}

Hierarchical policies whose corresponding sub-goals achieve a success rate greater than 0.5 are considered successfully trained, and these sub-goals are added to the $List_{do}$ set for subsequent rounds of learning and training. We evaluate the robustness and sensitivity of the algorithm to the selection of this success-rate threshold in Section~\ref{sec:success_ratio_sentivity}.

\subsection{Simplified-DECHRL}
Towards better scalability in realistic applications, our empirical findings show that hierarchical policies are robust to moderate delay variations. This indicates that while modeling each \(\tau \in [1, \tau_{\max}]\) individually is effective (as in our full framework), practical deployments can benefit from approximations such as coarse-grained delay bins that reduce computational overhead. Moreover, the multi-process architecture in our design, where each process performs independent causal discovery to identify a specific delay $\tau$, incurs a computational cost that scales linearly with the maximum delay $\tau_{\max}$, limiting both scalability and deployment feasibility. Thus, we subsample delays by a stride \(\kappa\) and retain only those divisible by \(\kappa\), restricting the delay distribution to support only these values. The resulting variant is termed as \textbf{Simplified-DECHRL}, a variant designed to reduce resource usage while preserving performance. In Section~\ref{sec:exp_scalability}, we evaluate the performance of Simplified-DECHRL under different delay-modeling granularities $\kappa$ across various maximum environment delays.



We emphasize that scalability issues arise from memory-based methods, as their augmented state space grows with the delay horizon. Our DECHRL circumvents this issue at its source by modeling the delay distribution directly, eschewing state augmentation altogether. This not only eliminates the combinatorial explosion in state representation but also equips the agent with the ability to accurately infer the true delay distribution of state transitions in the environment. To further improve efficiency and scalability, we introduce Simplified-DECHRL, which substantially reduces the computational and representational costs of DECHRL while preserving its core functionality. 

\section{Experiments and Analysis}

\subsection{Environmental Setup}\label{sec:env_setup}

\begin{figure}[htbp]
    \centering
    \begin{subfigure}[b]{0.82\textwidth}
        \centering
        \includegraphics[width=\linewidth]{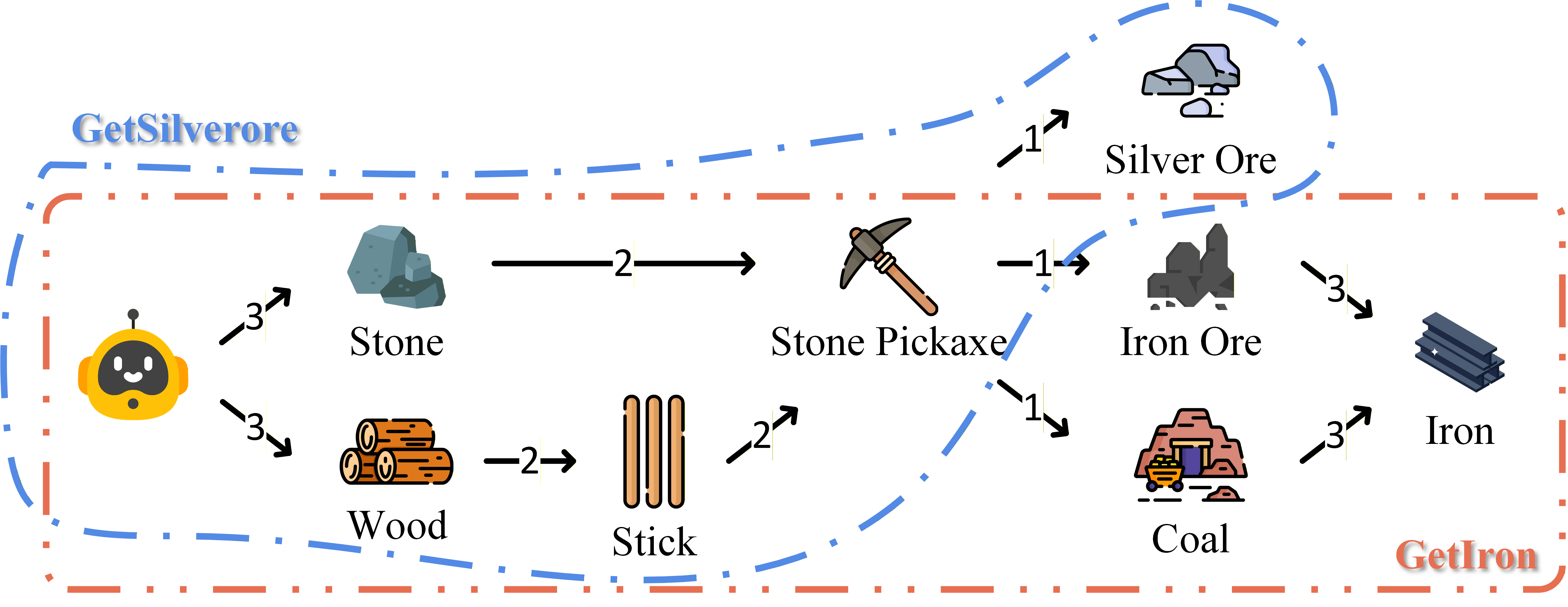}
        \caption{\textit{GetSilverore} and \textit{GetIron}}
        \label{fig:state_transition_silverore}
    \end{subfigure}

    \vspace{0.6em}
    
    \begin{subfigure}[b]{0.45\textwidth}
        \centering
        \includegraphics[width=\linewidth]{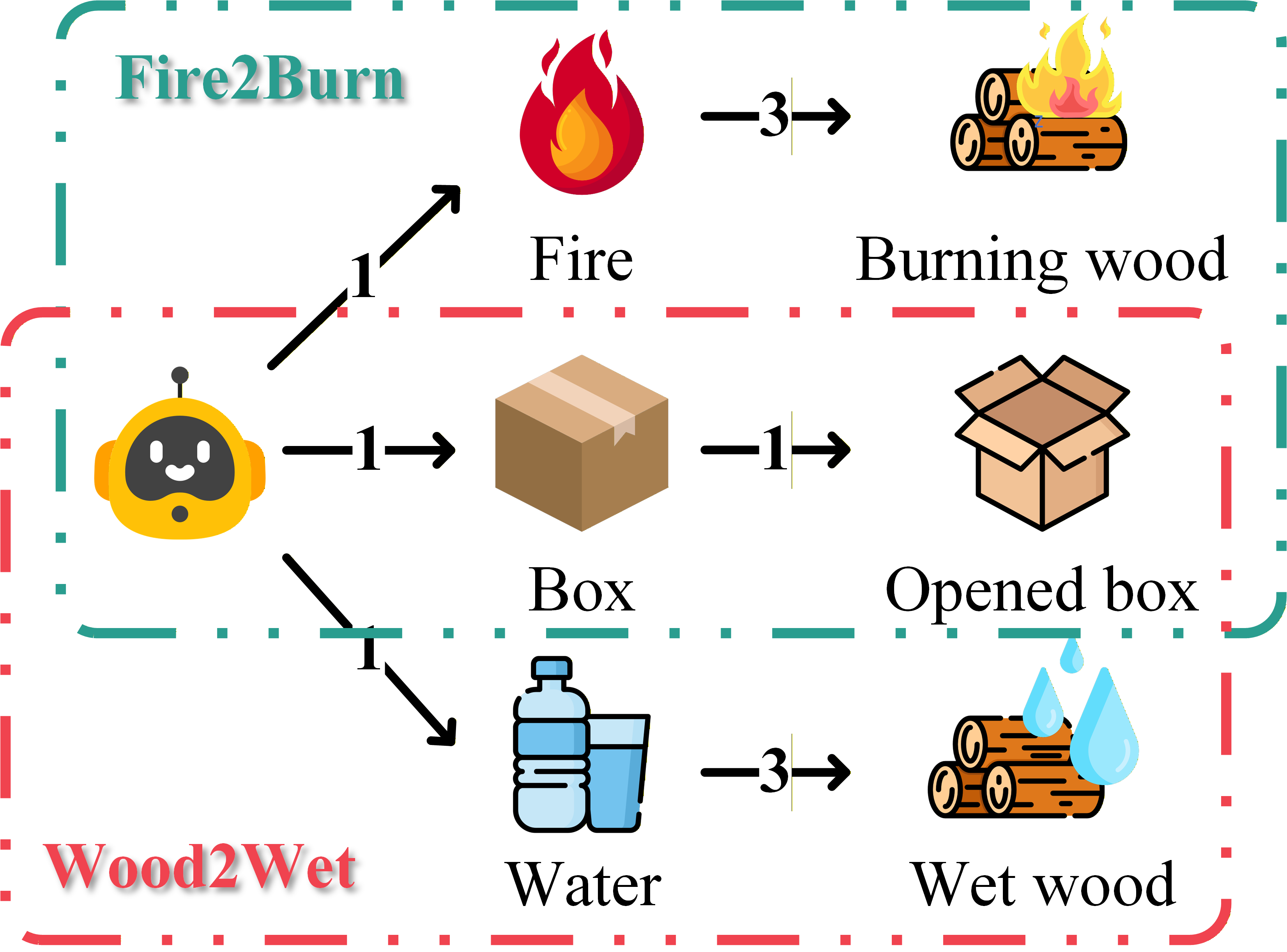}
        \caption{\textit{Fire2Burn} and \textit{Wood2Wet}}
        \label{fig:mini_state_trans}
    \end{subfigure}
    
    \caption{State-transition dynamics and their average delay characteristics involved in the MineCraft and MiniGrid tasks ($\tau_{\max}=4$).}
    \label{fig:task_state_transitions}
\end{figure}

\begin{figure*}[htbp]
    \centering
    \begin{tabular}{c}
        \subcaptionbox{$\sigma_{delay}=0.4$}{
            \includegraphics[width=1.0\textwidth]{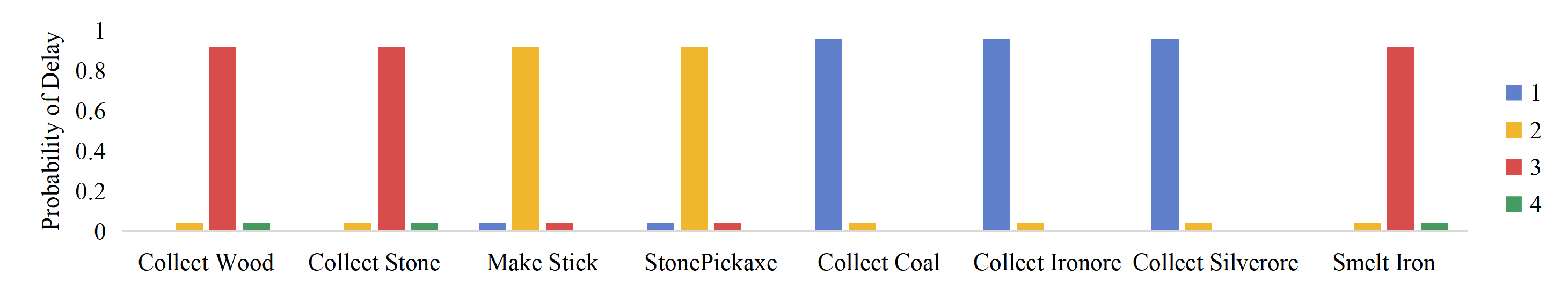}} \\
        \subcaptionbox{$\sigma_{delay}=0.6$}{
            \includegraphics[width=1.0\textwidth]{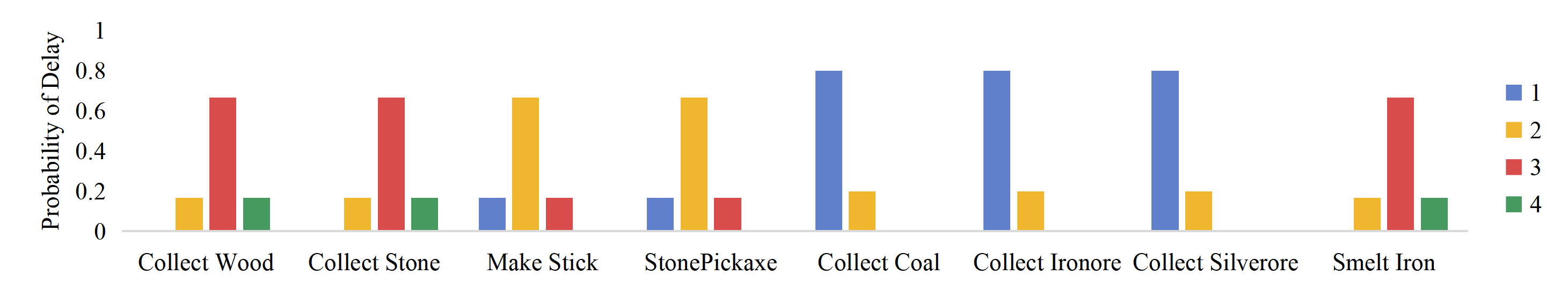}}\\
        \subcaptionbox{$\sigma_{delay}=0.8$}{
            \includegraphics[width=1.0\textwidth]{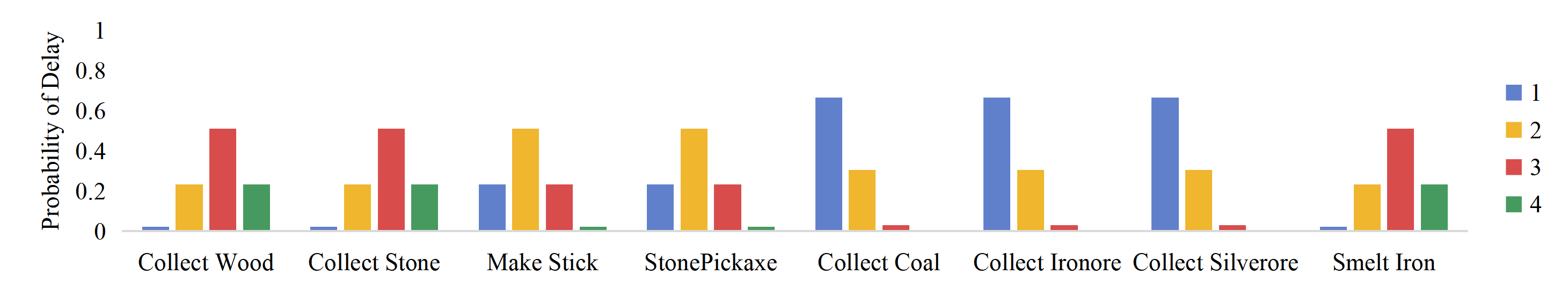}}\\
        \subcaptionbox{$\sigma_{delay}=1.0$}{
            \includegraphics[width=1.0\textwidth]{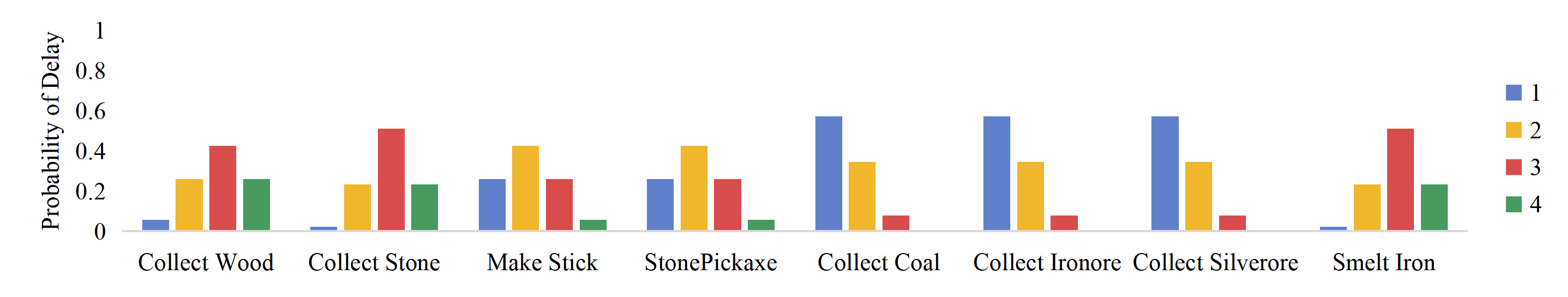}}
    \end{tabular}
    \caption{The delay distribution of transitions on the MineCraft tasks under different configuration of $\sigma_{delay}$.}\label{fig:minecraft_delay_dist} 
\end{figure*}

\begin{figure*}[htbp]
    \centering
    \includegraphics[width=0.95\textwidth]{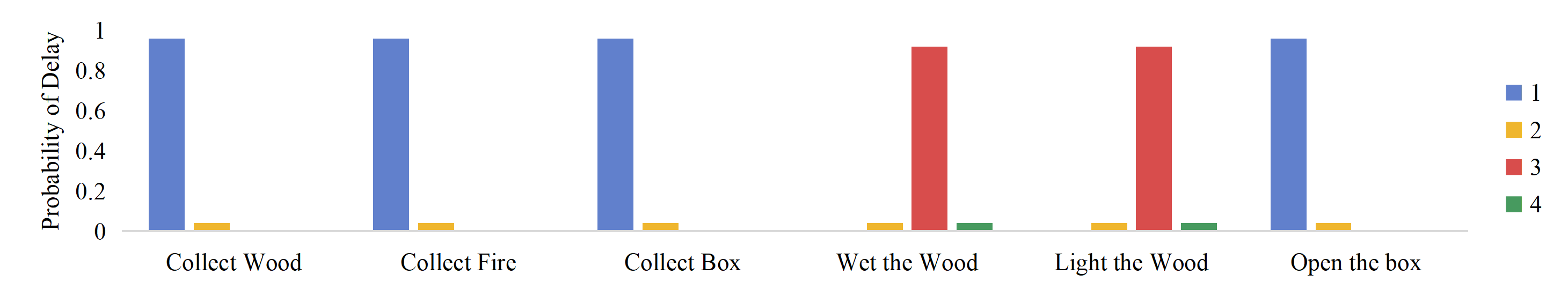}
    \caption{The delay distribution of transitions on the MiniGrid tasks under $\sigma_{delay}=0.4$.}\label{fig:minigrid_delay_dist} 
\end{figure*}

\textbf{Tasks.} We evaluate on 2D-Minecraft~\cite{sohn2018hierarchical} and MiniGrid~\cite{MinigridMiniworld23}, each with two designed tasks: \textit{GetSilverore} (T1) and \textit{GetIron} (T2) in 2D-Minecraft, and \textit{Fire2Burn} (T3) and \textit{Wood2Wet} (T4) in MiniGrid. 
Consider the \textit{GetIron} task as an illustrative example: within a finite time budget, the agent must complete a multi-stage procedure—first collecting wood and stone, then crafting sticks, using them to build a stone axe, subsequently mining iron ore and coal, and ultimately smelting these resources into iron. All tasks in our experimental setting follow this paradigm of structured, long-horizon sequential decision-making.
We introduce stochastic delays in state transitions, where each transition $\mathcal{PA}(S^i)\rightarrow S^i$ is governed by a distinct delay distribution following $\mathcal{N}(\mu_{i}, \sigma_{i}^2)$. The mean $\mu_{i}$ is a pre-defined integer not exceeding $\tau_{\max}$, while $\sigma_{i} \in \{0.4, 0.6, 0.8, 1.0\}$ controls the level of delay variability. In the following, we use \( \sigma_{delay} \) to represent the degree of stochasticity in state transitions. Figure~\ref{fig:task_state_transitions} presents the state transitions involved in each task along with their predefined average delays. Figures~\ref{fig:minecraft_delay_dist} and \ref{fig:minigrid_delay_dist} further illustrate, using the case of a maximum environment delay of $\tau_{\max} = 4$, the delay distributions of these state transitions under different levels of stochasticity $\sigma_{delay}$. The \textbf{agent \textit{has no access} to this prior information} and must \textbf{\textit{automatically discover}} the delay structure through interaction.

\textbf{Metrics.} We assess overall performance using two key metrics: \textbf{Average Success Ratio (ASR)} and \textbf{Average Sub-goal Distance to Completion (ADC)}, both computed over the last 100 training episodes. To evaluate how accurately the model captures temporal dynamics, we measure the \textbf{KL divergence} between the learned and ground-truth delay distributions. All reported results are averaged across five random seeds.

\textbf{Baselines.} We compare against representative DARL and HRL baselines under a unified hierarchical framework:
\textbf{(1) For DARL} baselines, we exclude oracle-assisted approaches that bypass delay modeling, and instead integrate \textbf{\textit{memory-based}}~\cite{DBLP:journals/prl/AgarwalA21} and \textbf{\textit{prior knowledge-based}}~\cite{yudelay,schuitema2010control} delay handling methods into the same HRL architecture as ours, ensuring fair comparison.
\textbf{(2) For HRL} baselines, we enhance non-causal HRL baselines (\textbf{\textit{HAC}}~\cite{levy2017learning}, \textbf{\textit{Option-Critic}}~\cite{bacon2017option}, \textbf{\textit{LESSON}}~\cite{kim2023lesson}) with causal supervision via curriculum learning, enabling them to capture causal dependencies on par with CHRL baselines (\textbf{\textit{D3HRL}}~\cite{ZHAO2026108275}, \textbf{\textit{CDHRL}}~\cite{hu2022causality}). All HRL baselines are further enhanced to handle a fixed uniform delay of $\tau_{\max}$. We provide a detailed description of the enhancement methodology and present a comparative analysis of algorithmic performance before and after enhancement in \ref{app:enhancement}.

\subsection{Results and Analysis}\label{sec:experiments}

To systematically evaluate the proposed DECHRL framework, we design a series of experiments that address three key research questions:

\begin{enumerate}
    \item \textbf{Baselines Comparison – DARL:} How does DECHRL compare with delay-aware RL (DARL) baselines? This comparison is used to evaluate the performance of different strategies for handling delays.
    \item \textbf{Baselines Comparison – HRL:} How does DECHRL compare with standard HRL baselines? This comparison is used to evaluate the performance of different hierarchical reinforcement learning methods in environments with stochastic delays.
    \item \textbf{Delay Modeling Accuracy:} How closely does the learned delay distribution match the ground-truth delay distribution? This evaluates the core capability of our causal delay modeling module to recover stochastic temporal dynamics from delayed observations.
    \item \textbf{Gradient Estimator Impact:} How does the choice of gradient estimator for modeling the delay distribution affect DECHRL’s performance? This assesses the sensitivity and suitability of different gradient approximation strategies in delay distribution modeling.
    \item \textbf{Empowerment Ablation:} How does the inclusion of the delay-aware empowerment objective affect overall performance? This isolates the role of empowerment as a stabilizing exploration mechanism under delay-induced uncertainty.
    \item \textbf{Robustness to Delay Stochasticity:} How does DECHRL perform under varying degrees of delay stochasticity? This tests the framework’s robustness to increasingly complex and uncertain delay structures.
    \item \textbf{Modeling Granularity \& Scale:} How does Simplified-DECHRL perform under different delay modeling granularities and maximum delays? This examines practical trade-offs in model capacity and scalability.
    \item \textbf{Hyperparameter Sensitivity:} How sensitive is DECHRL to its hyperparameters $\lambda_1$ and $\lambda_2$ (for delay modeling) and to the sub-goal success ratio threshold? This evaluates the ease of tuning and stability of the method in practice.
    \item \textbf{Sensitivity to Sub-Goal Success Criterion:} How sensitive is DECHRL to the threshold used to determine sub-goal success? This evaluates the robustness of the hierarchical credit assignment mechanism to variations in the success criterion, which directly influences high-level policy learning and temporal abstraction quality.
\end{enumerate}
The subsequent subsections present empirical results corresponding to each question, providing a comprehensive evaluation of DECHRL’s practical effectiveness and algorithmic design.

\subsubsection{How does DECHRL compare with the DARL baselines?}\label{sec:delay_method_comparison}

\begin{figure}[htbp]
    \centering
    \renewcommand{\arraystretch}{0.6} 
    \resizebox{\textwidth}{!}{
    \setlength{\tabcolsep}{2pt}
    \begin{tabular}{c >{\centering\arraybackslash}m{3cm} >{\centering\arraybackslash}m{3cm} >{\centering\arraybackslash}m{3cm} >{\centering\arraybackslash}m{3cm} >{\centering\arraybackslash}m{3cm}}
     & Get Wood & Get Stone & Get Stick & Get Stone Axe & Get Silver Ore\\
    \makecell*[c]{$\tau_{\max}=4$\\$\sigma_{delay}=0.4$}
    &\includegraphics[width=1.0\linewidth]{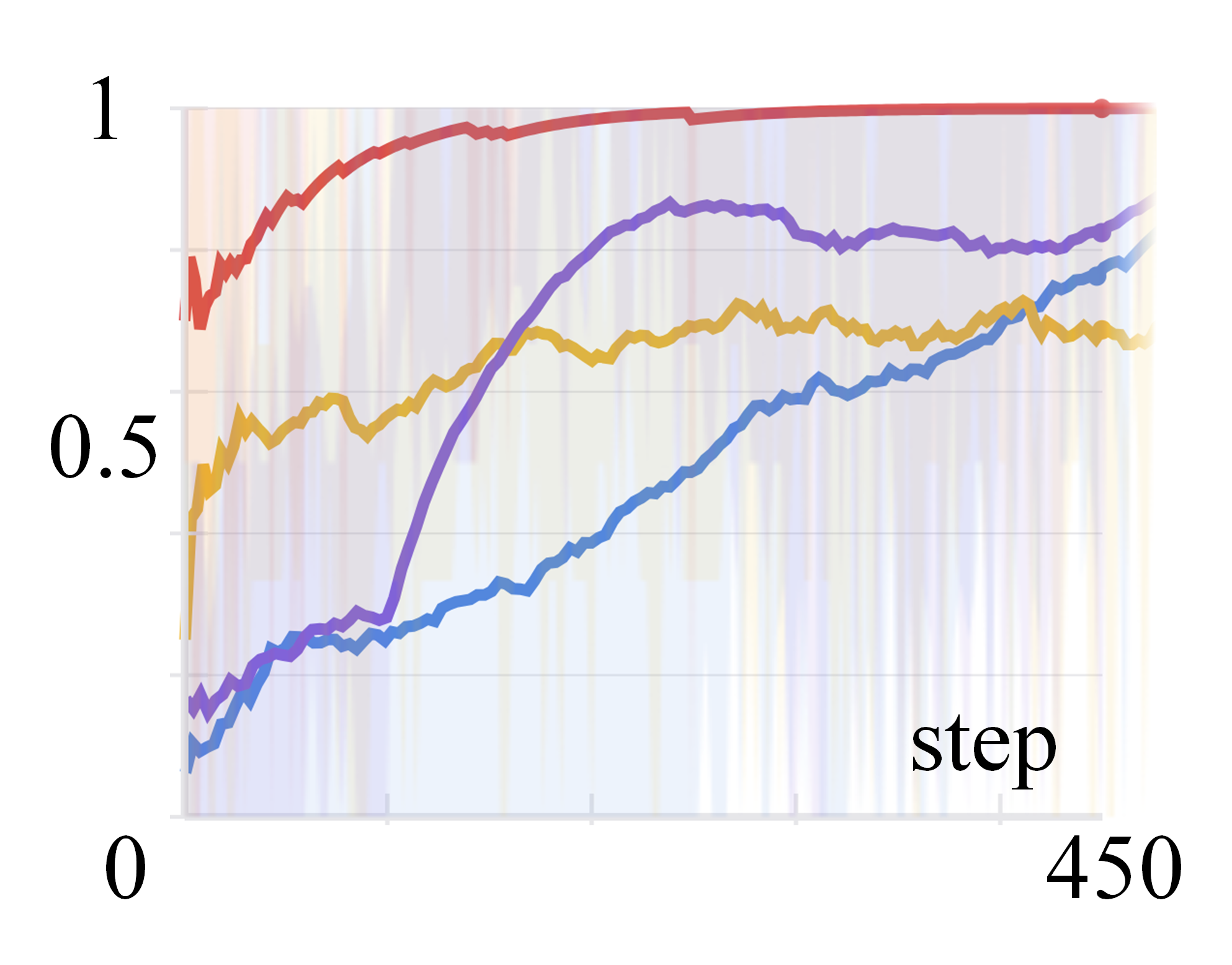}
    &\includegraphics[width=1.0\linewidth]{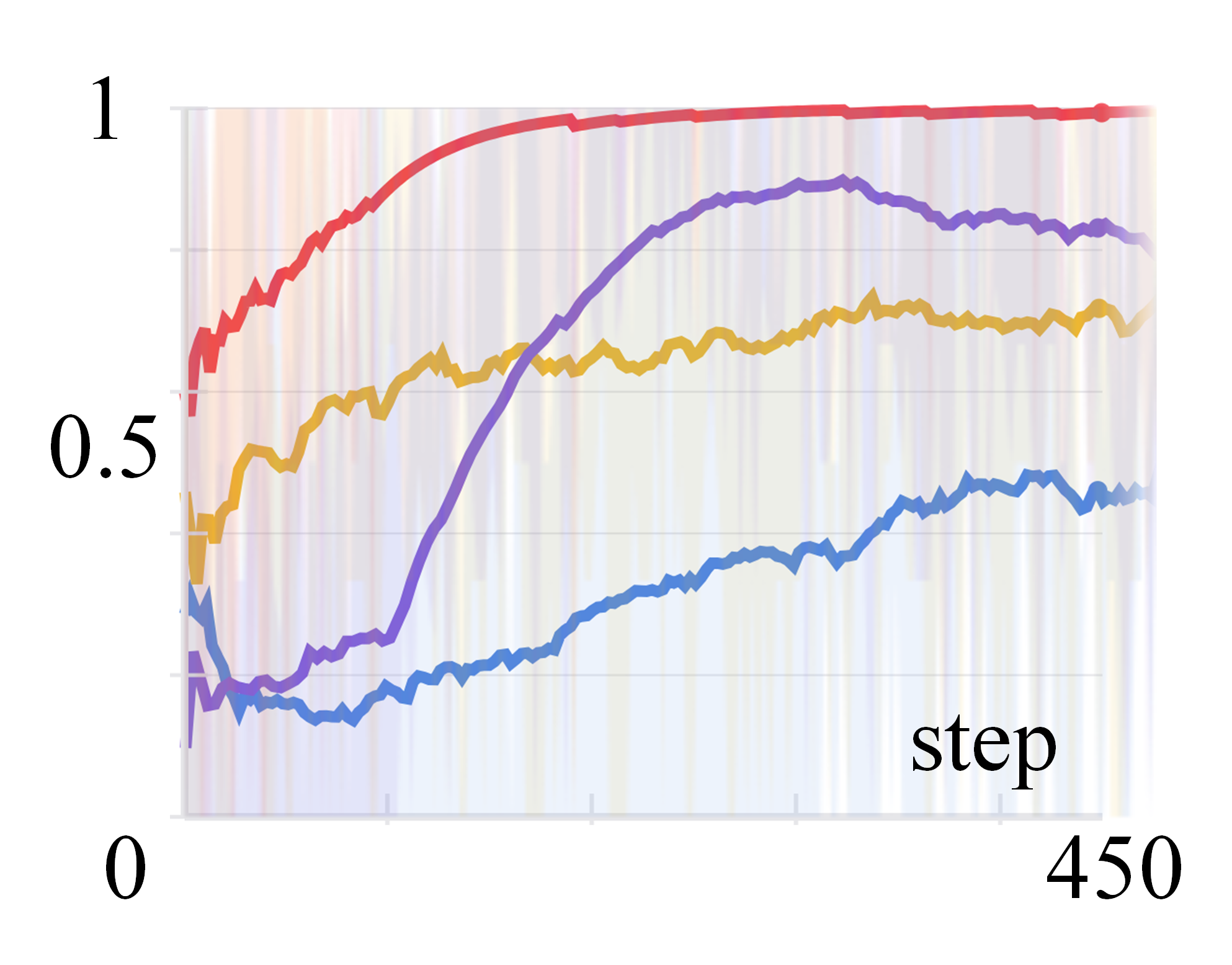}
    &\includegraphics[width=1.0\linewidth]{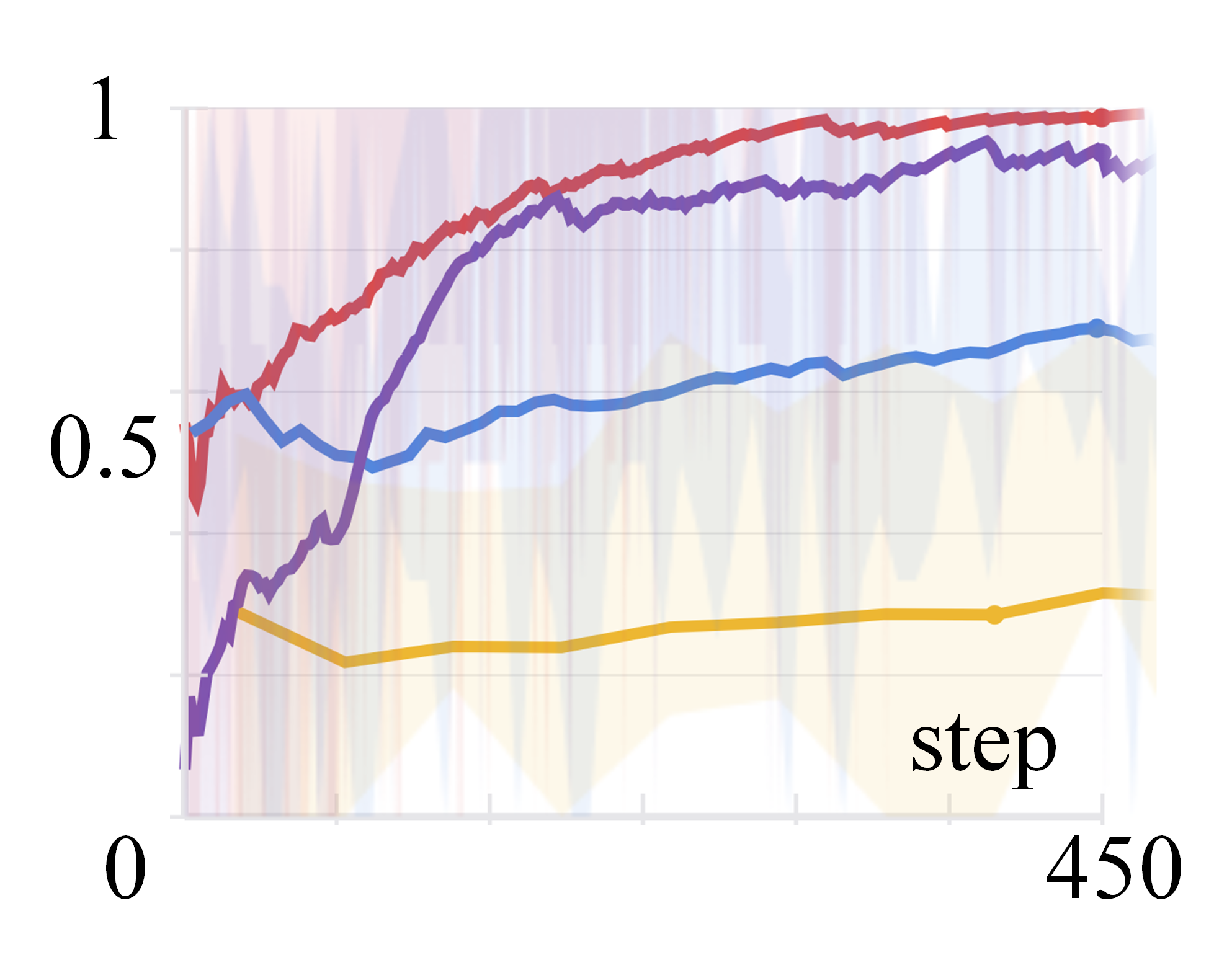}
    &\includegraphics[width=1.0\linewidth]{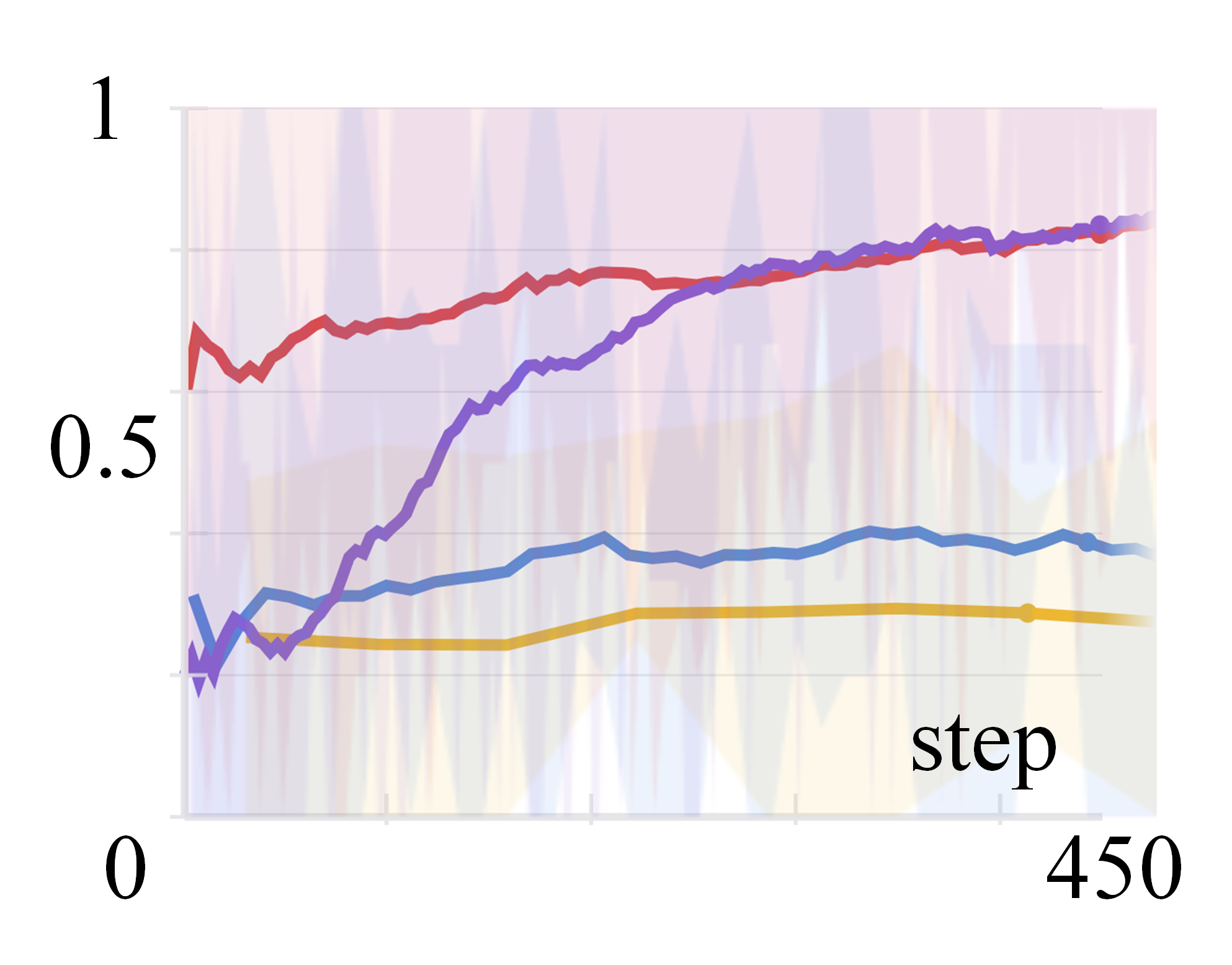}
    &\includegraphics[width=1.0\linewidth]{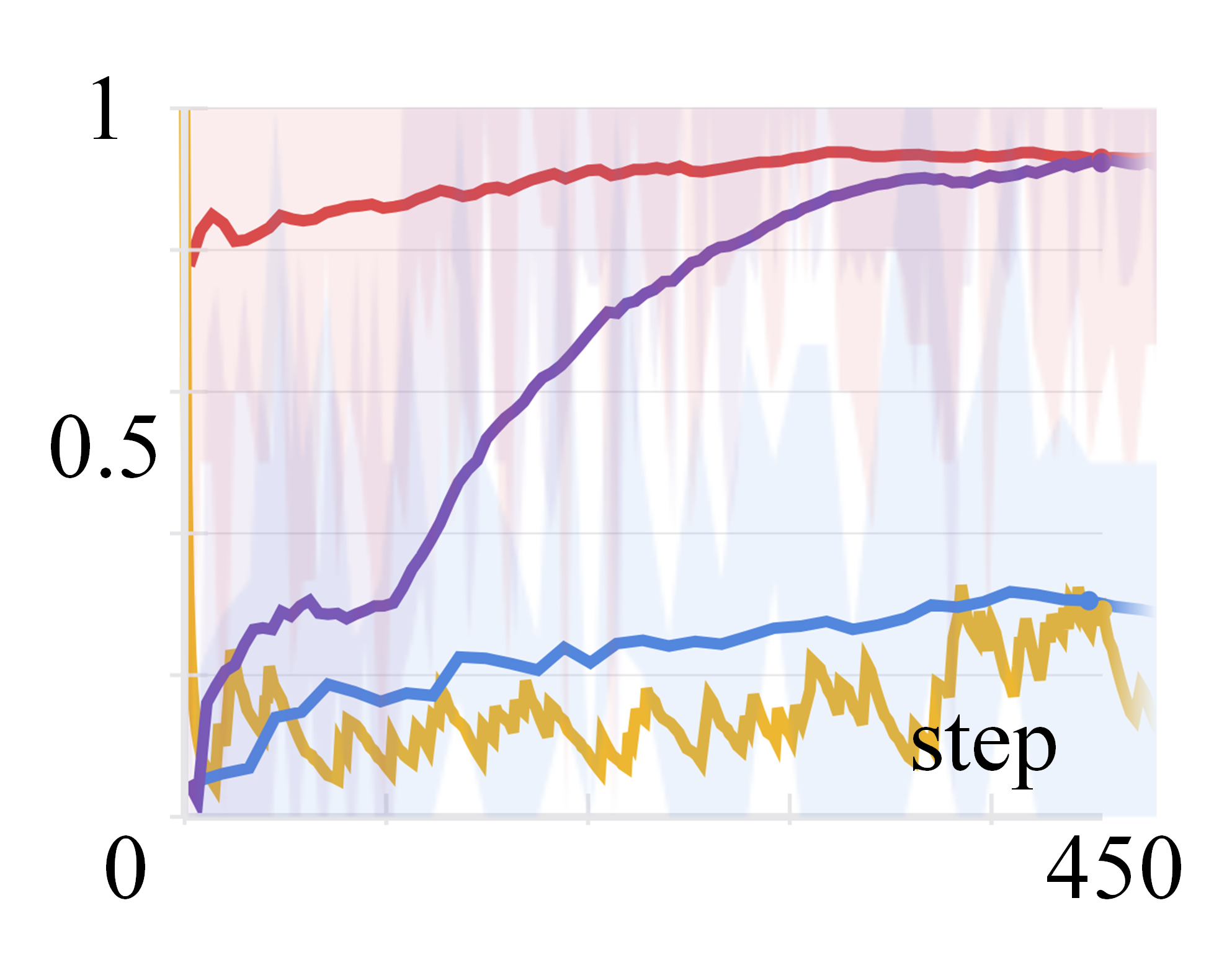} \\
    
    \makecell*[c]{$\tau_{\max}=4$\\$\sigma_{delay}=0.8$}
    &\includegraphics[width=1.0\linewidth]{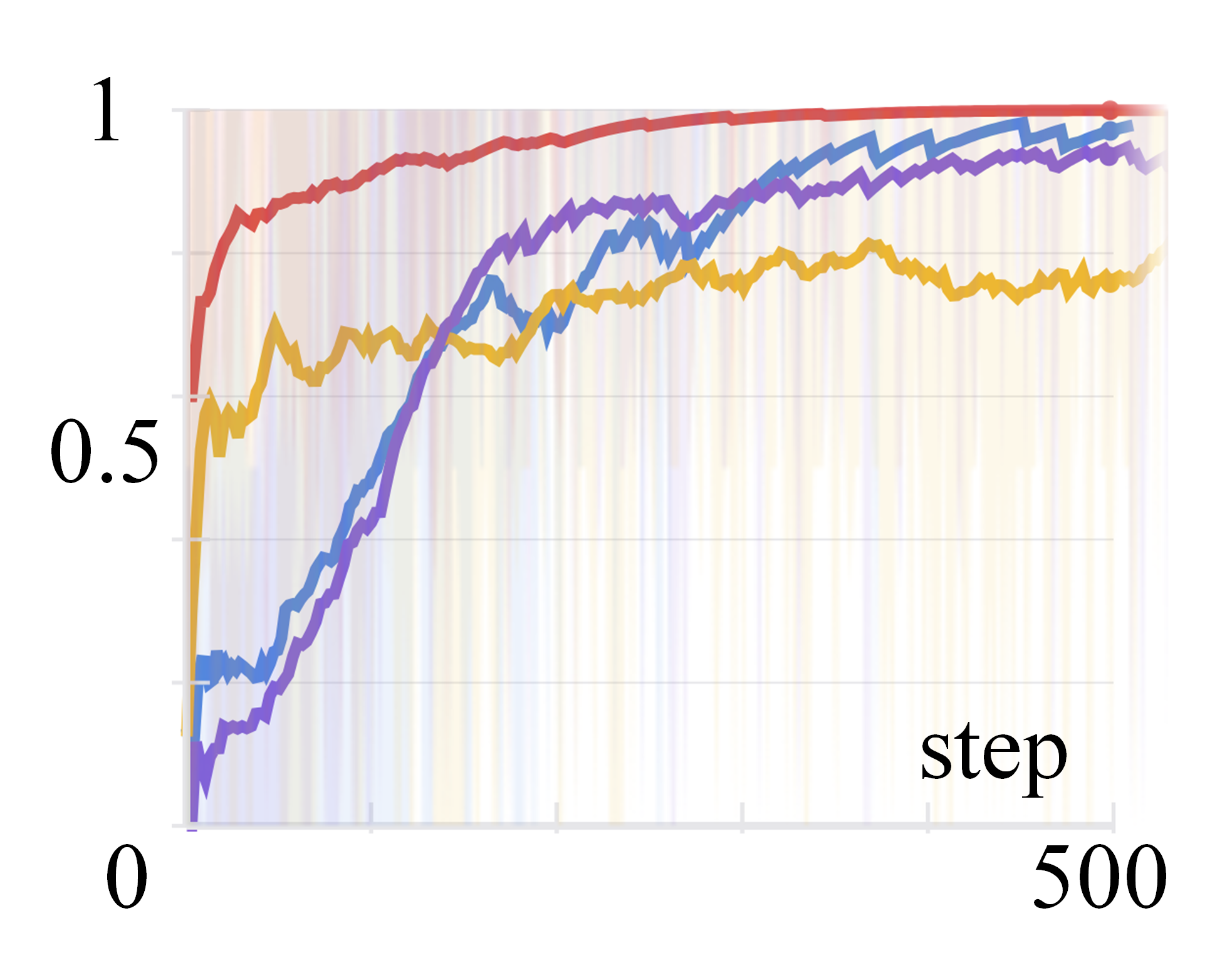}
    &\includegraphics[width=1.0\linewidth]{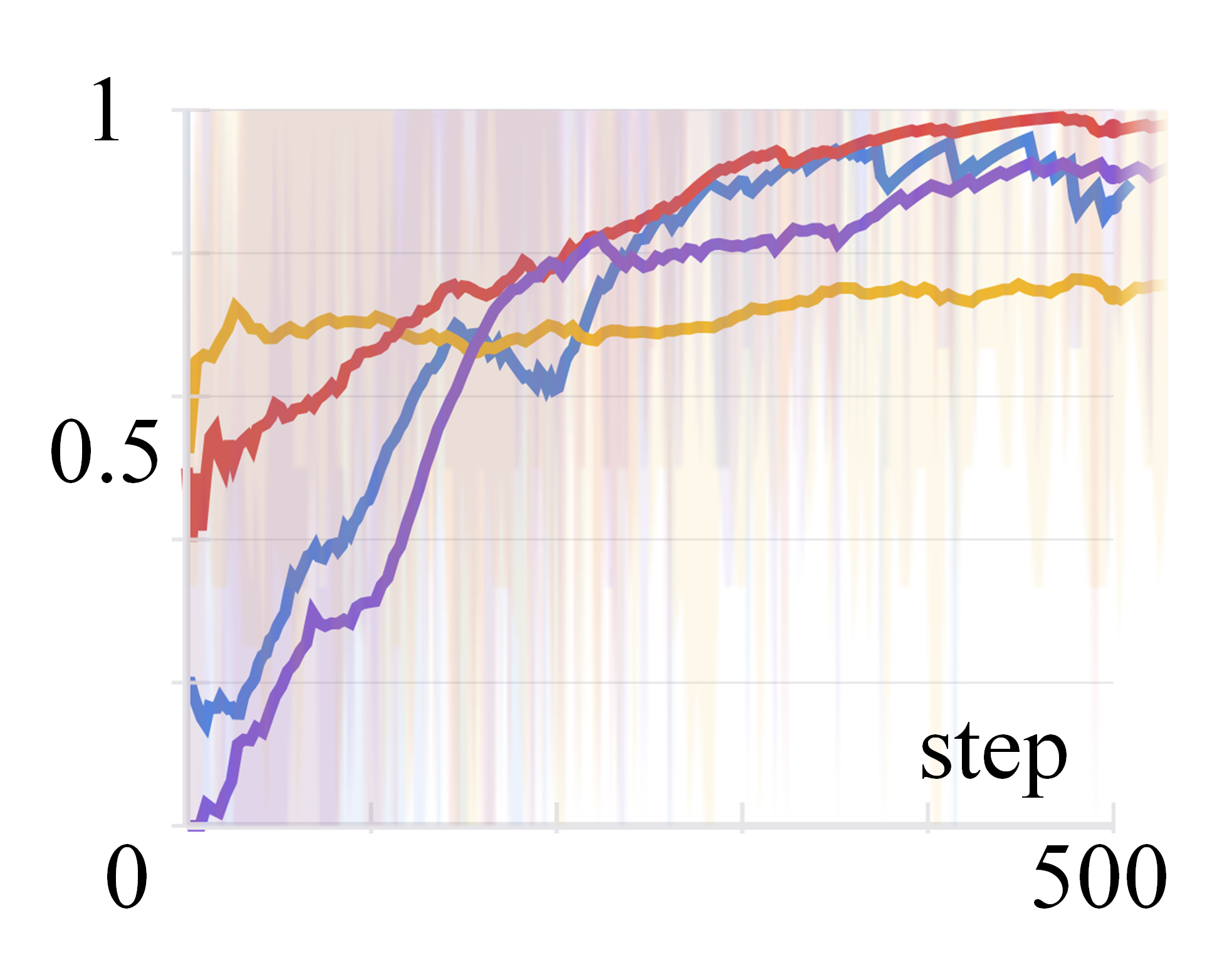}
    &\includegraphics[width=1.0\linewidth]{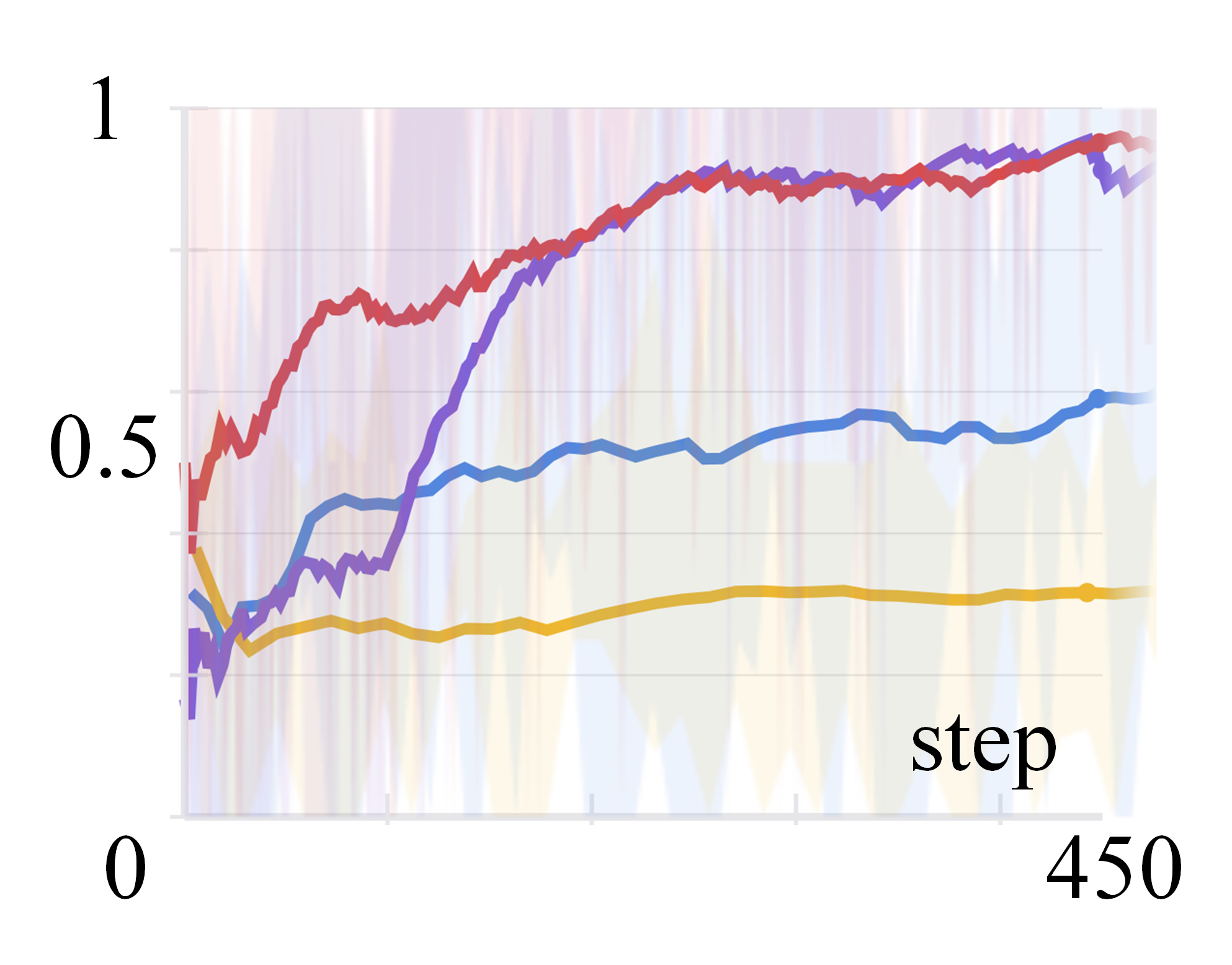}
    &\includegraphics[width=1.0\linewidth]{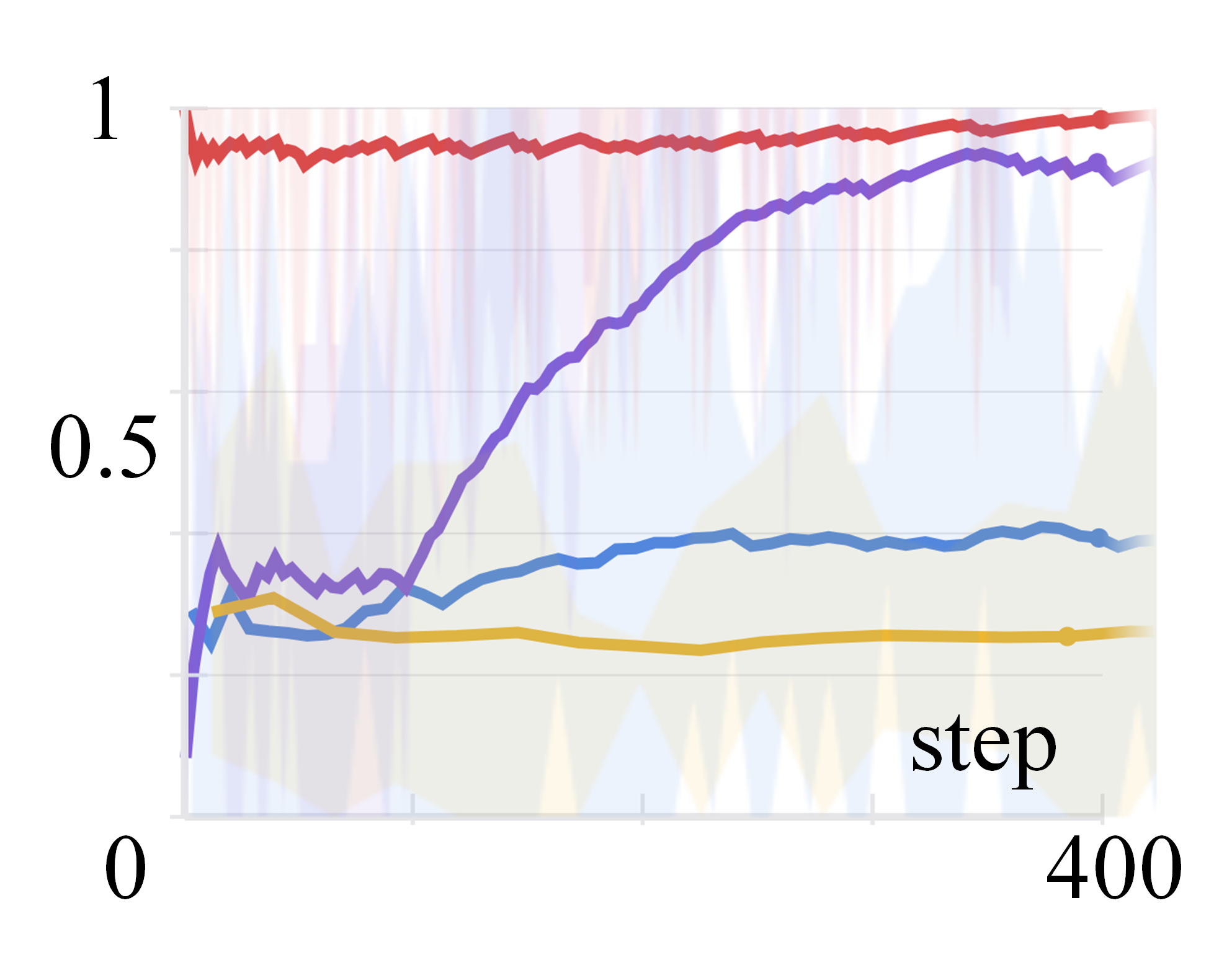}
    &\includegraphics[width=1.0\linewidth]{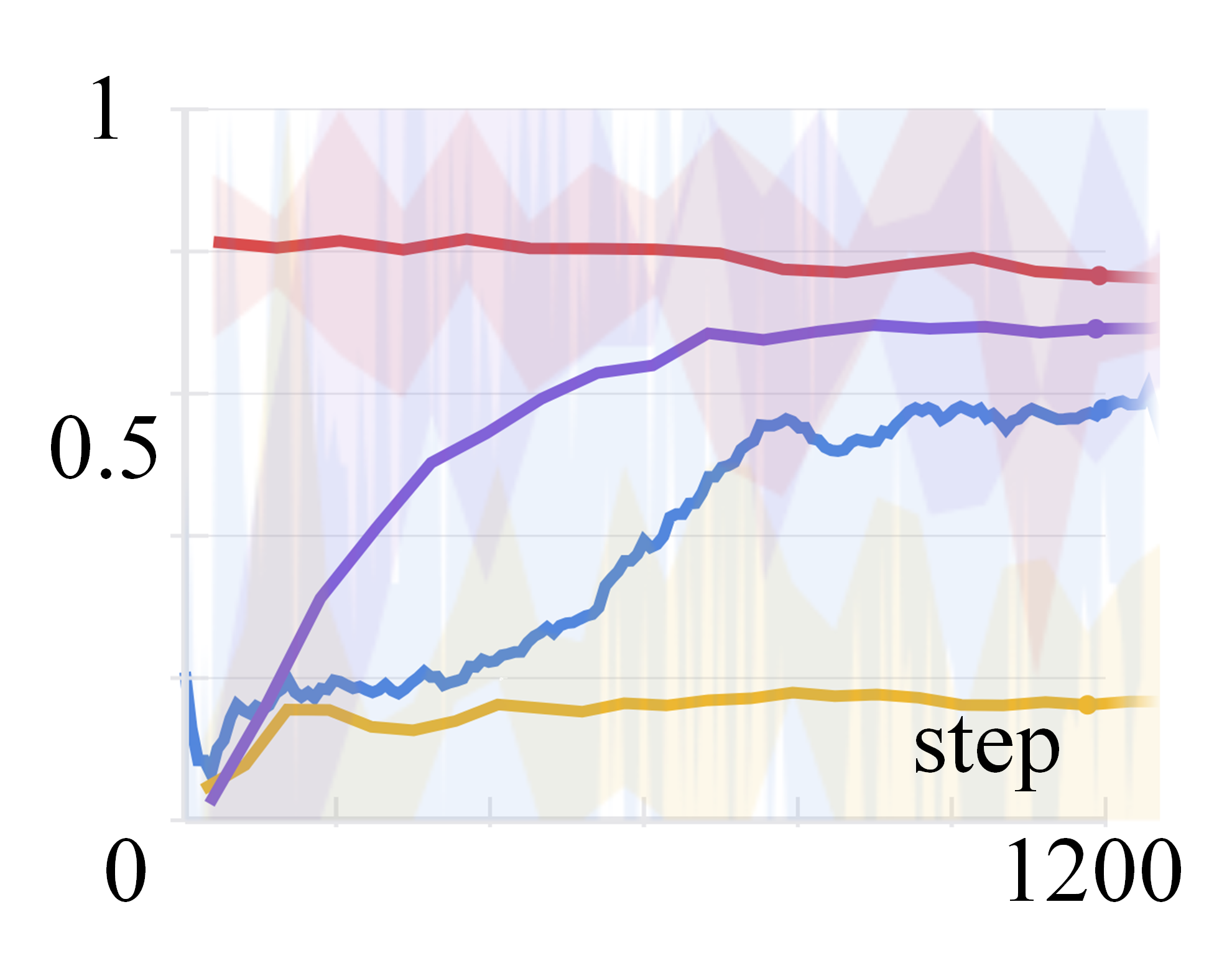} \\
    
    \makecell*[c]{$\tau_{\max}=8$\\$\sigma_{delay}=0.4$}
    &\includegraphics[width=1.0\linewidth]{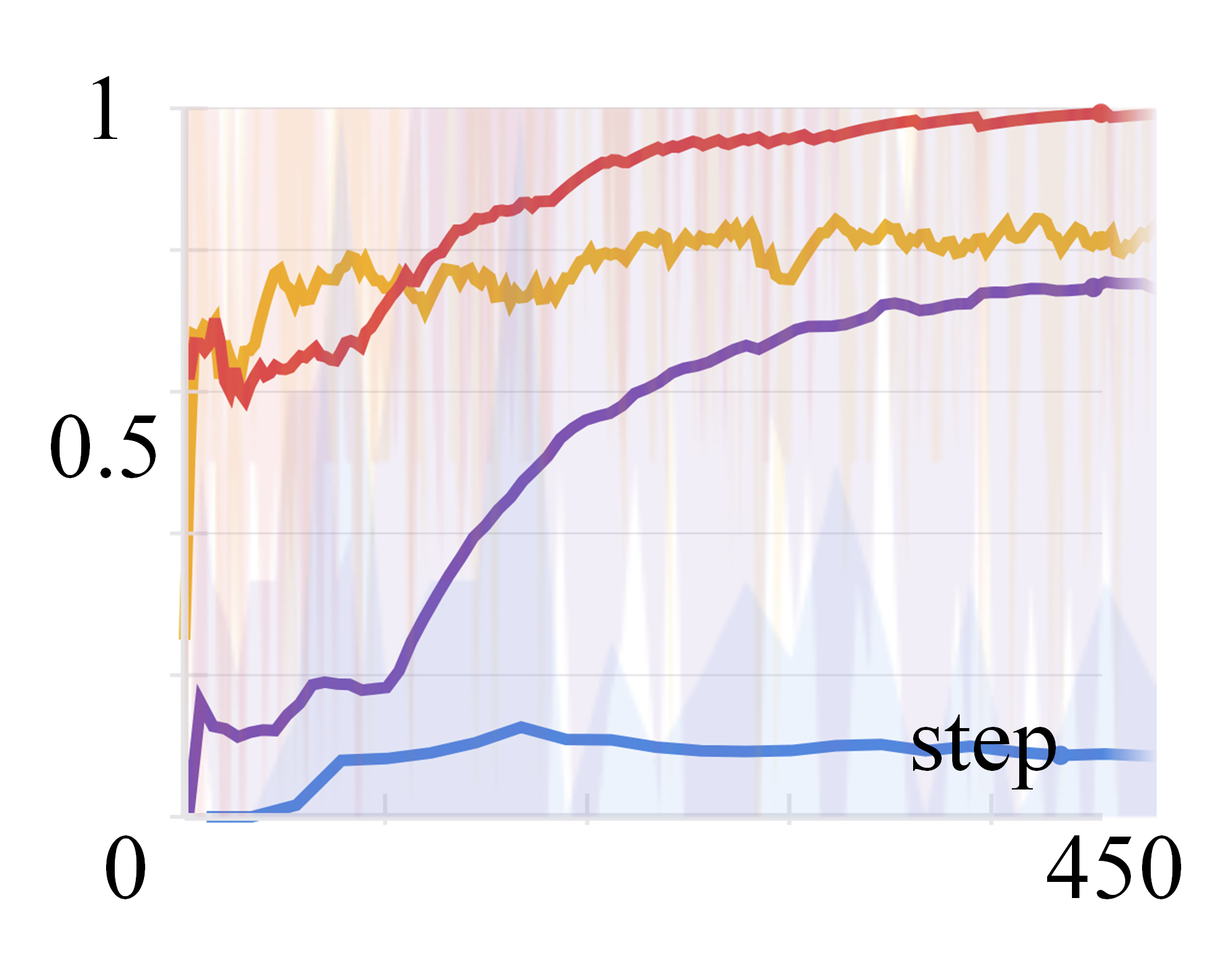}
    &\includegraphics[width=1.0\linewidth]{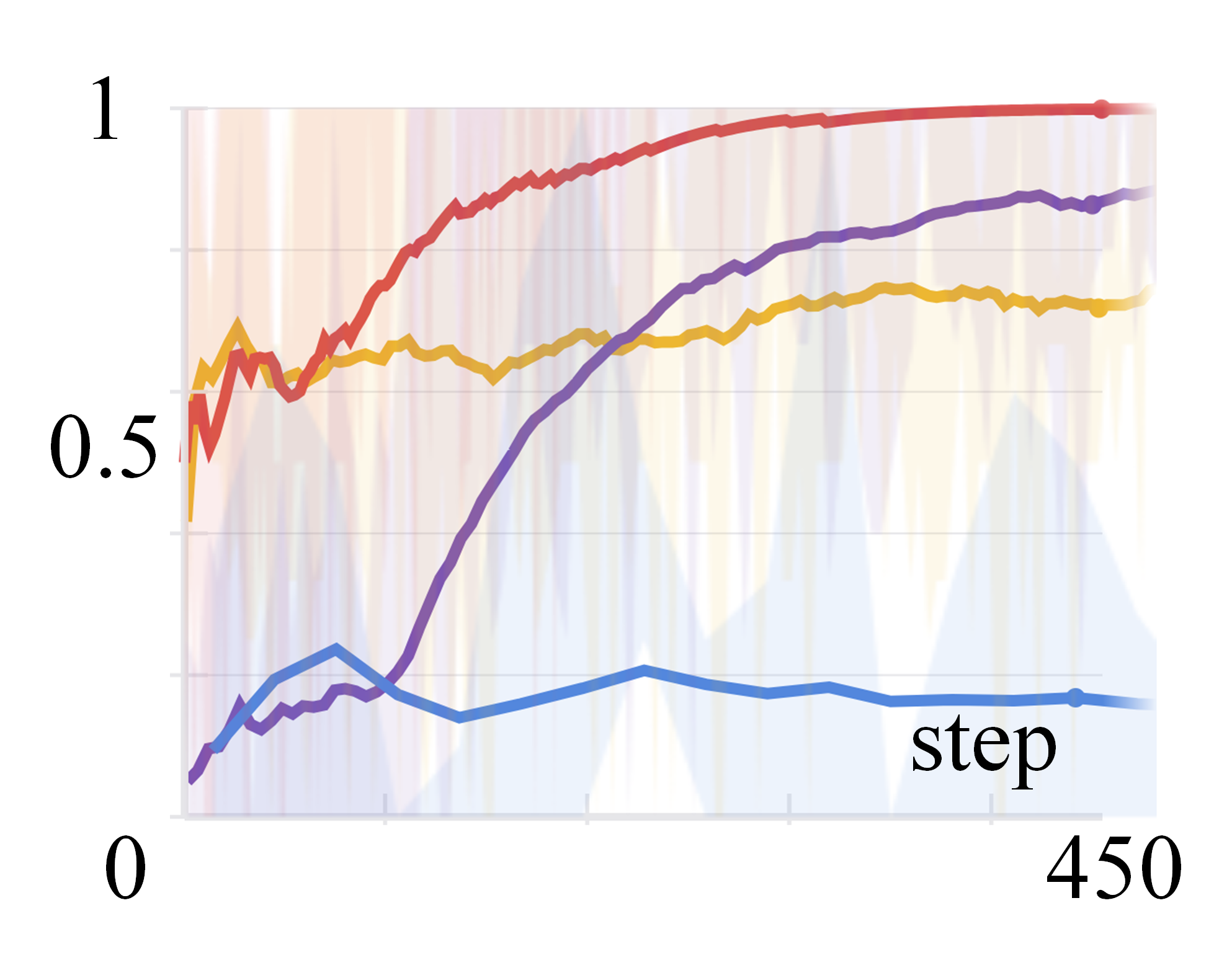}
    &\includegraphics[width=1.0\linewidth]{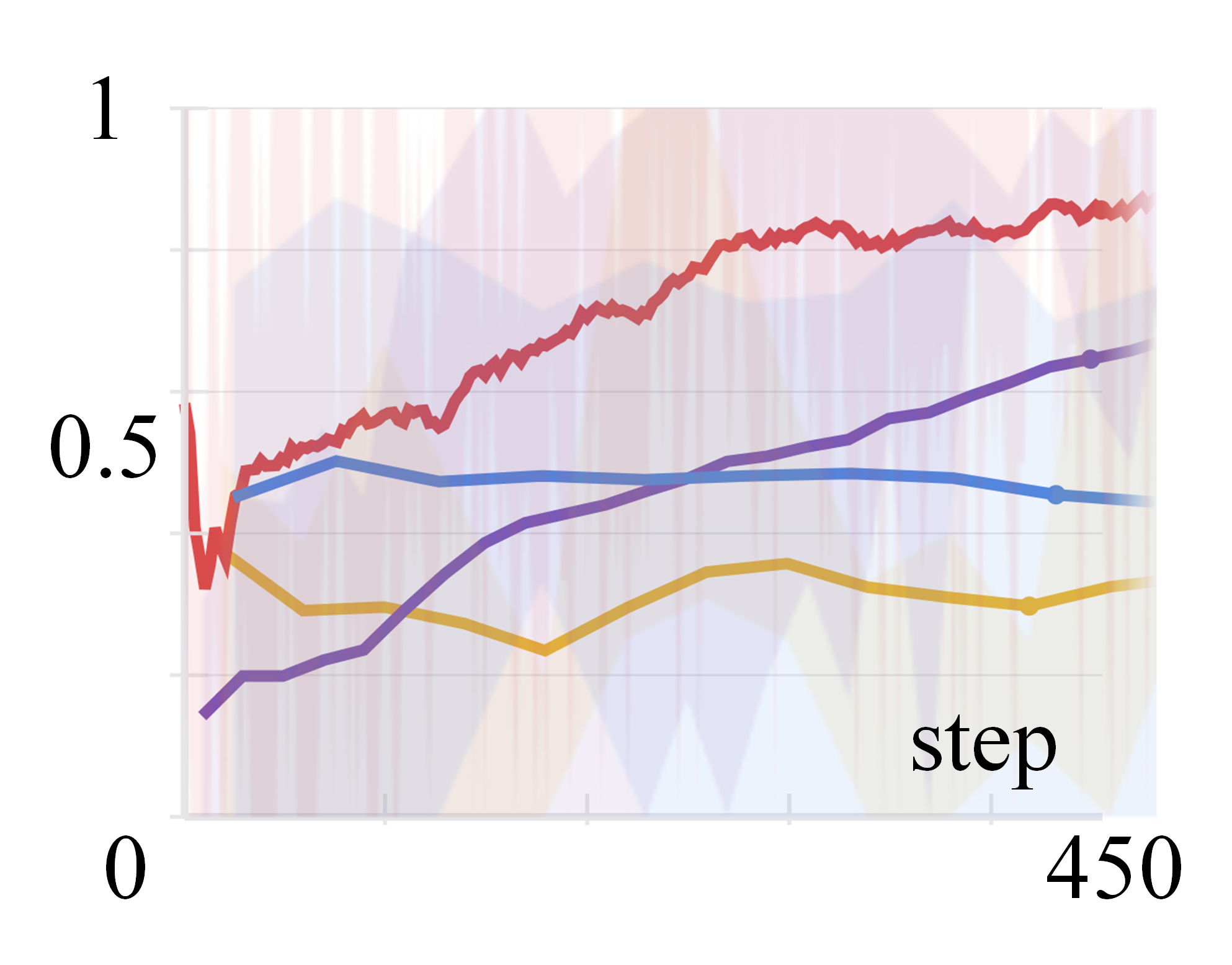}
    &\includegraphics[width=1.0\linewidth]{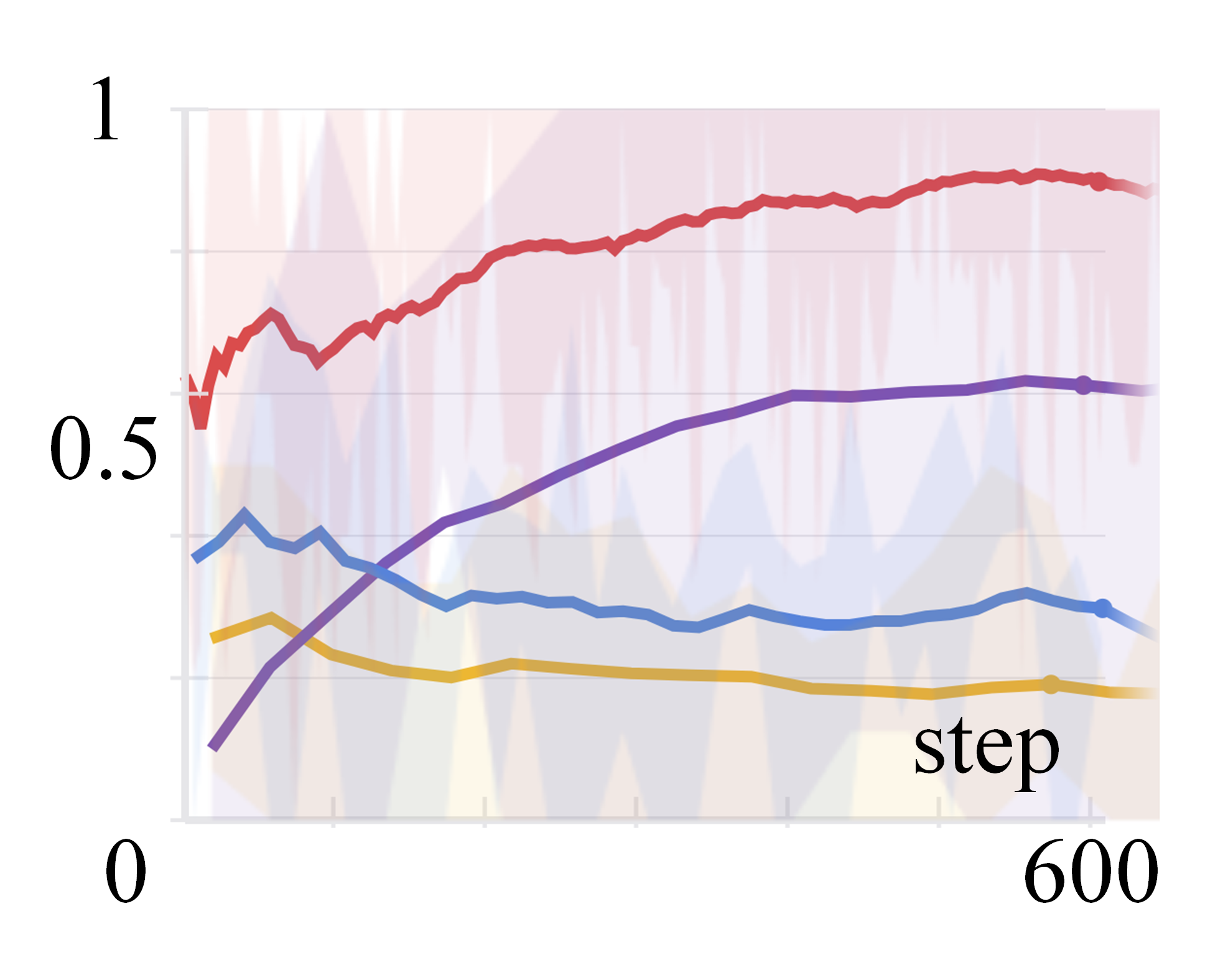}
    &\includegraphics[width=1.0\linewidth]{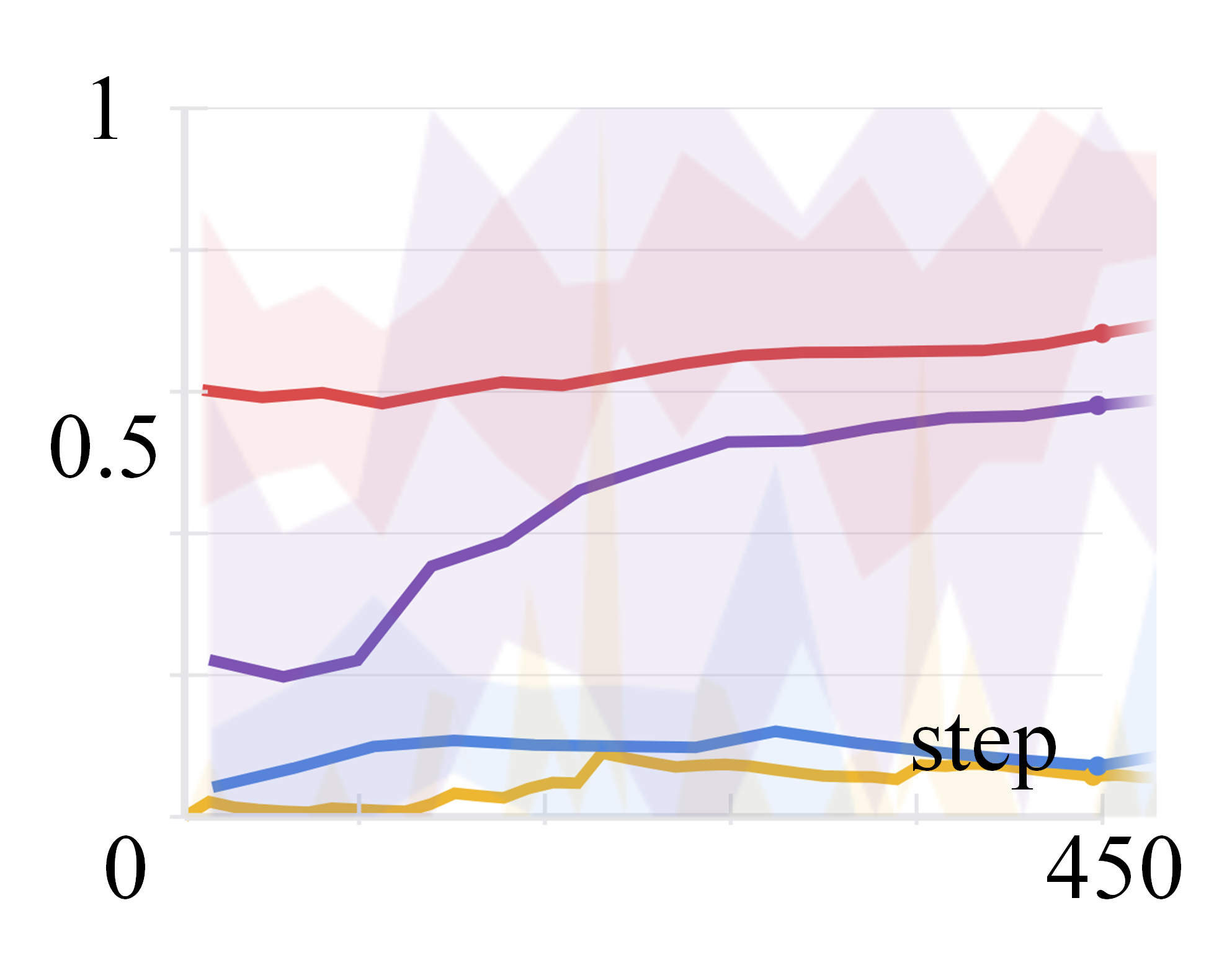} \\
    
    \makecell*[c]{$\tau_{\max}=8$\\$\sigma_{delay}=0.8$}
    &\includegraphics[width=1.0\linewidth]{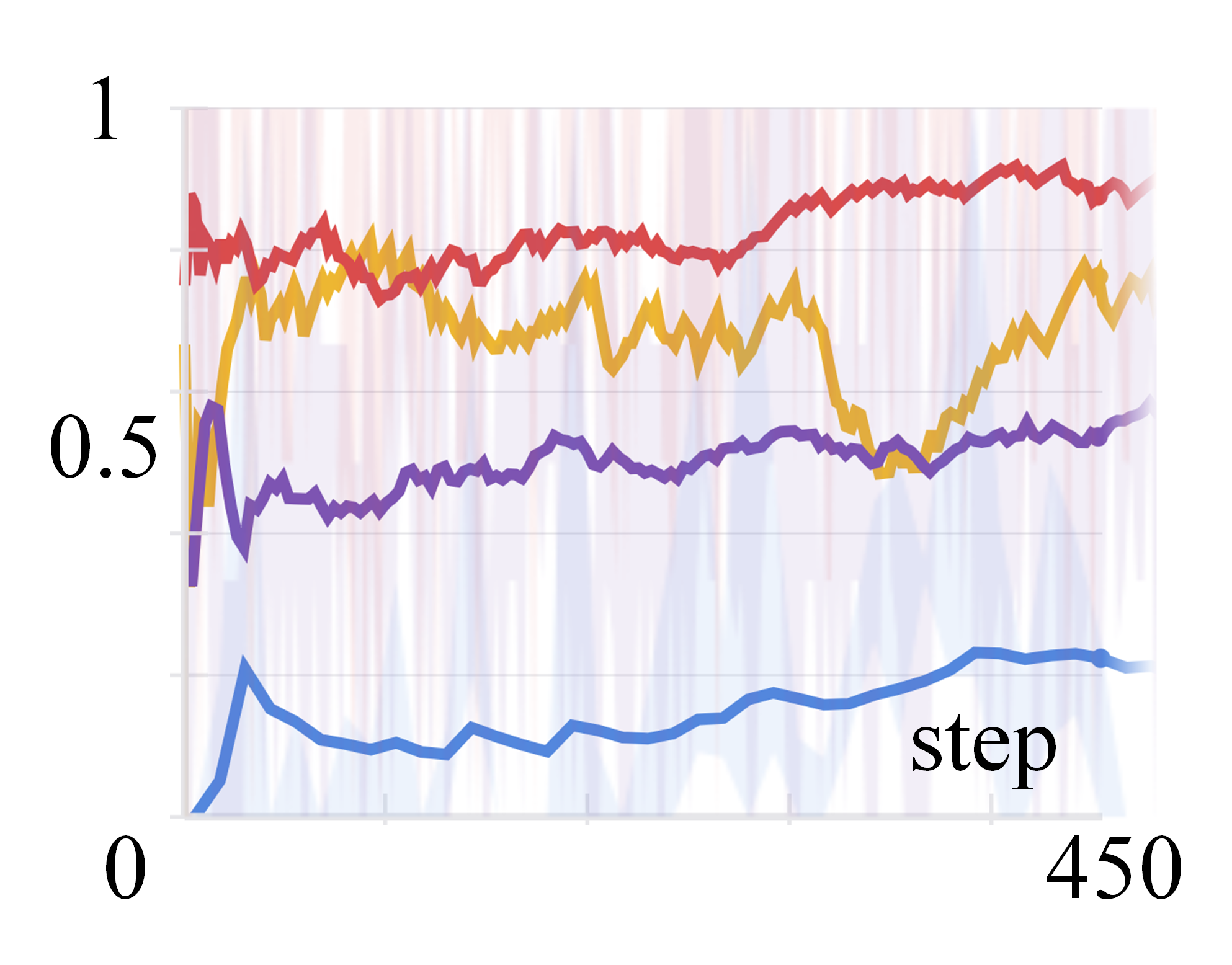}
    &\includegraphics[width=1.0\linewidth]{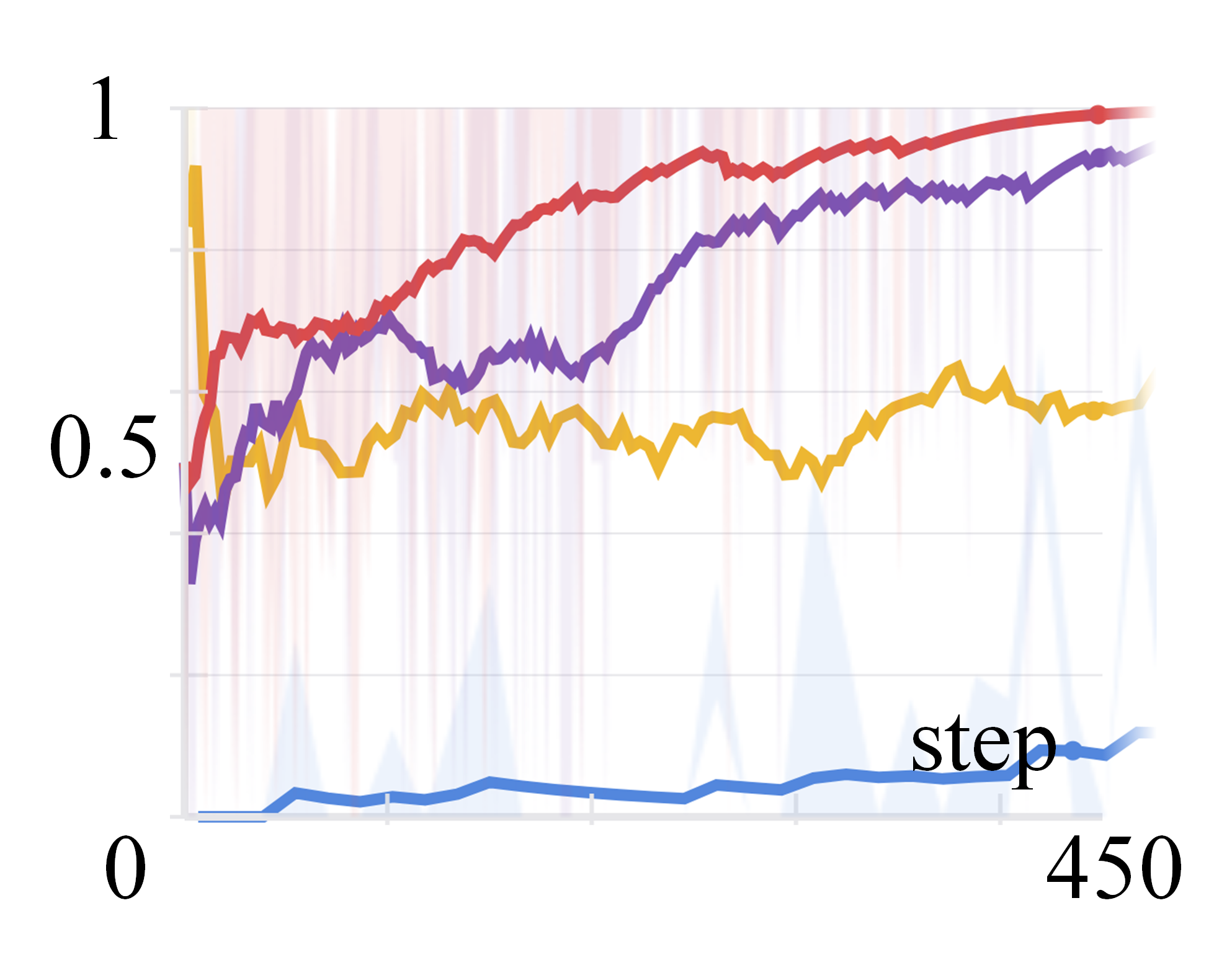}
    &\includegraphics[width=1.0\linewidth]{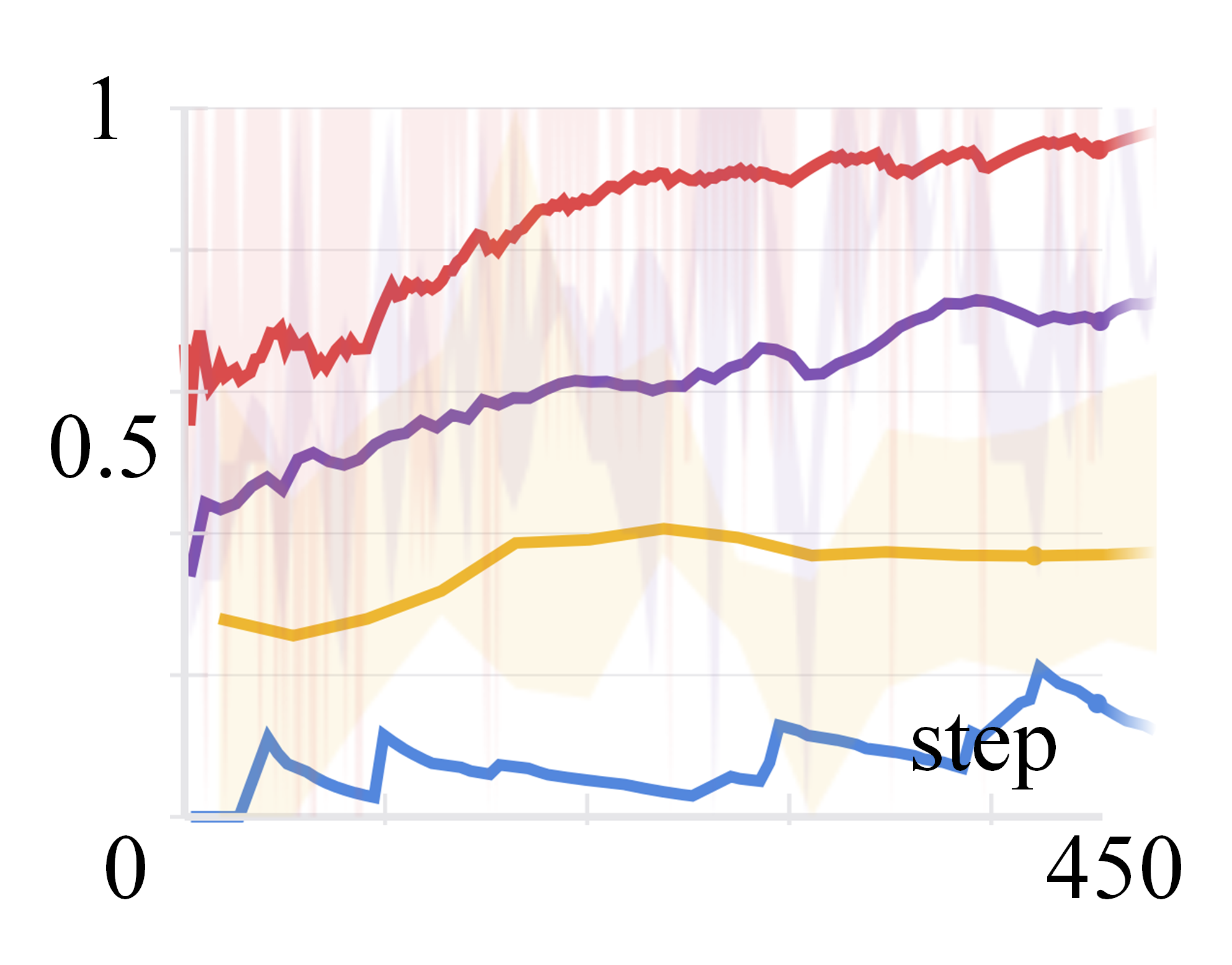}
    &\includegraphics[width=1.0\linewidth]{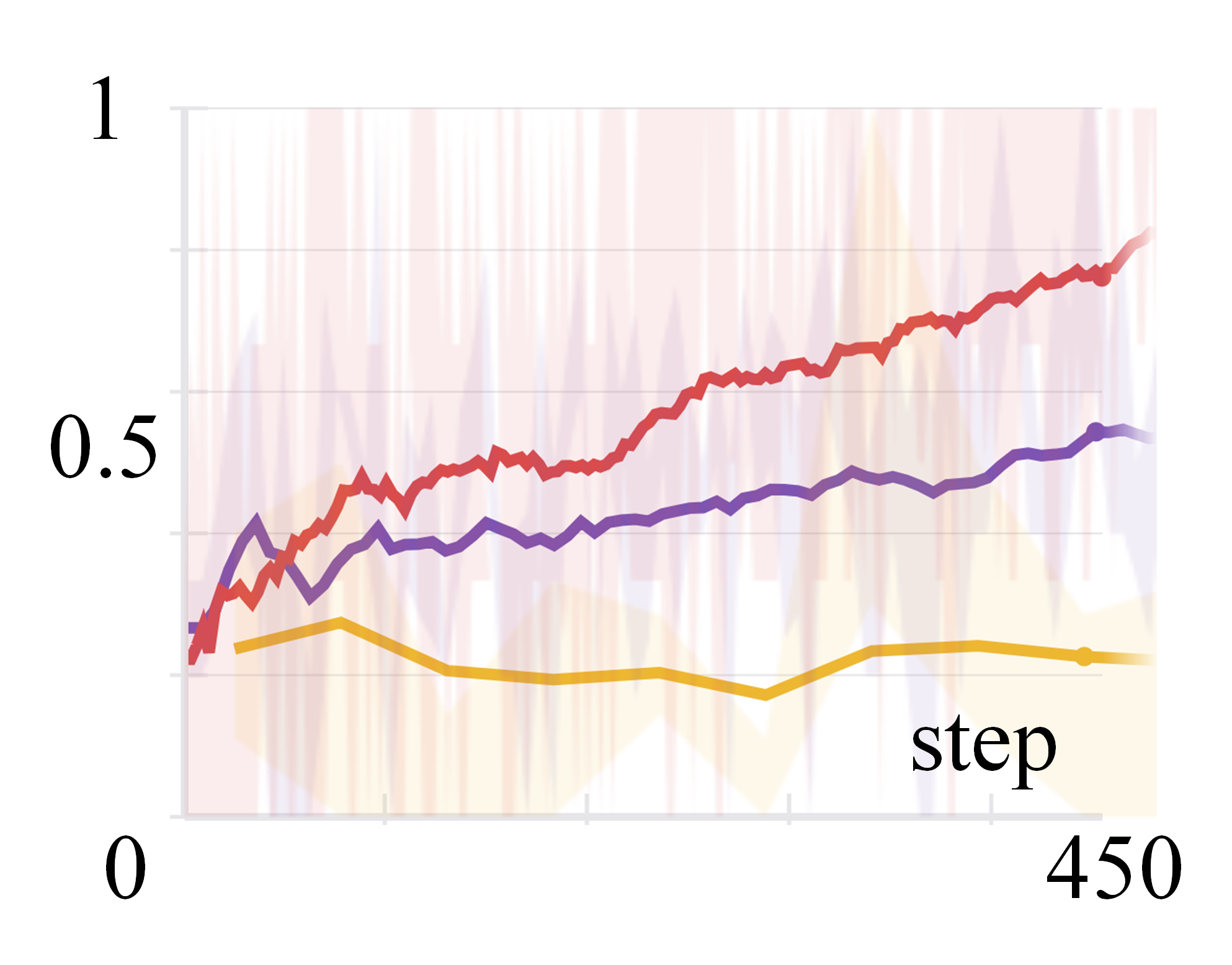}
    &\includegraphics[width=1.0\linewidth]{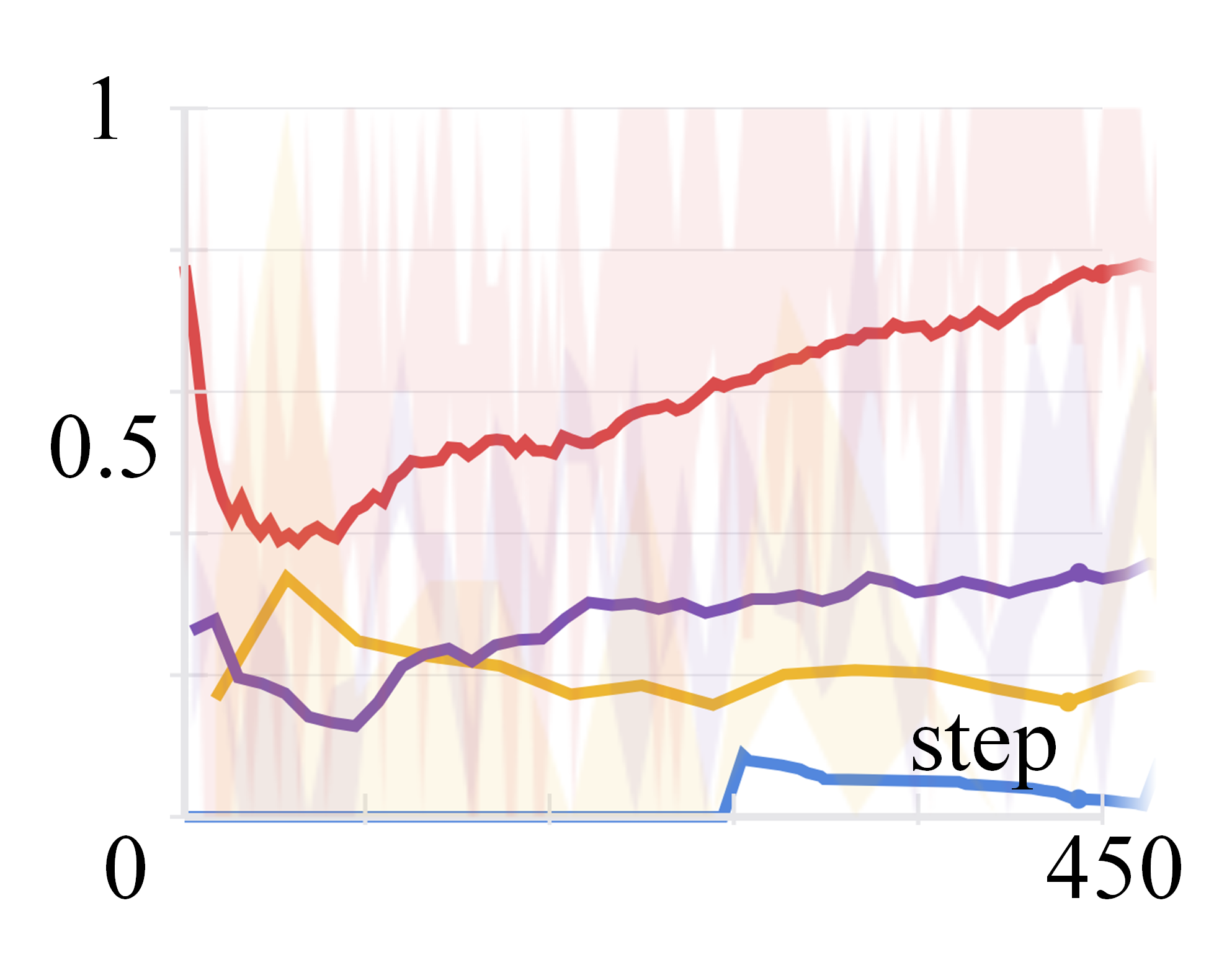} \\

    \multicolumn{6}{c}{\includegraphics[width=1.2\linewidth]{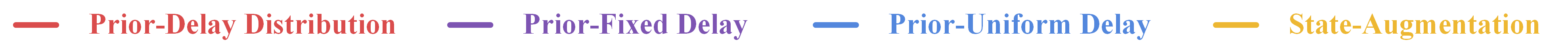}}\\
    
    \end{tabular}
    }
    \caption{The sub-goal training efficiency of different delay-handling methods across varying delay settings on the \textit{GetSilverore} task.}\label{tab:delay_handling_results} 
\end{figure}

\indent\textbf{\textit{Experimental Design.}} 
We integrate the delay handling method of two foundational DARL categories into the same HRL architecture as ours:
\textbf{(1) Memory-Based}: Augments states with historical trajectories during policy training; we term this variant \textit{State-Augmentation}.
\textbf{(2) Prior Knowledge-Based}: Leverages prior knowledge about delays for policy training. We implement two variants that differ only in their delay prior representation:
\textit{(i) Prior-Fixed Delay}: Applies a fixed, transition-specific delay $\mu_{\text{delay}}$;
\textit{(ii) Prior-Uniform Delay}: Applies a uniform delay $\tau_{\max}$ to all transitions.
To ensure a fair comparison, we implement \textit{Prior-Delay Distribution}, which incorporates the delay distribution learned by DECHRL as a prior into the same HRL architecture but omits the delay-aware empowerment bonus.
We evaluate the sub-goal training efficiency of these variants on the \textit{GetSilverore} task across $\tau_{\max} = \{4,8\}$ and $\sigma_{delay} = \{0.4,0.8\}$, as shown in Figure~\ref{tab:delay_handling_results}.

\indent\textbf{\textit{Experimental Results.}} 
\textit{State-Augmentation} achieves moderate robustness but consistently underperforms our \textit{Prior-Delay Distribution}. \textit{Prior-Uniform Delay} degrades sharply with increasing $\tau_{\max}$, while \textit{\textit{Prior-Fixed Delay}} suffers under high $\sigma_{delay}$ due to unmodeled uncertainty. In contrast, our \textit{Prior-Delay Distribution}, leveraging the learned delay distribution—consistently outperforms all baselines across all settings, demonstrating superior robustness.
\textit{Notably}, within the unified HRL framework without Delay-Aware Empowerment, ablating the delay distribution modeling module reduces the algorithm to using fixed delays, learned via causal discovery, as priors for policy training. Thus, \textbf{the comparison between \textit{Prior-Delay Distribution} and \textit{Prior-Fixed Delay} serves as an ablation study on modeling the delay distribution, rendering a separate evaluation unnecessary.}

How does DECHRL compare with the HRL baselines?

\subsubsection{How does DECHRL compare with the HRL baselines?}\label{sec:exp_genernal}

\begin{figure}[htbp]
    \centering
    \includegraphics[width=1.0\textwidth]{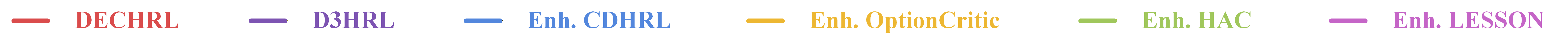}
    \setlength{\tabcolsep}{3pt} 
    \begin{tabular}{ccc}
        \subcaptionbox{T1 $(\tau_{\max}{=}4, \sigma_{delay}{=}0.4)$}{
            \includegraphics[width=0.16\textwidth]{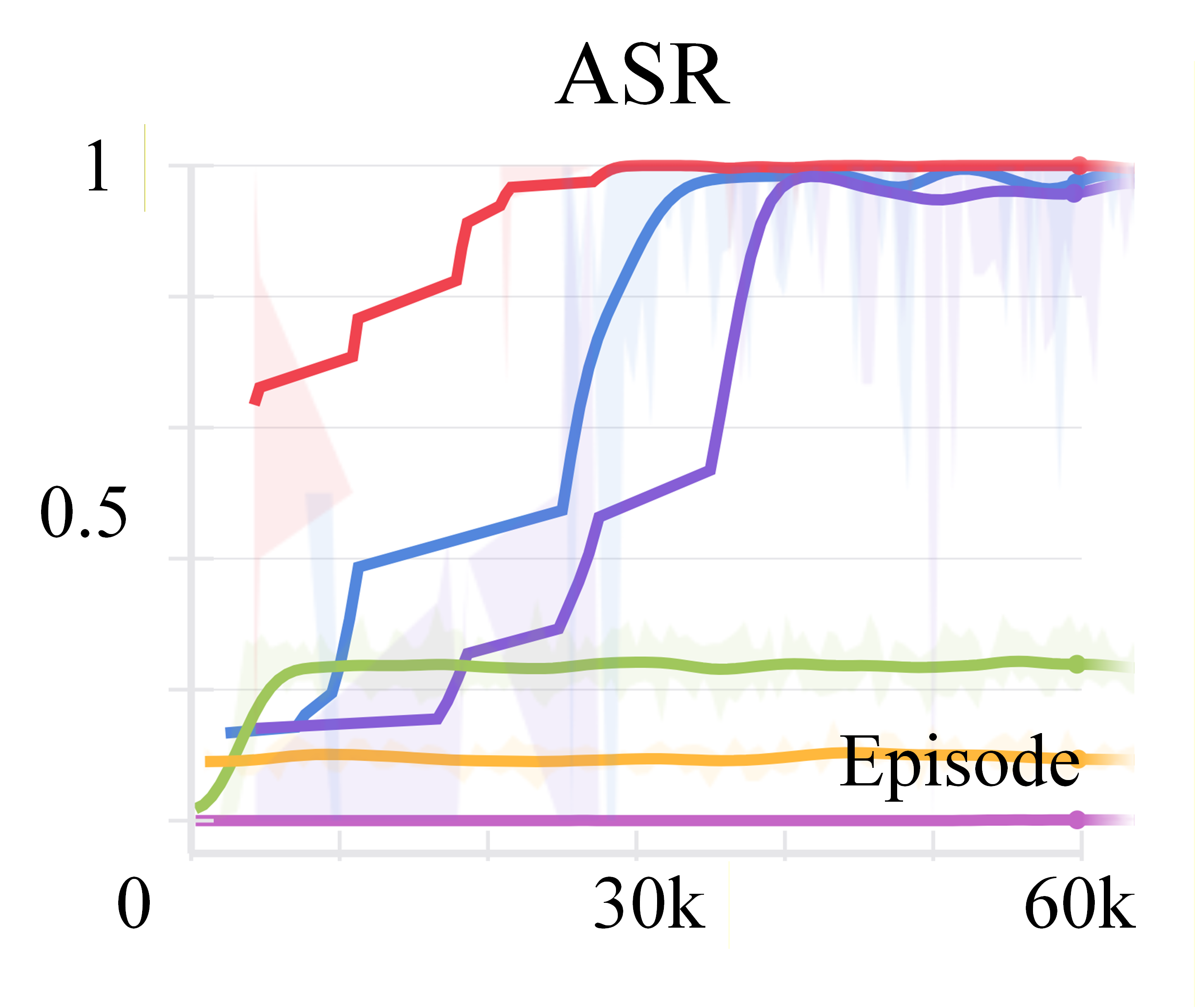}
            \hspace{-0.16cm}
            \includegraphics[width=0.16\textwidth]{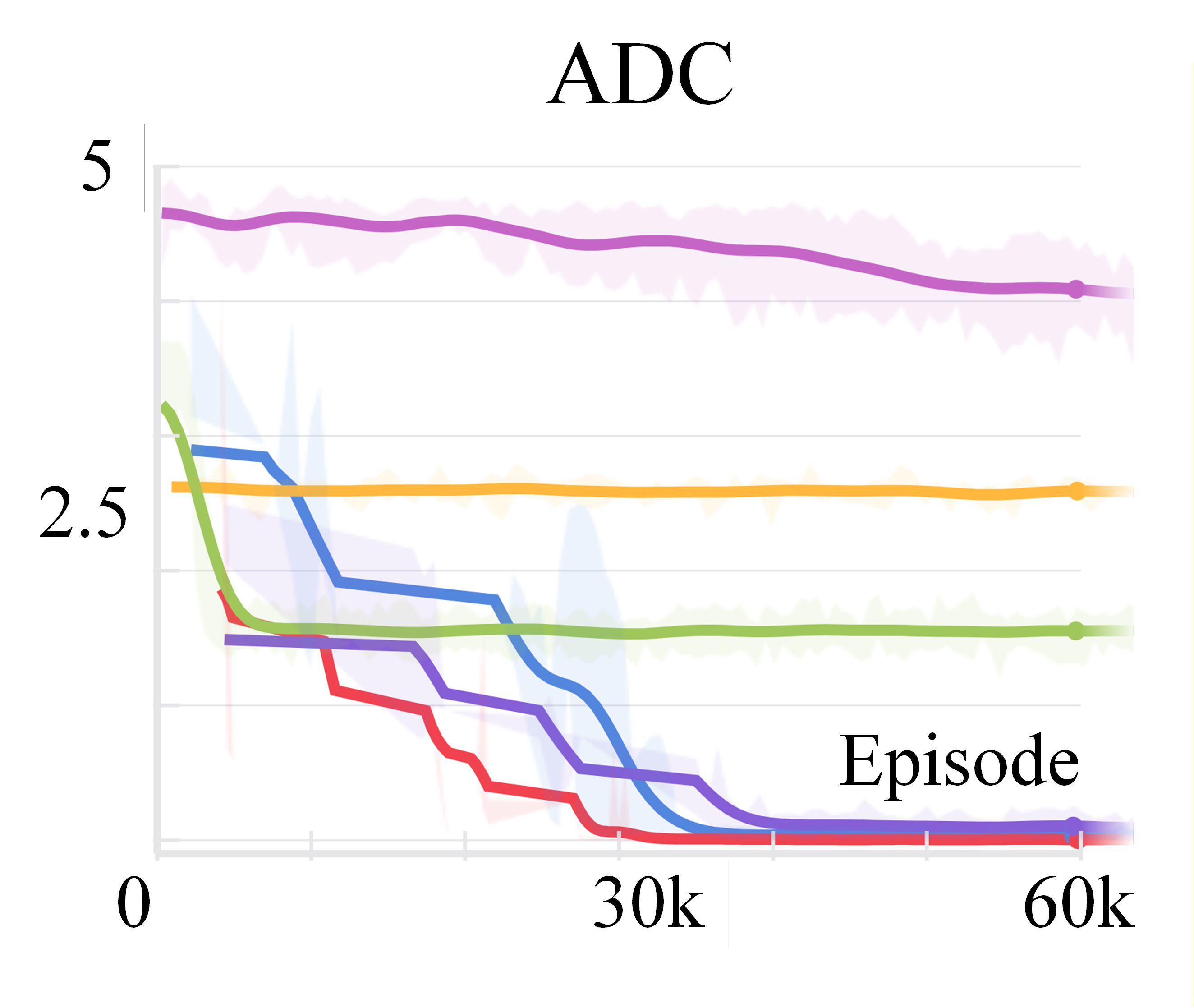}} &
        \subcaptionbox{T2 $(\tau_{\max}{=}4, \sigma_{delay}{=}0.4)$}{
            \includegraphics[width=0.16\textwidth]{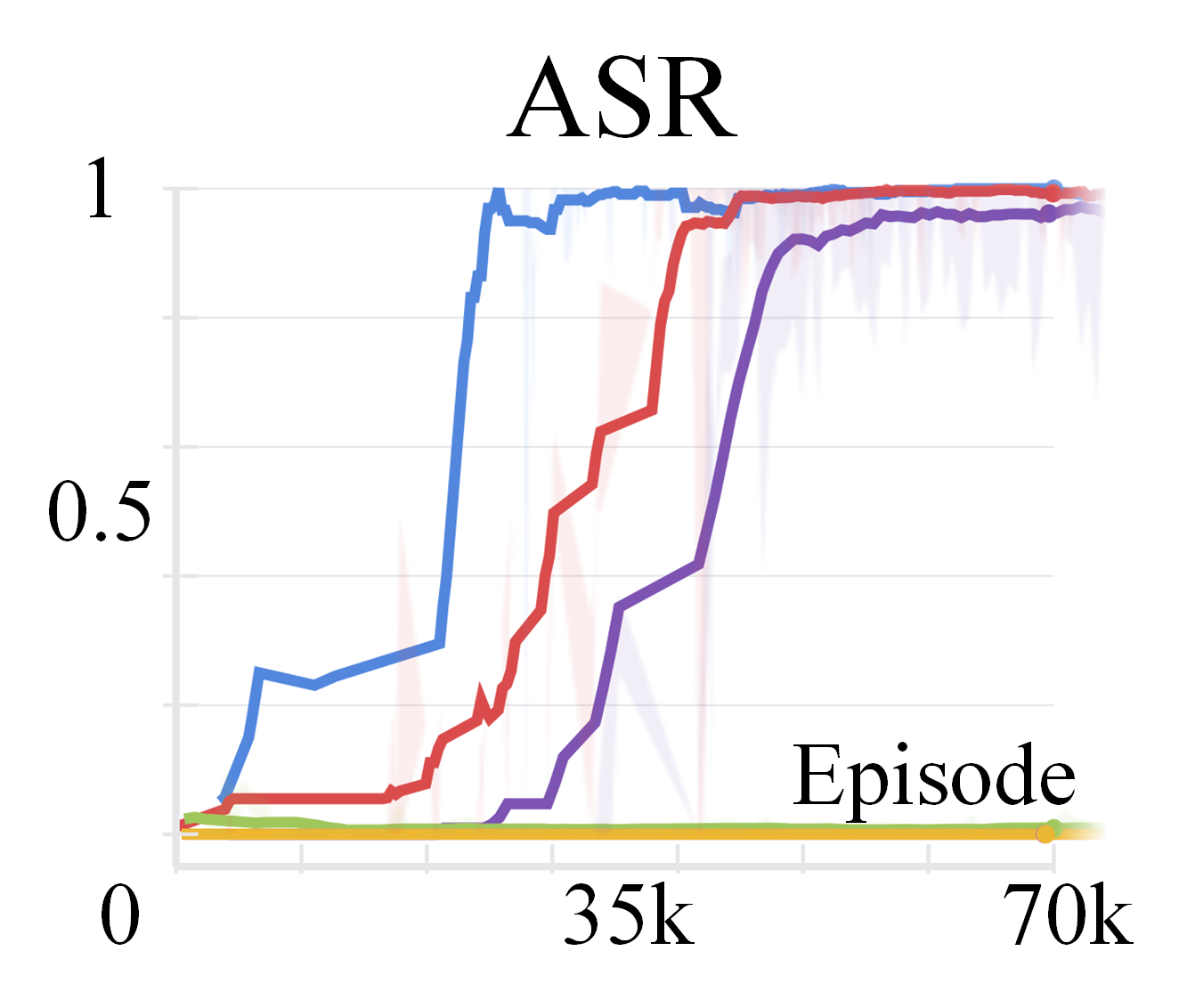}
            \hspace{-0.16cm}
            \includegraphics[width=0.16\textwidth]{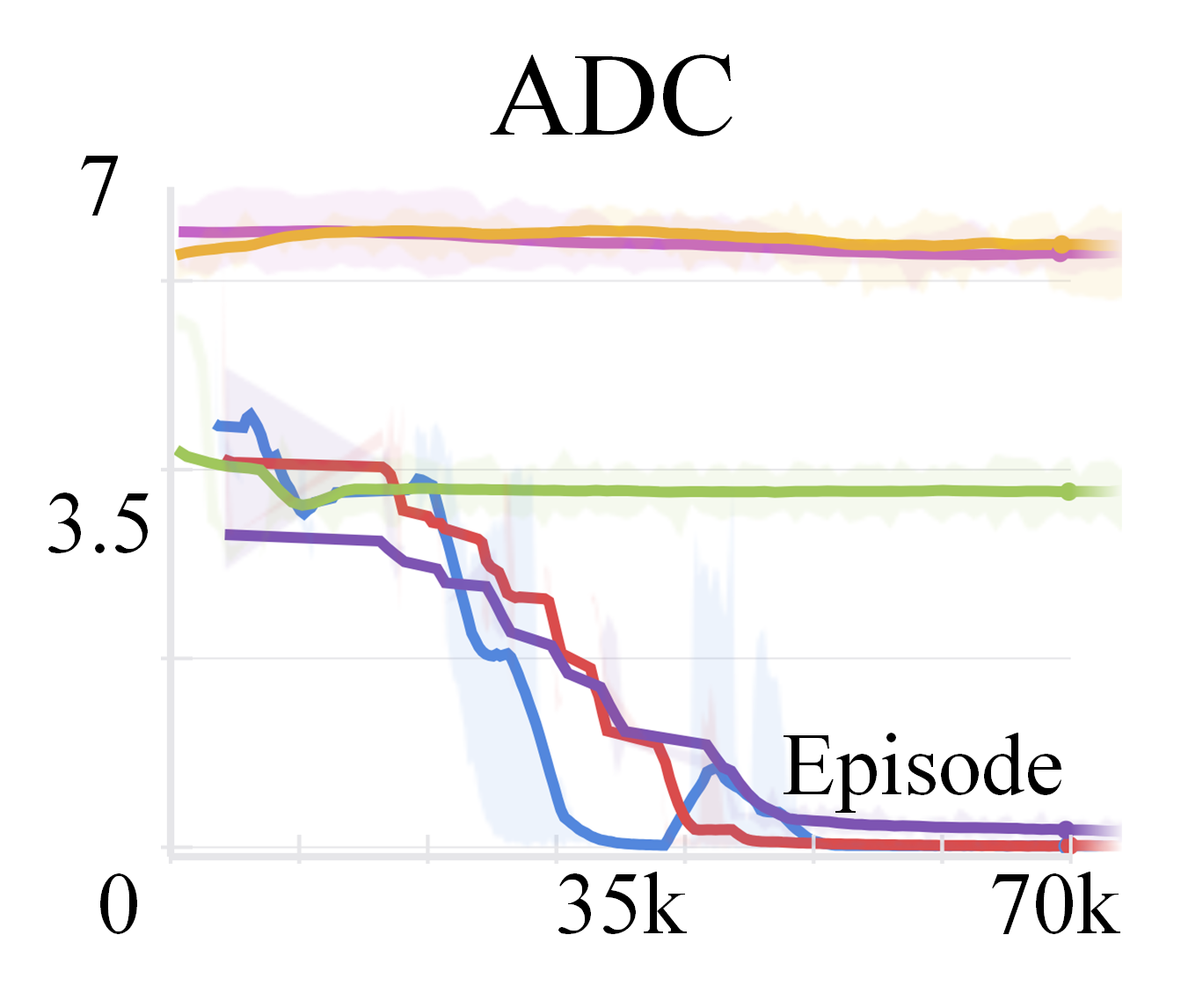}} &
        \subcaptionbox{T3 $(\tau_{\max}{=}4, \sigma_{delay}{=}0.4)$}{
            \includegraphics[width=0.16\textwidth]{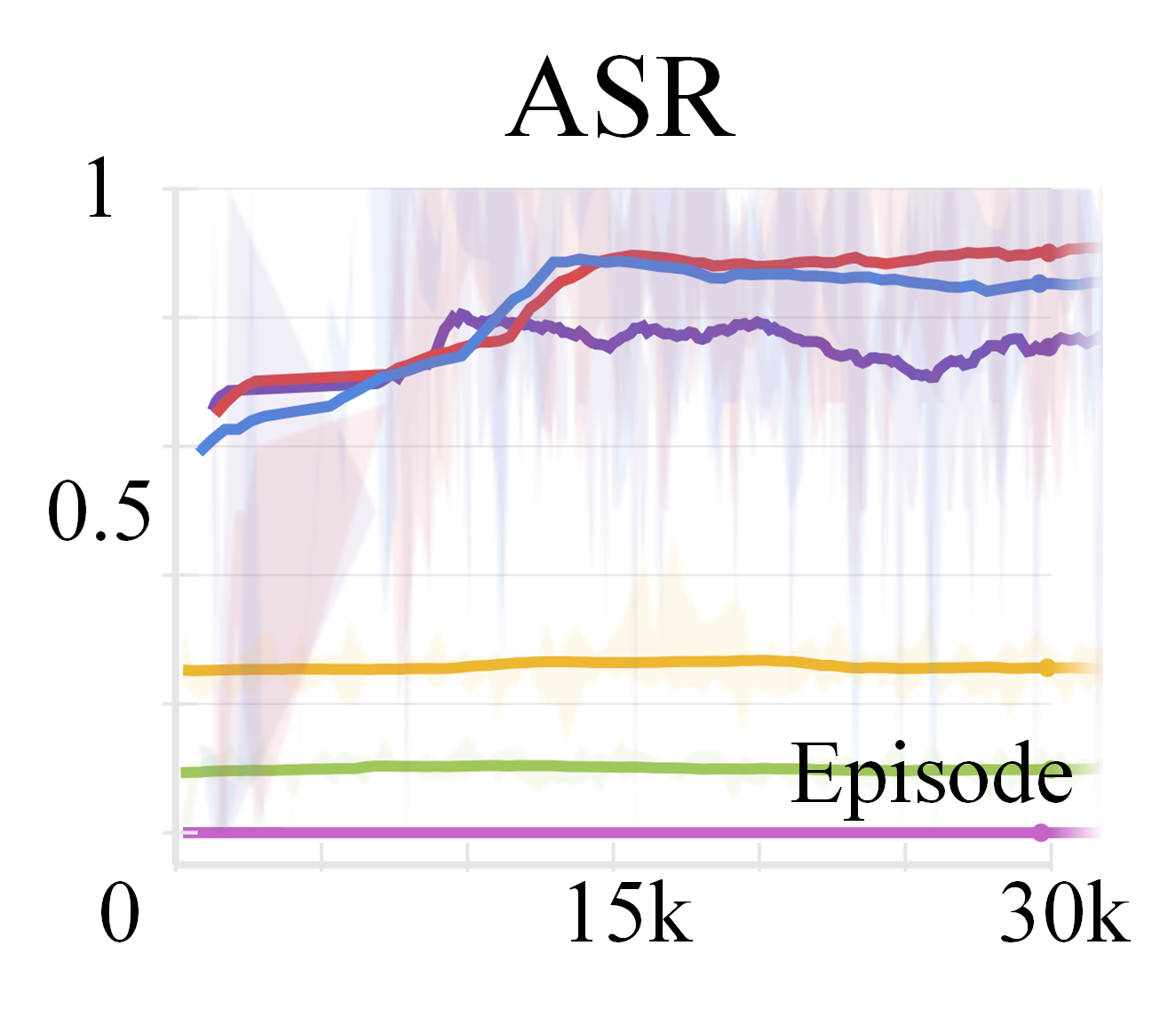}
            \hspace{-0.16cm}
            \includegraphics[width=0.16\textwidth]{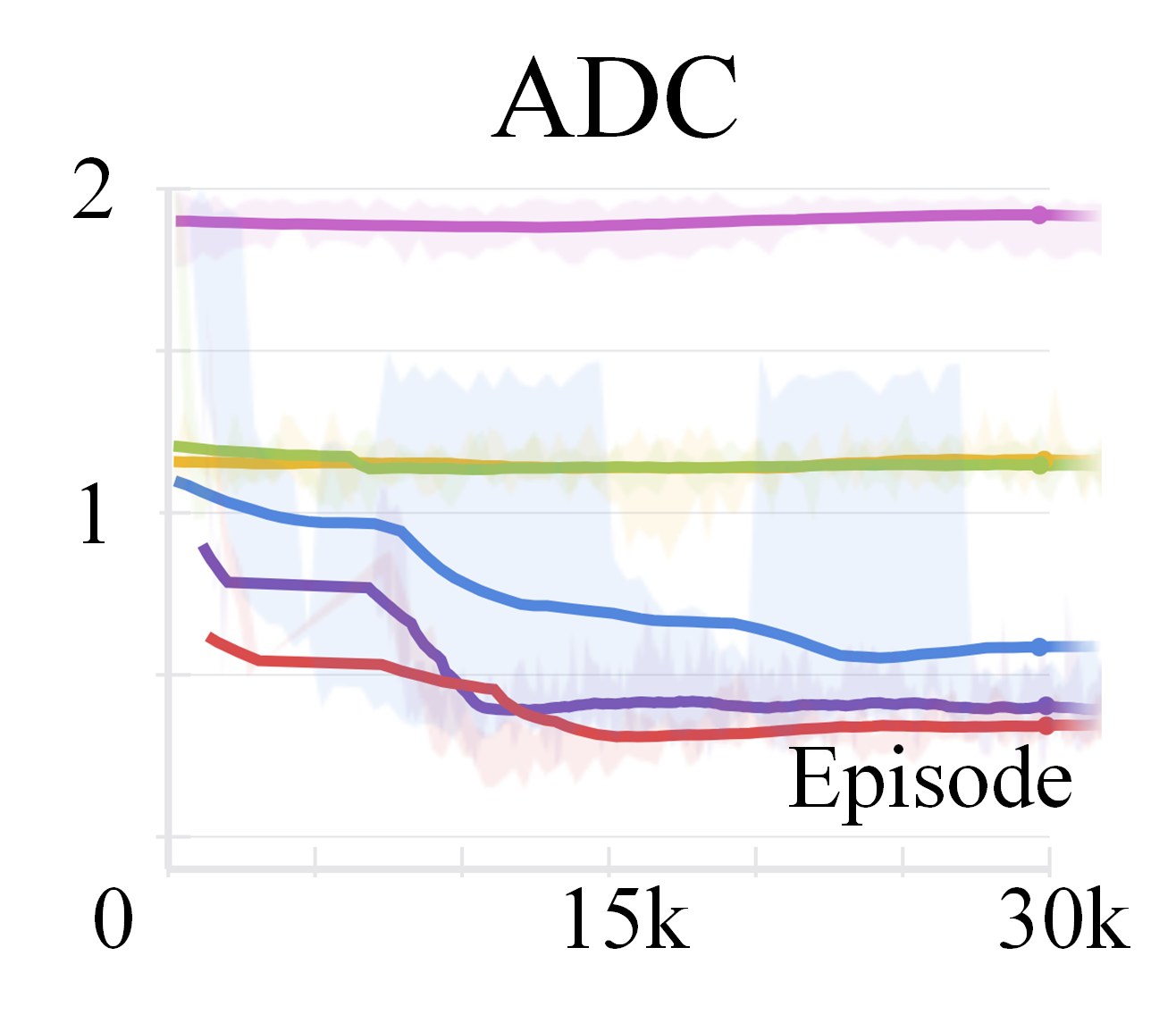}} \\
            
        \subcaptionbox{T1 $(\tau_{\max}{=}4, \sigma_{delay}{=}0.8)$}{
            \includegraphics[width=0.16\textwidth]{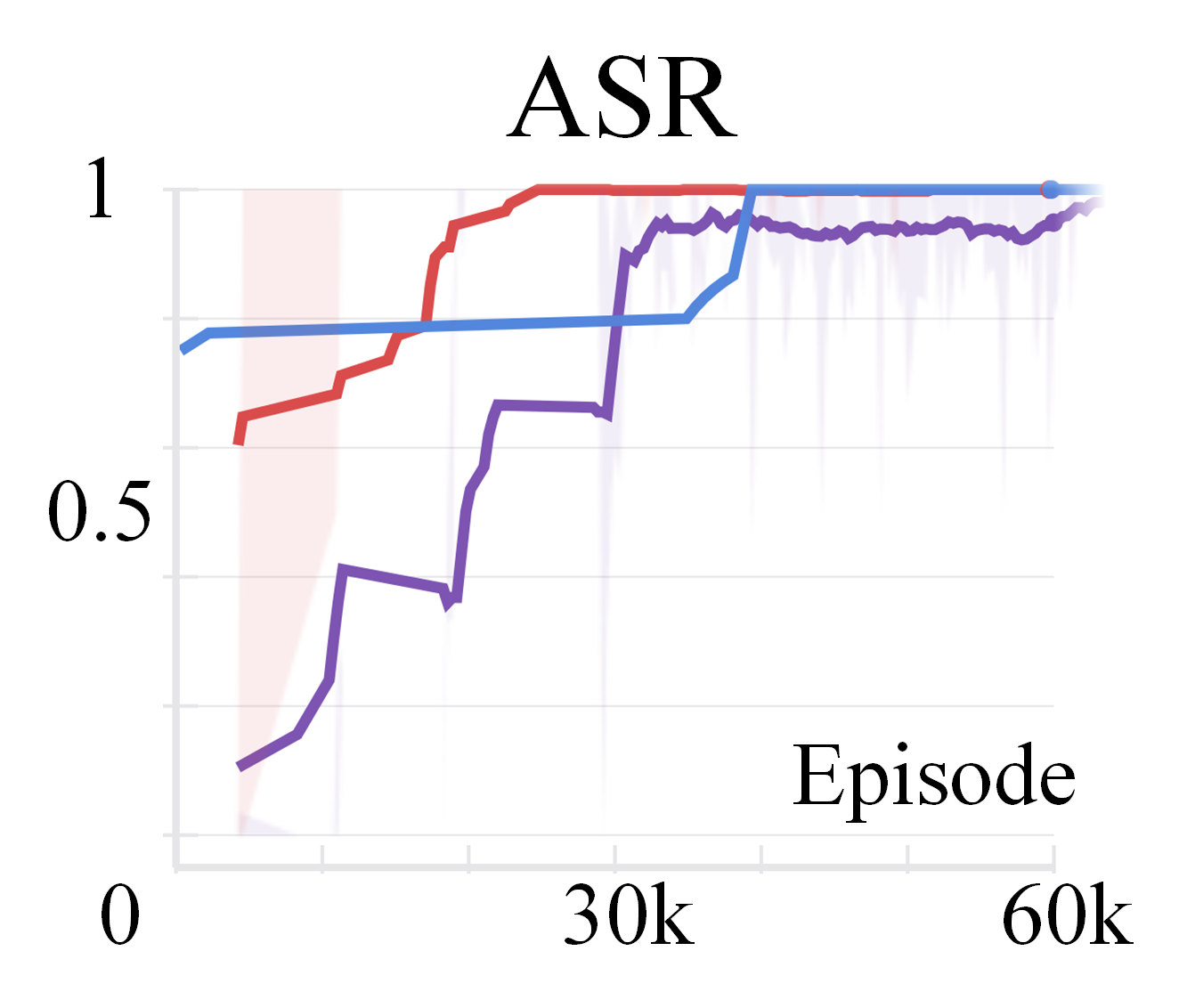}
            \hspace{-0.16cm}
            \includegraphics[width=0.16\textwidth]{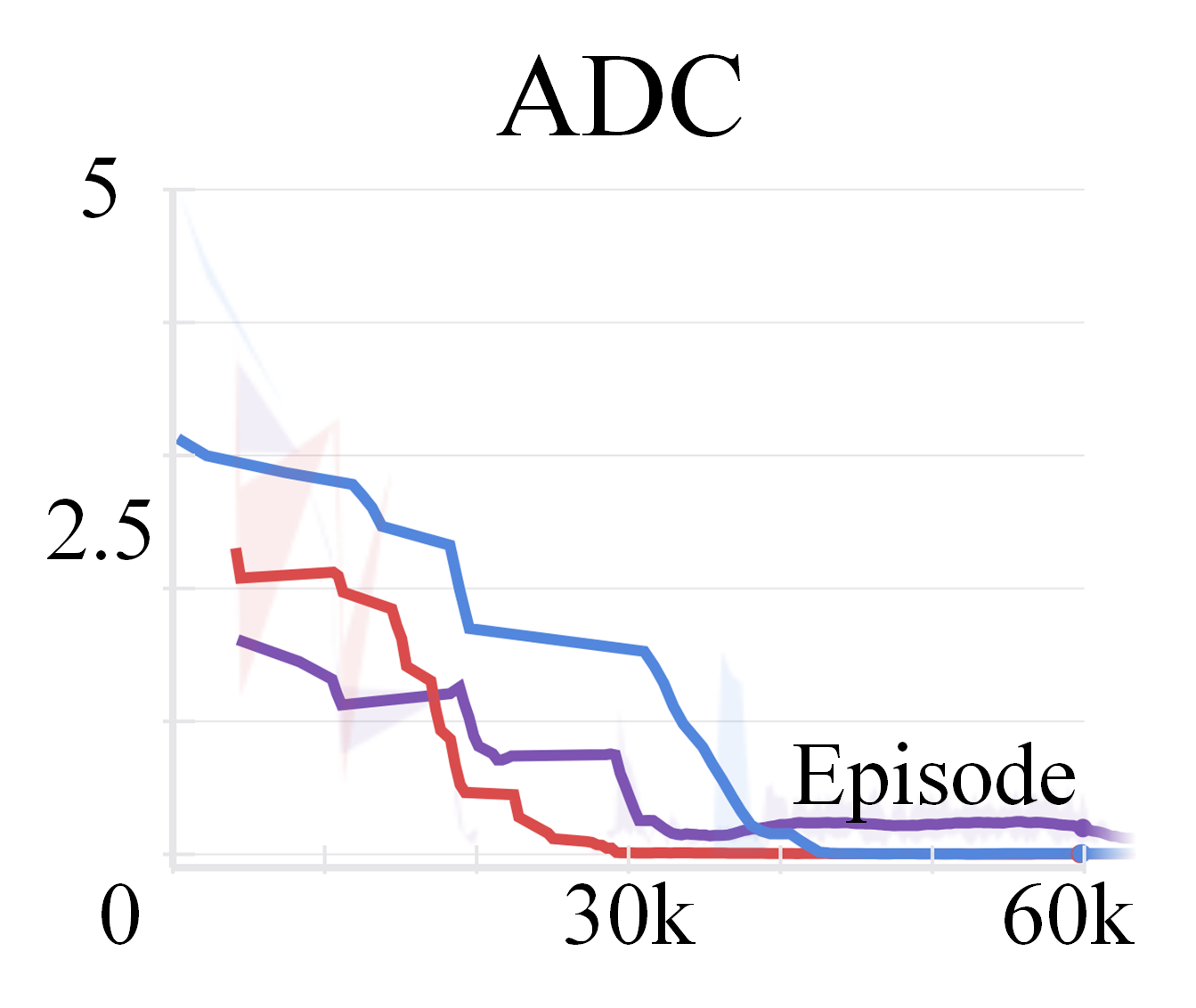}} &
        \subcaptionbox{T2 $(\tau_{\max}{=}4, \sigma_{delay}{=}0.8)$}{
            \includegraphics[width=0.16\textwidth]{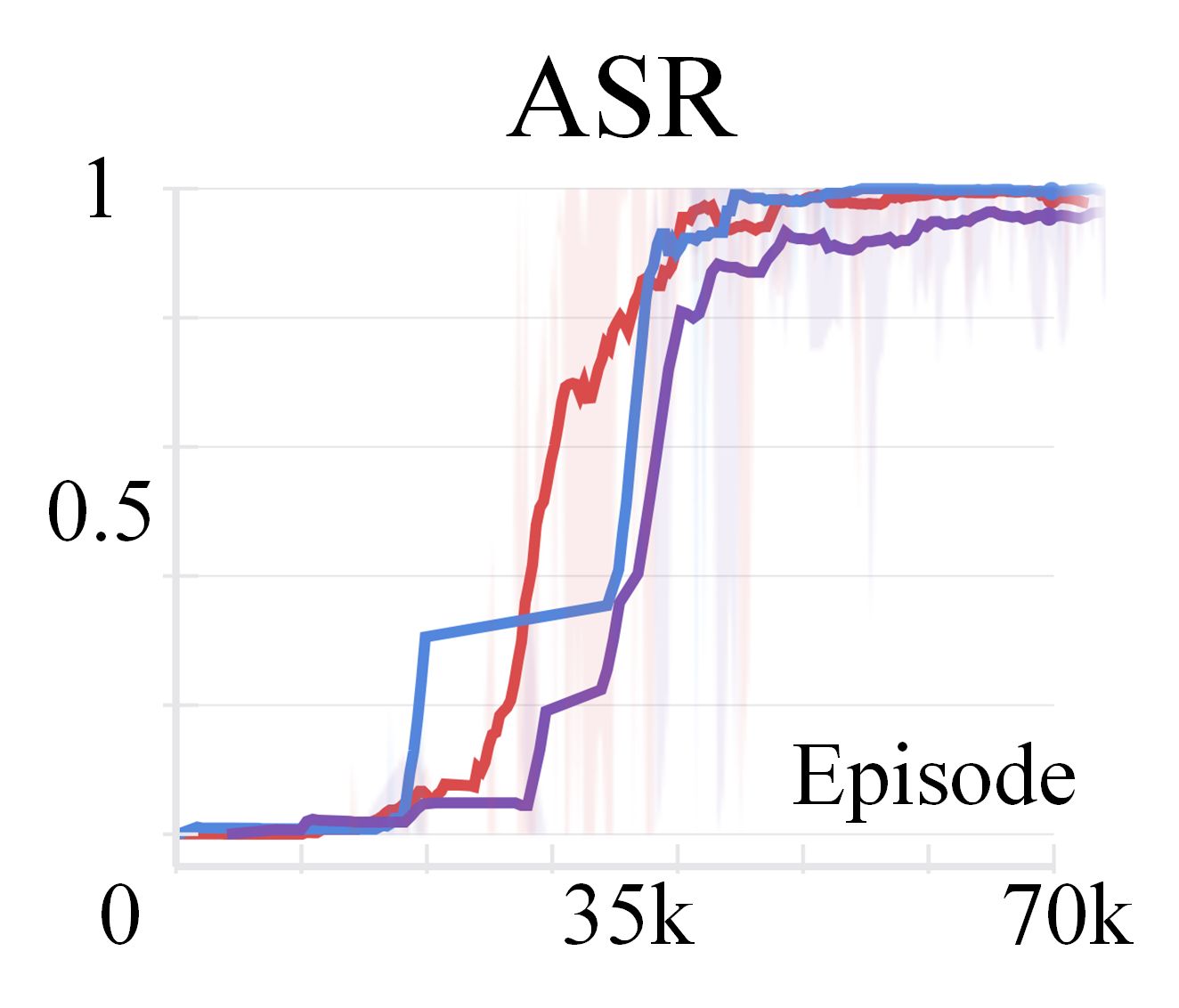}
            \hspace{-0.16cm}
            \includegraphics[width=0.16\textwidth]{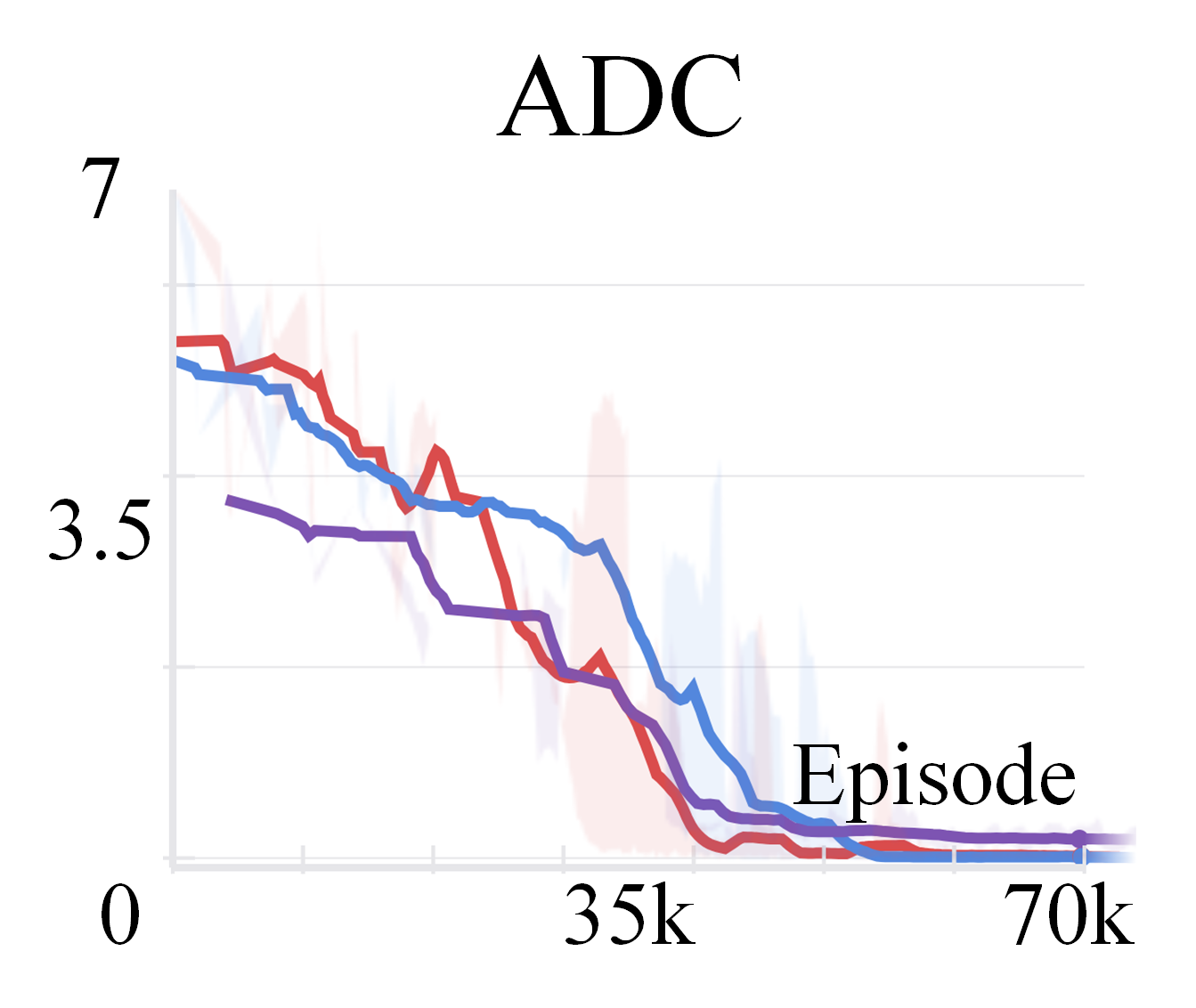}}&
        \subcaptionbox{T3 $(\tau_{\max}{=}8, \sigma_{delay}{=}0.4)$}{
            \includegraphics[width=0.16\textwidth]{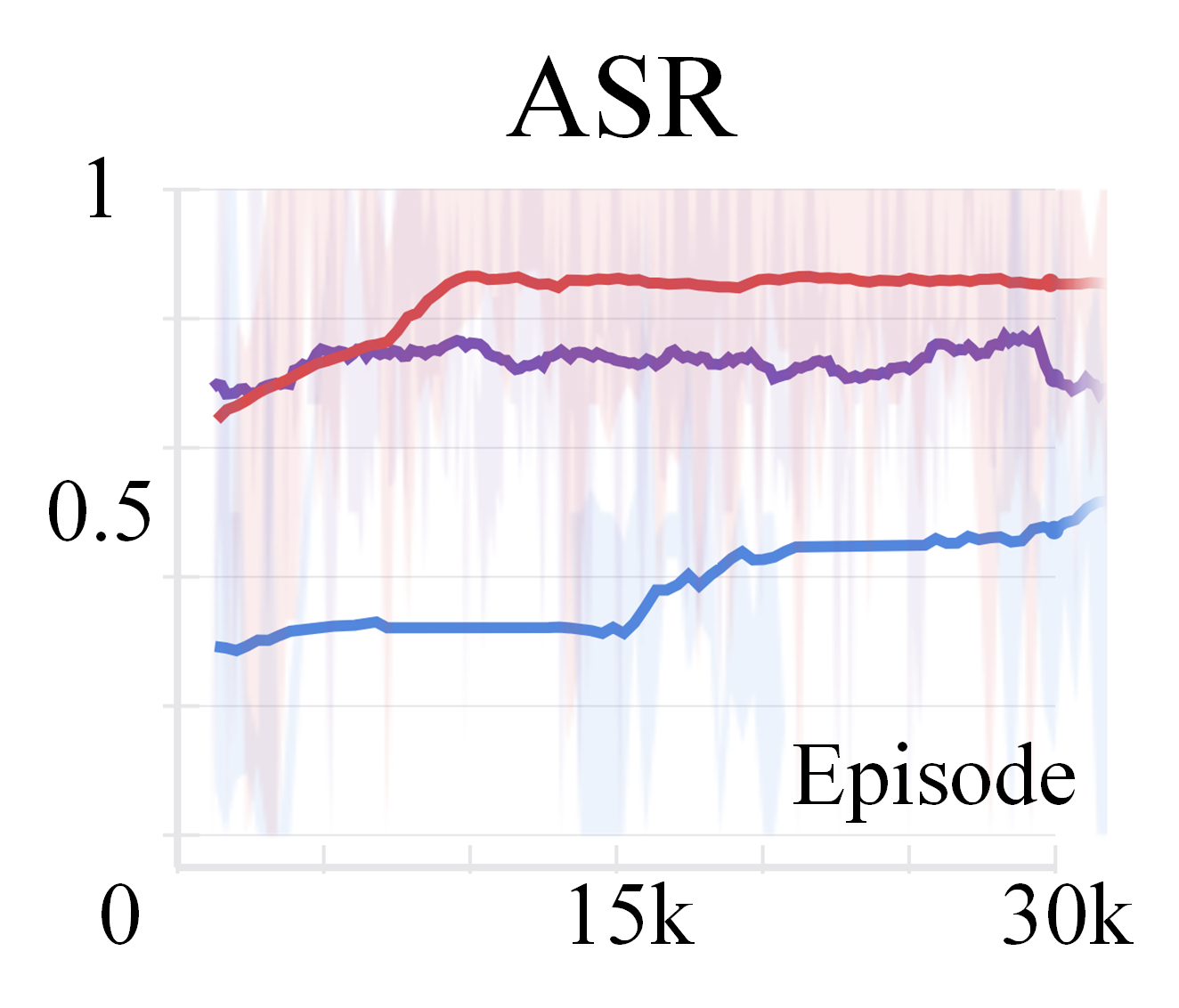}
            \hspace{-0.16cm}
            \includegraphics[width=0.16\textwidth]{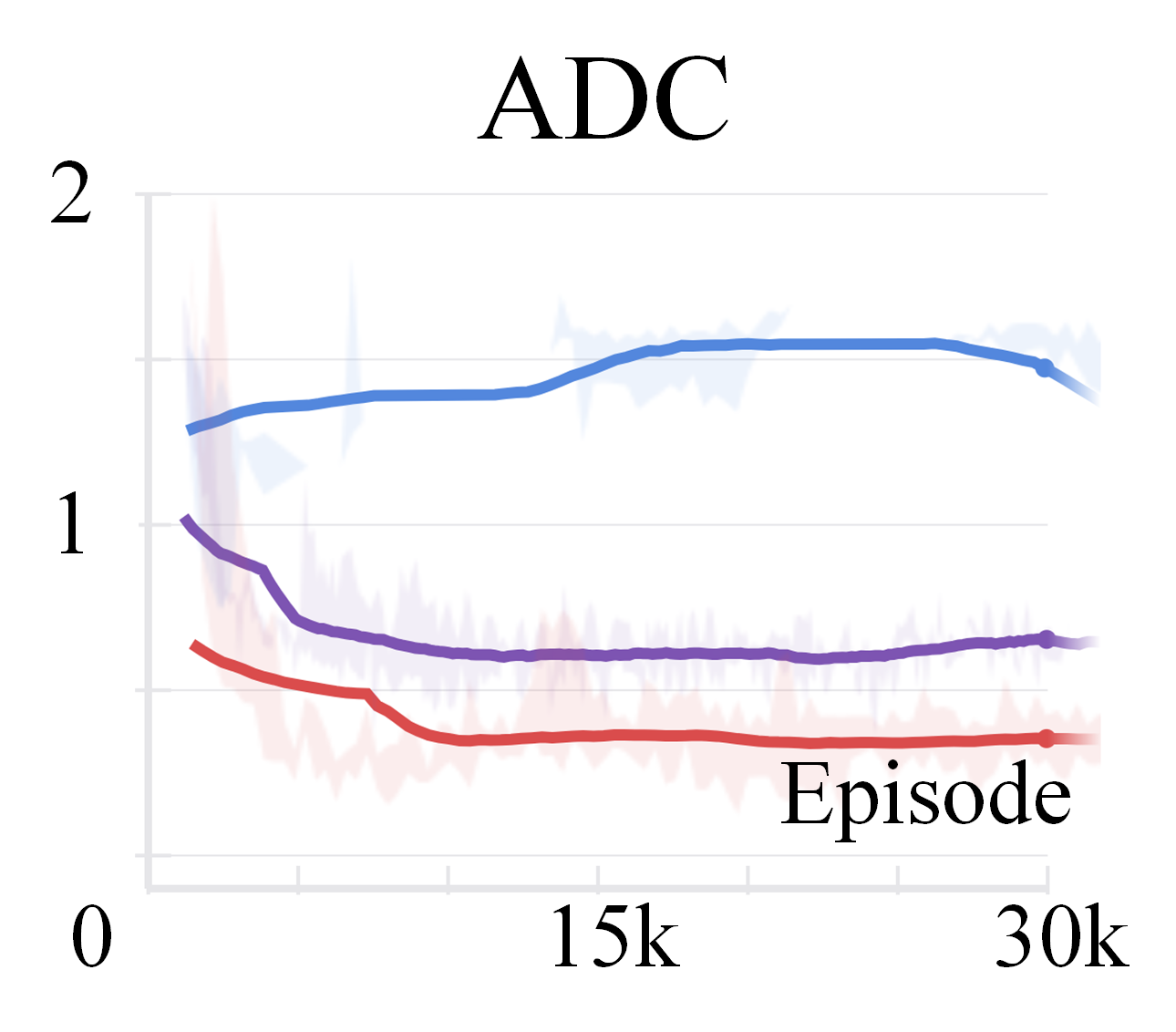}}\\
            
        \subcaptionbox{T1 $(\tau_{\max}{=}8, \sigma_{delay}{=}0.4)$}{
            \includegraphics[width=0.16\textwidth]{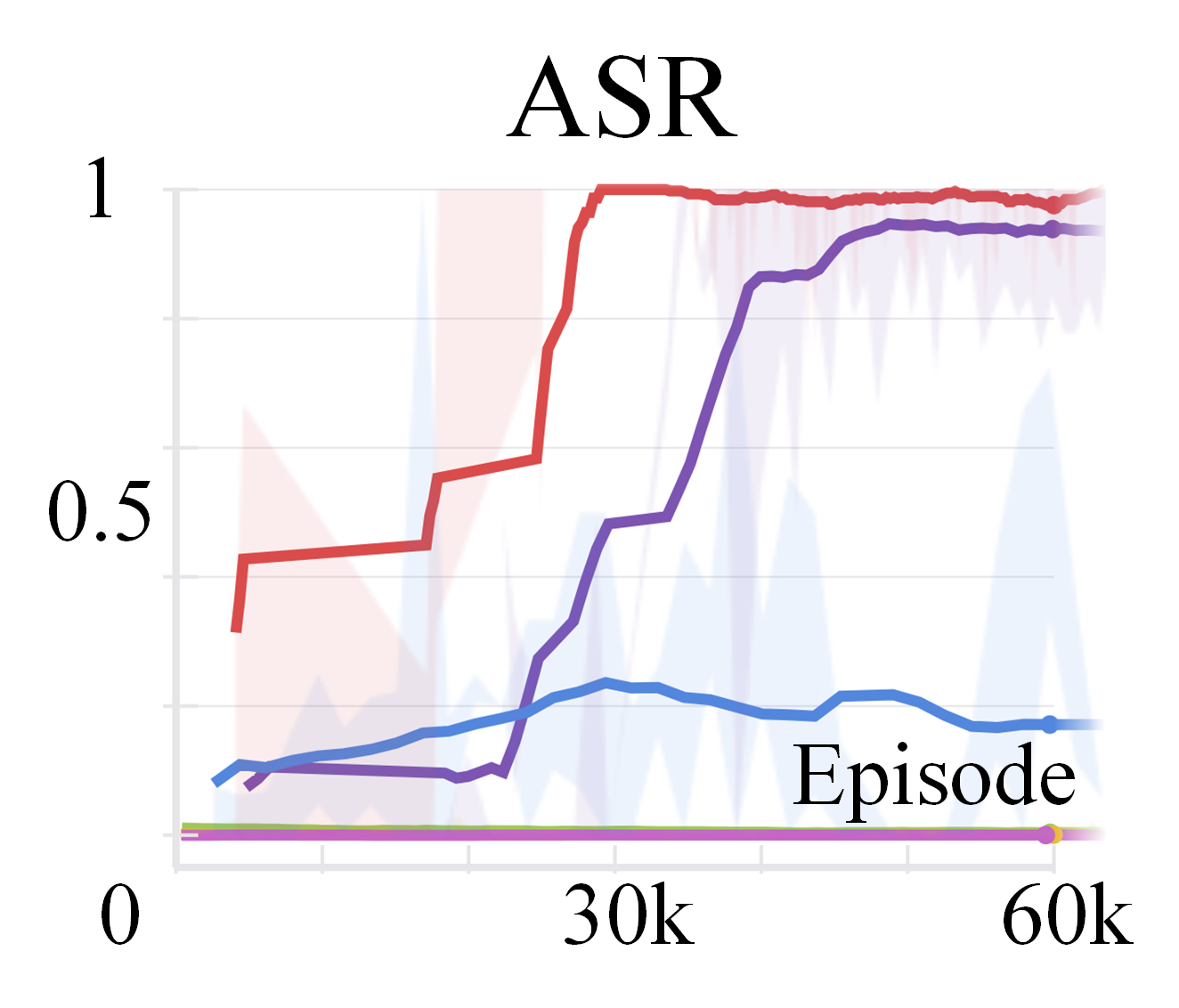}
            \hspace{-0.16cm}
            \includegraphics[width=0.16\textwidth]{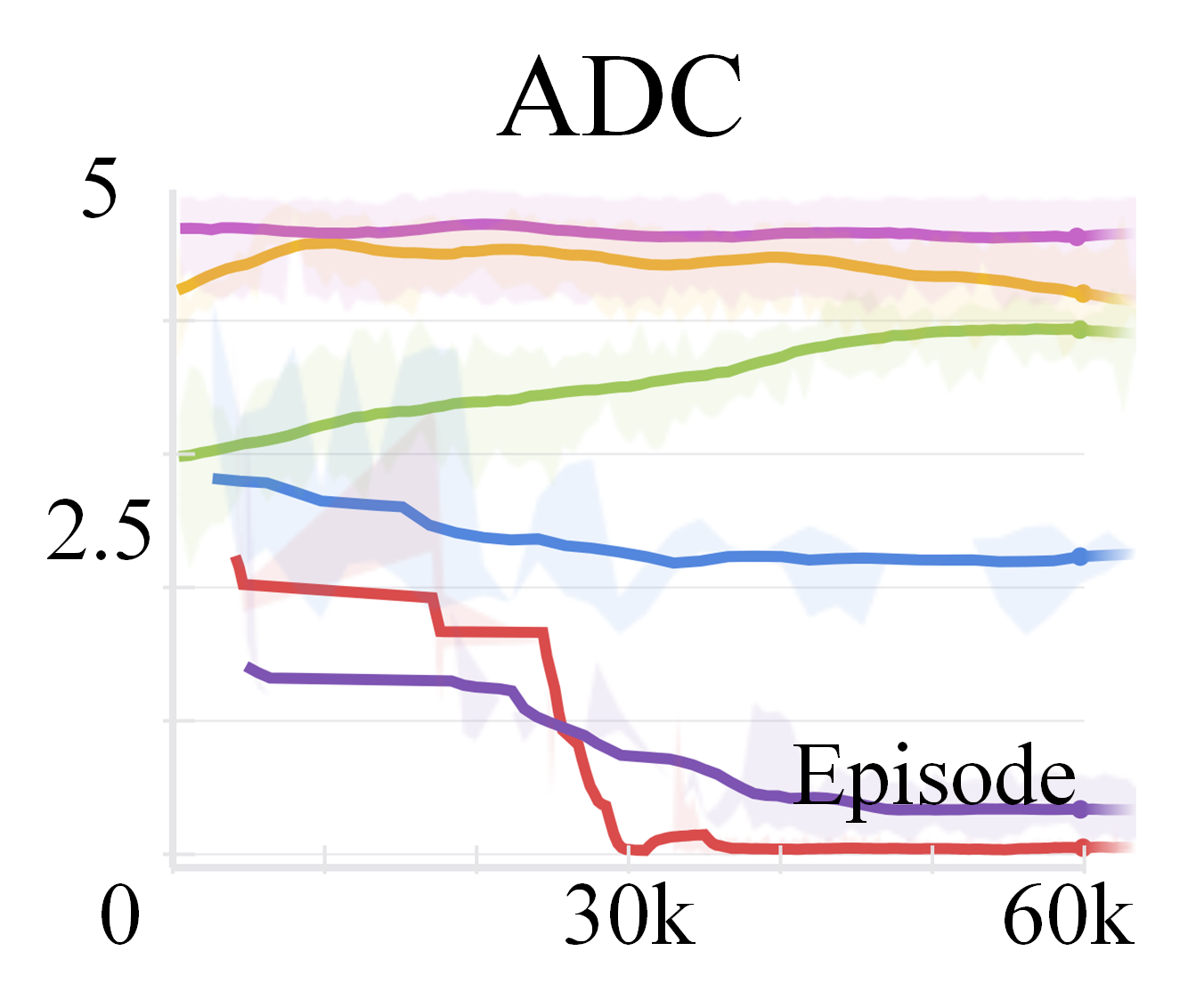}} &
        \subcaptionbox{T2 $(\tau_{\max}{=}8, \sigma_{delay}{=}0.4)$}{
            \includegraphics[width=0.16\textwidth]{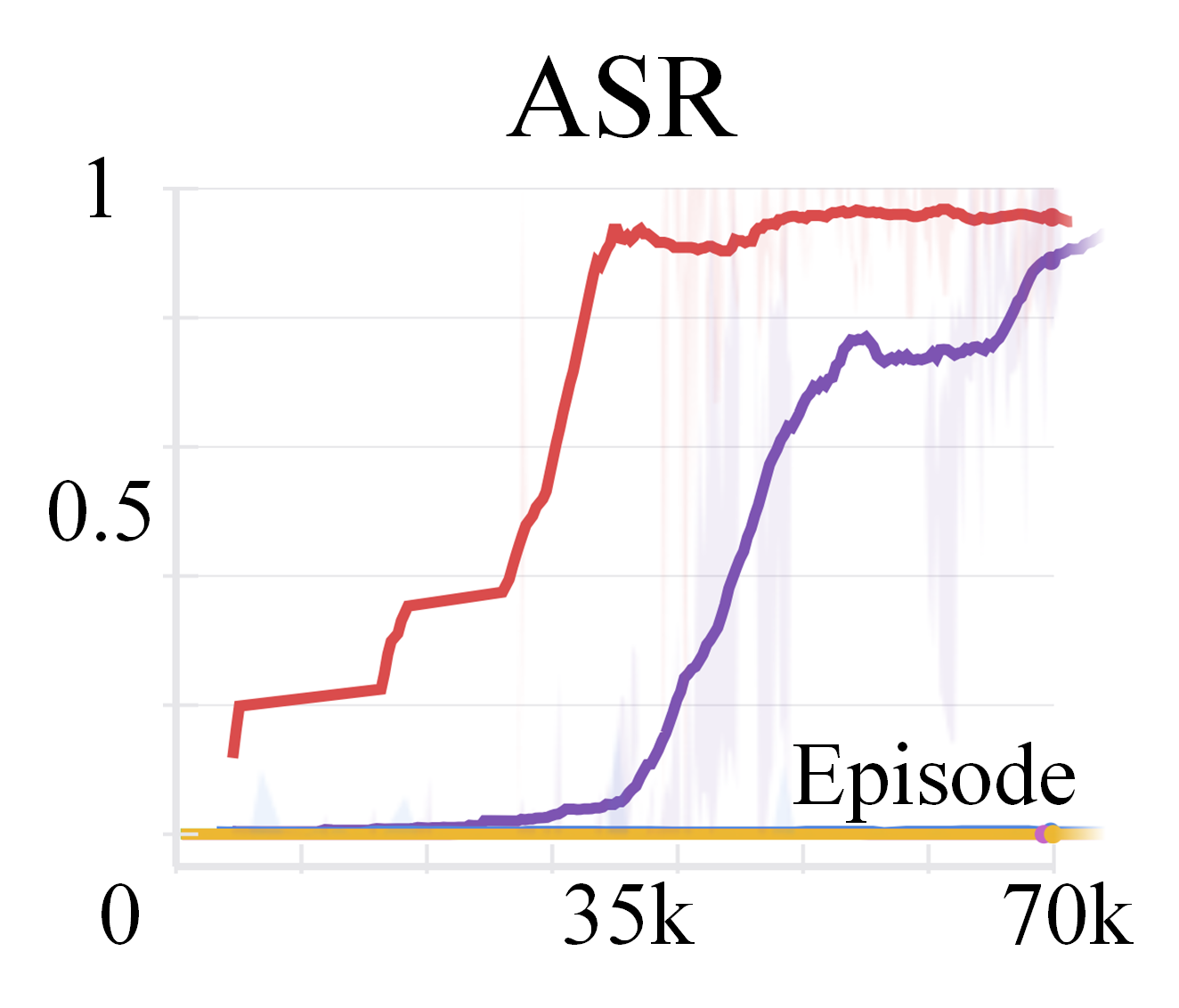}
            \hspace{-0.16cm}
            \includegraphics[width=0.16\textwidth]{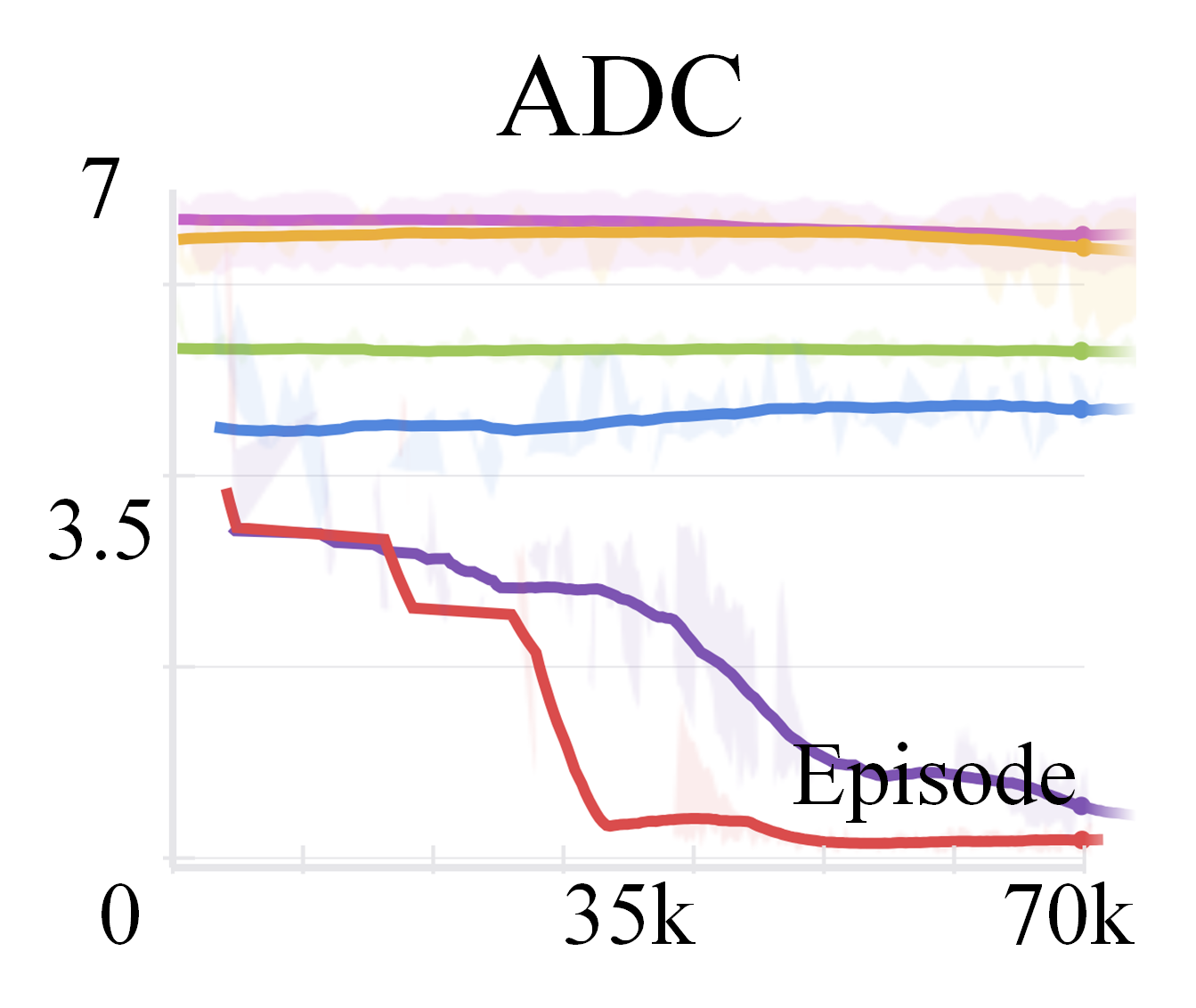}}&
        \subcaptionbox{T4 $(\tau_{\max}{=}4, \sigma_{delay}{=}0.4)$}{
            \includegraphics[width=0.16\textwidth]{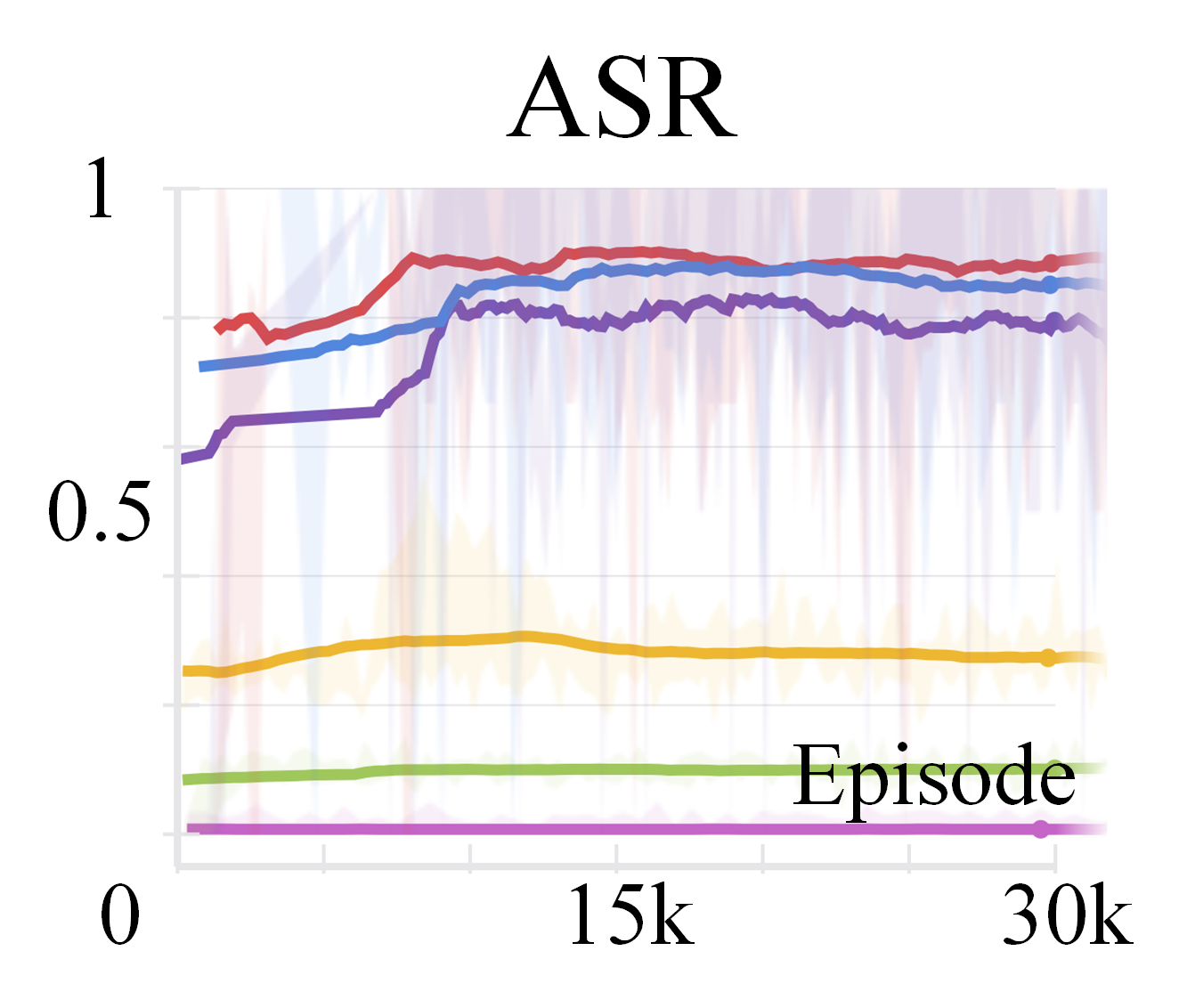}
            \hspace{-0.16cm}
            \includegraphics[width=0.16\textwidth]{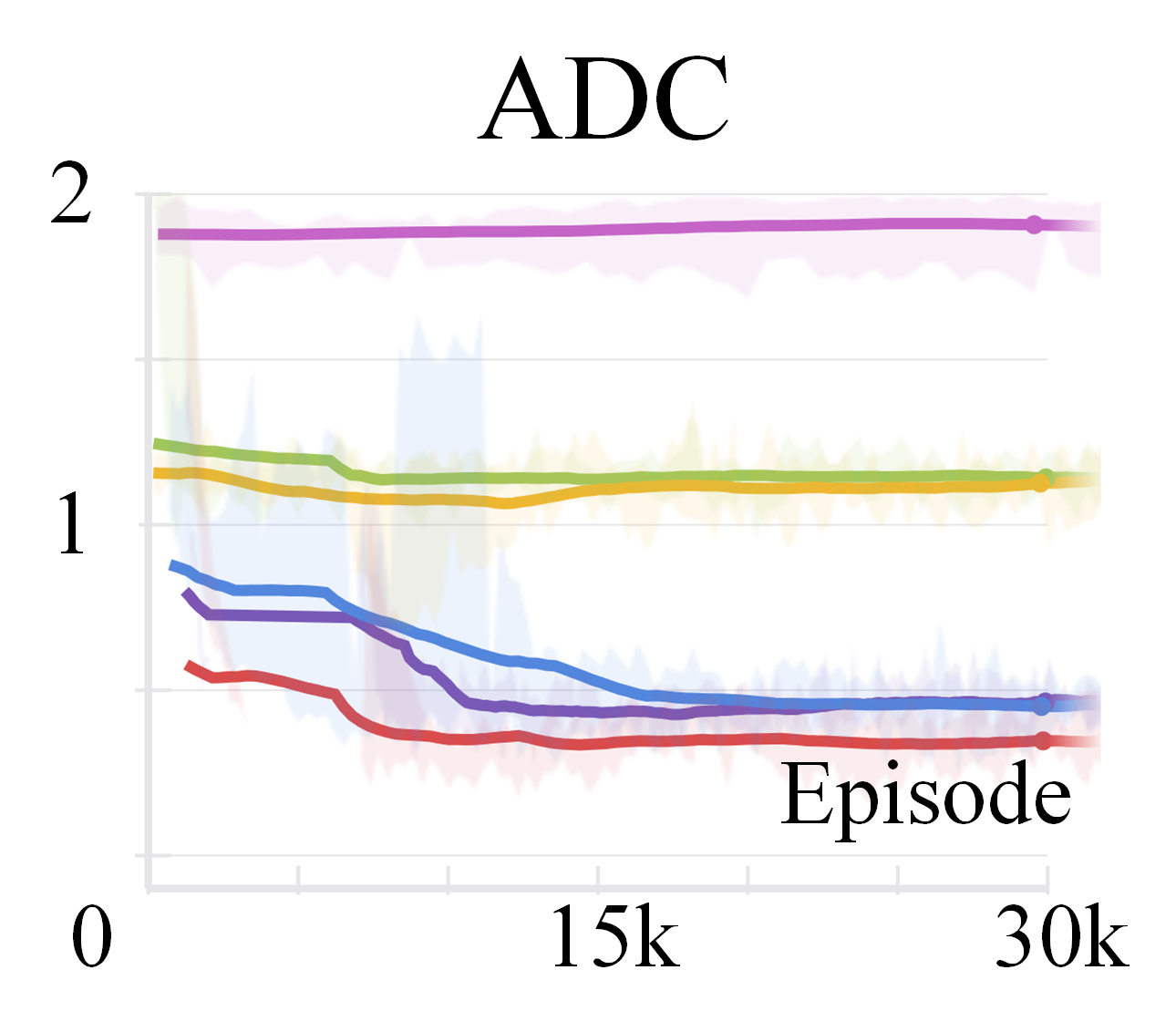}}\\

        \subcaptionbox{T1 $(\tau_{\max}{=}8, \sigma_{delay}{=}0.8)$}{
            \includegraphics[width=0.16\textwidth]{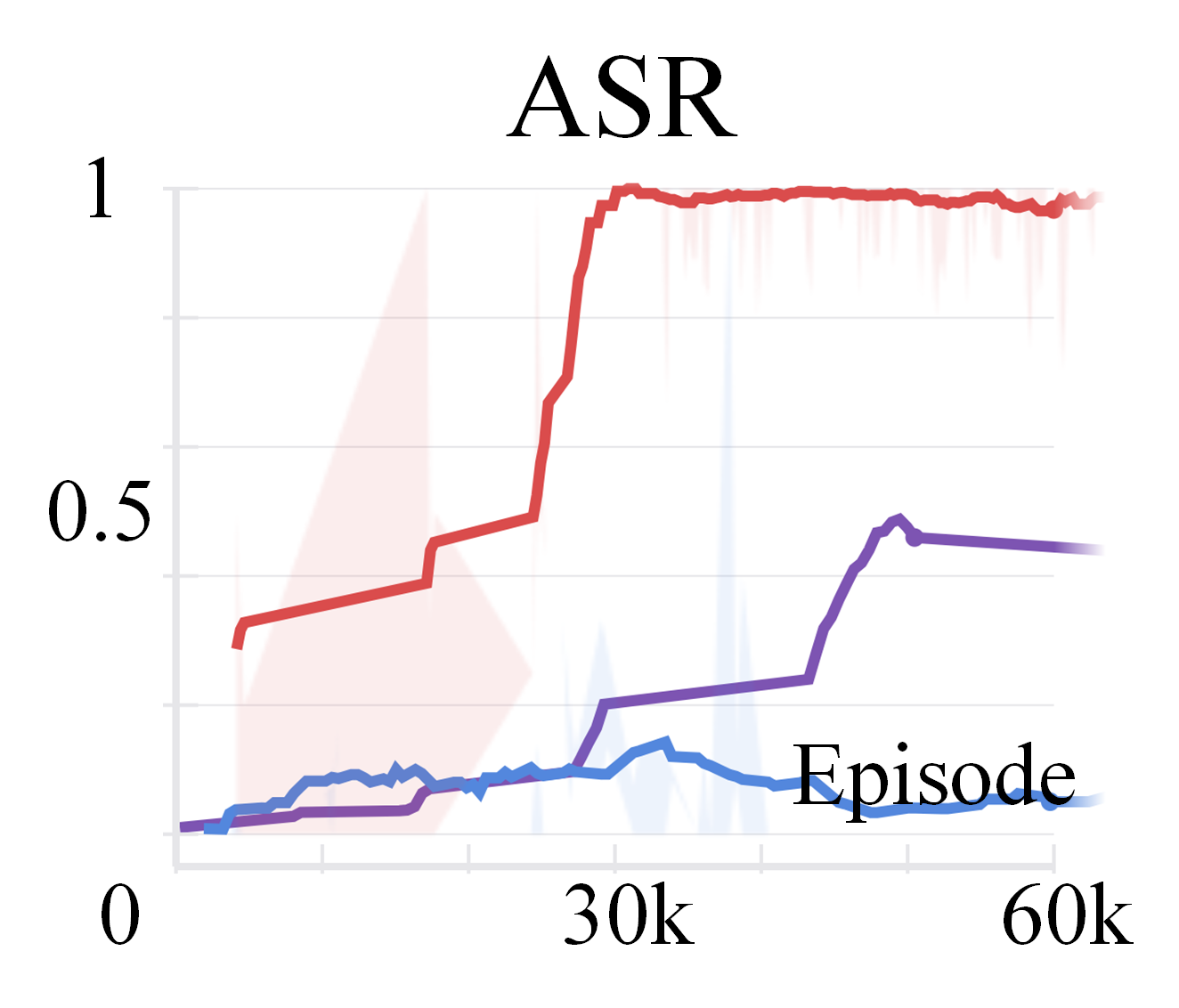}
            \hspace{-0.16cm}
            \includegraphics[width=0.16\textwidth]{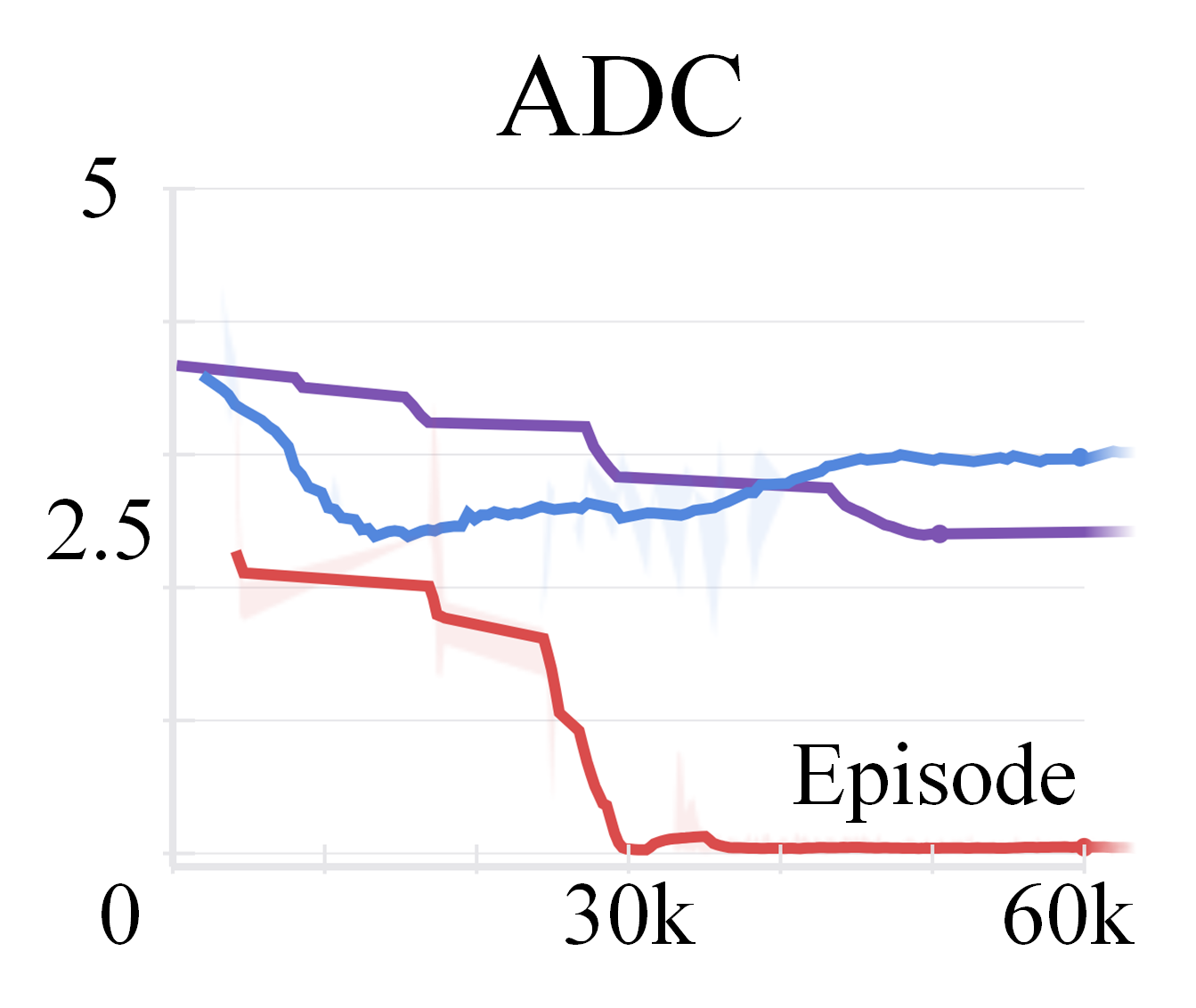}} &
        \subcaptionbox{T2 $(\tau_{\max}{=}8, \sigma_{delay}{=}0.8)$}{
            \includegraphics[width=0.16\textwidth]{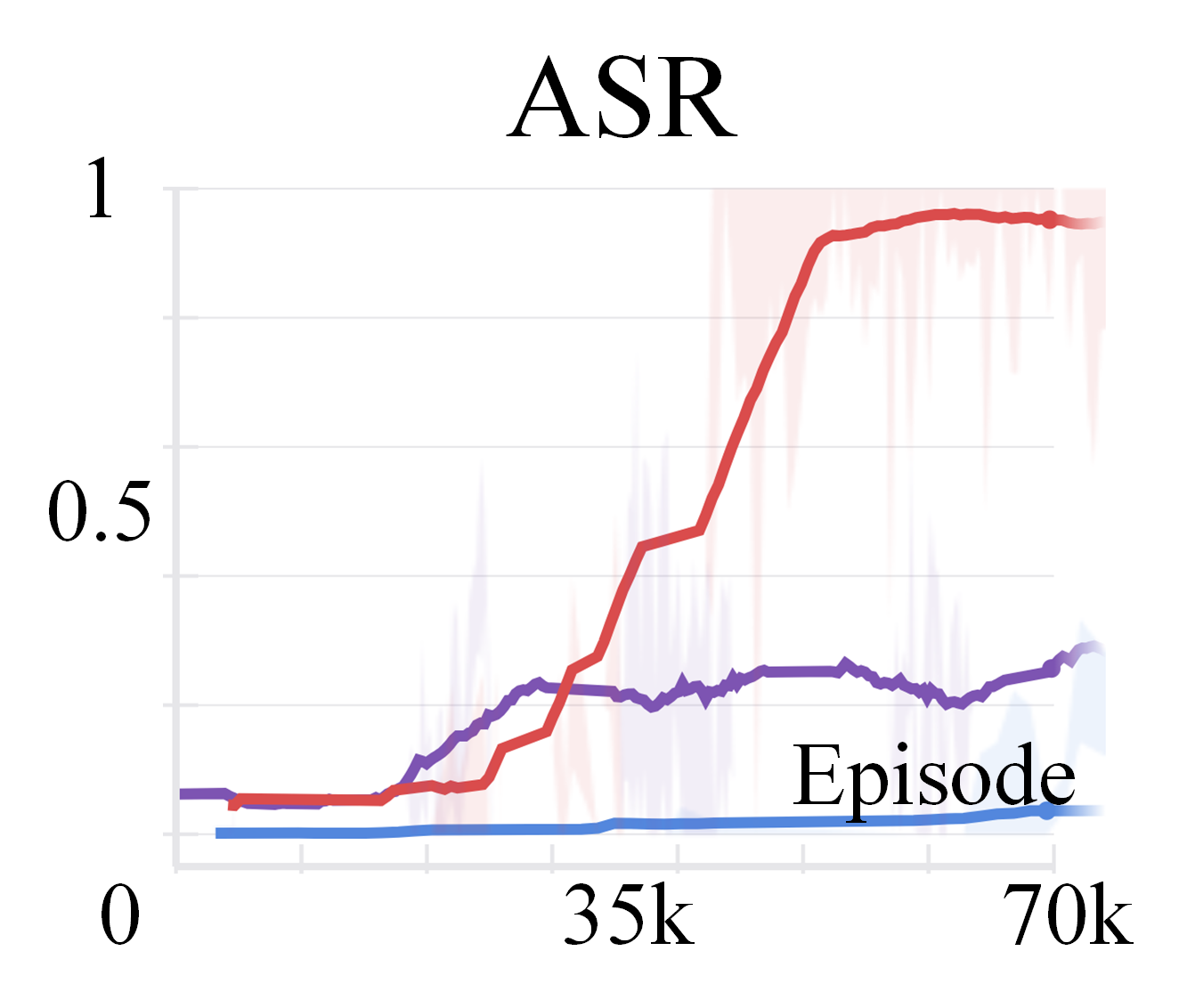}
            \hspace{-0.16cm}
            \includegraphics[width=0.16\textwidth]{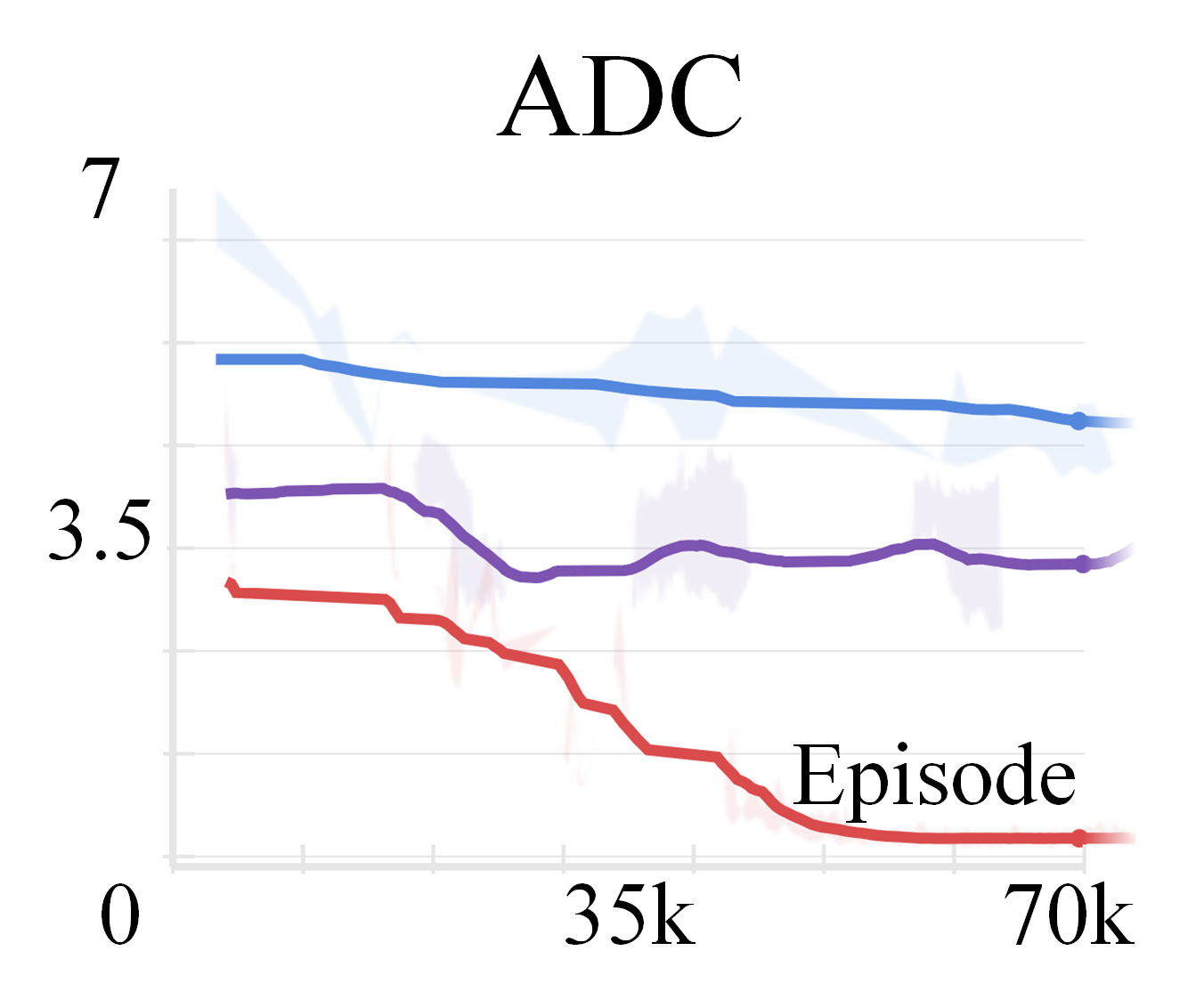}} &
        \subcaptionbox{T4 $(\tau_{\max}{=}8, \sigma_{delay}{=}0.4)$}{
            \includegraphics[width=0.16\textwidth]{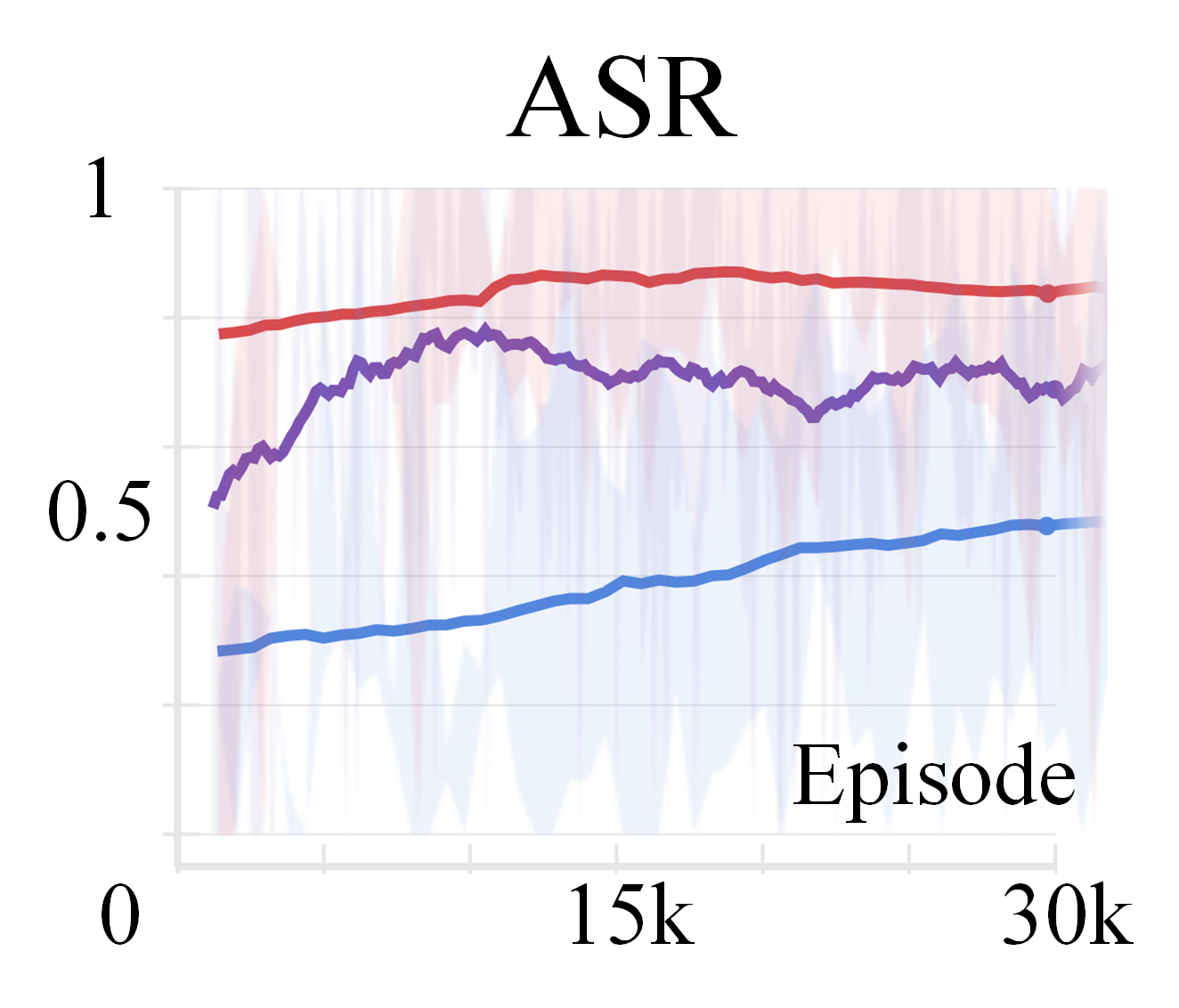}
            \hspace{-0.16cm}
            \includegraphics[width=0.16\textwidth]{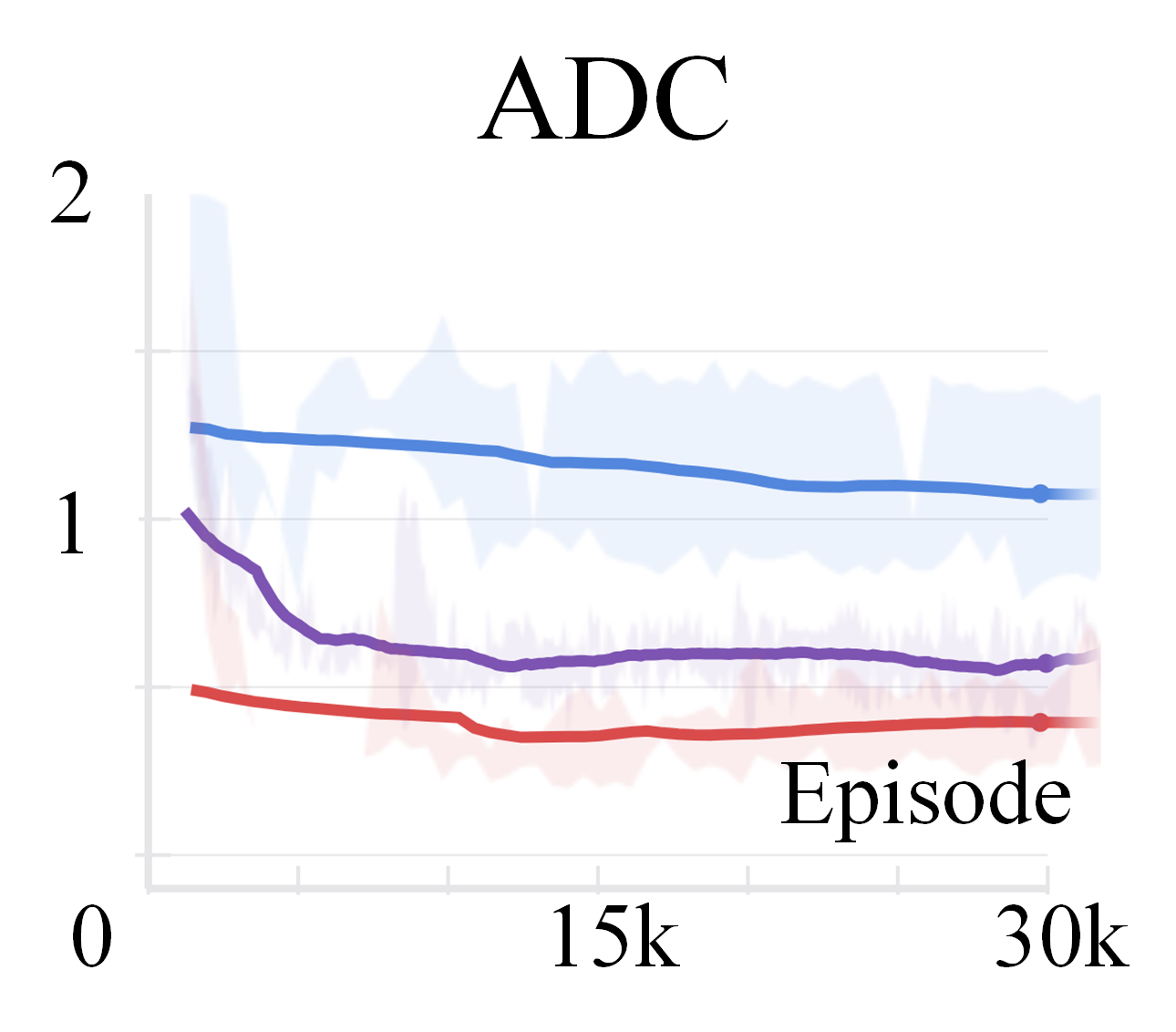}}
    \end{tabular}
    \caption{The ASR and ADC of DECHRL and enhanced HRL baselines across four tasks under varying settings.}
    \label{fig:genernal_exp}
\end{figure}

\indent\textbf{\textit{Experimental Design.}} 
We evaluate the overall performance of \textit{DECHRL} and the enhanced HRL baselines (prefixed ``\textit{Enh.}'') across all tasks under $\tau_{\max} \in \{4, 8\}$ and $\sigma_{delay} \in \{0.4, 0.8\}$, with the results summarized in Figure~\ref{fig:genernal_exp}.

\indent\textbf{\textit{Experimental Results.}} 
Our \textbf{DECHRL} demonstrates strong performance across tasks by effectively modeling delayed effects and enabling proactive, delay-aware empowerment-driven exploration. Although it slightly trails CDHRL on the \textit{GetIron} task ($\tau_{\max}=4, \sigma_{delay}=0.4$) in Figure~\ref{fig:genernal_exp}(b), it outperforms baselines as delay stochasticity $\sigma_{delay}$ and maximum delay $\tau_{\max}$ increase, highlighting its robustness in handling complex, uncertain delay dynamics.
\textbf{D3HRL} relies on accurate delay estimation. Under low $\sigma_{delay}$, delays are more predictable, allowing its fixed-delay model to perform well. However, as $\sigma_{delay}$ increases, it becomes harder to align transitions with true delays, leading to reduced performance and weaker robustness.
\textbf{Enh.CDHRL} can perceive fixed delays, but exhibits limited sensitivity to the maximum delay $\tau_{\max}$: performance is strong at $\tau_{\max} = 4$ but degrades for $\tau_{\max} = 8$. This limitation arises because it cannot accurately model variable delay distributions—larger $\tau_{\max}$ increases perceptual bias, hindering hierarchical policy learning.
\textbf{Enh.HAC} incorporates curriculum learning and causal supervision, but does not support stochastic delay modeling or intrinsic motivation. Consequently, it performs adequately on simpler tasks like \textit{Fire2Burn} (T3) and \textit{Wood2Wet} (T4) but deteriorates on more complex ones such as \textit{GetSilverore} (T1) and \textit{GetIron} (T2).
\textbf{Enh.Option-Critic} and \textbf{Enh.LESSON} lack multi-level hierarchical structures. Despite being guided via curriculum learning, their flat architectures struggle to retain prior knowledge across subgoal levels. Consequently, their performance remains poor across all tasks. Notably, Option-Critic slightly outperforms LESSON due to its inherent option-based temporal abstraction. 
We note that HAC, Option-Critic, and LESSON struggle to match the performance of other methods even under the relatively easy setting ($\sigma_{delay} = 0.4$). Consequently, we omit these algorithms from evaluation under the more challenging condition ($\sigma_{delay} = 0.8$).
\textit{Among the various baselines, CDHRL and D3HRL are the two baselines most closely related to our work. Accordingly, we provide a detailed comparison of CDHRL, D3HRL, and our proposed DECHRL in \ref{app:comparison_chrl_complexity}, with a focus on implementation details and computational complexity.}

\subsubsection{How closely does the learned delay distribution match the ground-truth delay distribution?}

\indent\textbf{\textit{Experimental Design.}} 
We evaluate DECHRL on the \textit{GetSilverore} task under varying degrees of delay stochasticity $\sigma_{delay}=\{0.4, 0.6, 0.8, 1.0\}$, with a maximum delay of $\tau_{\max} = 4$. Specifically, we compute the KL divergence between the delay distributions learned by DECHRL for each state transition and their corresponding ground-truth distributions. The results reported are averaged over 5 independent runs and presented in Table~\ref{tab:kl_distrbution}.

\begin{table*}[htbp]
\centering
\resizebox{\textwidth}{!}{
\begin{tabular}{c|ccccc}
\toprule
\diagbox[width=4cm, height=1.1cm]{$\sigma_{delay}$}{Transitions}
& \makecell{$\mathbf{A}$\\$\shortdownarrow$\\$wood$}
& \makecell{$\mathbf{A}$\\$\shortdownarrow$\\$stone$}
& \makecell{$wood$\\$\shortdownarrow$\\$stick$}
& \makecell{$stone, stick$\\$\shortdownarrow$\\$stonepickaxe$}
& \makecell{$stonepickaxe$\\$\shortdownarrow$\\$silverore$}\\

\midrule
$\sigma_{delay}=0.4$ & 0.0354 & 0.0217 & 0.0449 & 0.7876 & 0.0825\\
$\sigma_{delay}=0.6$ & 0.4051 & 0.0963 & 0.5748 & 0.2656 & 0.1293\\
$\sigma_{delay}=0.8$ & 0.3085 & 0.2431 & 0.9108 & 0.6557 & 0.1638\\
$\sigma_{delay}=1.0$ & 0.1697 & 0.4960 & 0.6287 & 0.2961 & 2.0625\\
\bottomrule
\end{tabular}
}
\caption{KL divergence between the true and learned delay distributions on the \textit{GetSilverore} ($\tau_{\max} = 4$) task under varying levels of delay stochasticity $\sigma_{\text{delay}} \in \{0.4, 0.6, 0.8, 1.0\}$.}
\label{tab:kl_distrbution}
\end{table*}

\indent\textbf{\textit{Experimental Results.}} 
For the majority of state transitions, the KL divergence between the delay distributions learned by DECHRL and the true distributions remains below $0.5$, indicating that, despite minor discrepancies, their core structures are well aligned. For a small subset of transitions, however, noticeable deviations are observed. As the stochasticity $\sigma_{delay}$ of state transitions in the environment increases, the discrepancy between the learned and ground-truth delay distributions exhibits a growing trend. Overall, our proposed delay modeling module effectively captures the underlying delay distributions of environmental state transitions. In conjunction with the experimental results presented in Sections~\ref{sec:delay_method_comparison} and \ref{sec:exp_genernal}, this demonstrates that the learned delay distributions enable the agent to perceive temporal delays in the environment and thereby facilitate robust decision-making under delayed feedback.

\subsubsection{How does the choice of gradient estimators method for modeling the delay distribution affect the performance of DECHRL?}\label{sec:gradient_exp}
\begin{figure}[htbp]
    \centering
    \includegraphics[width=0.29\textwidth]{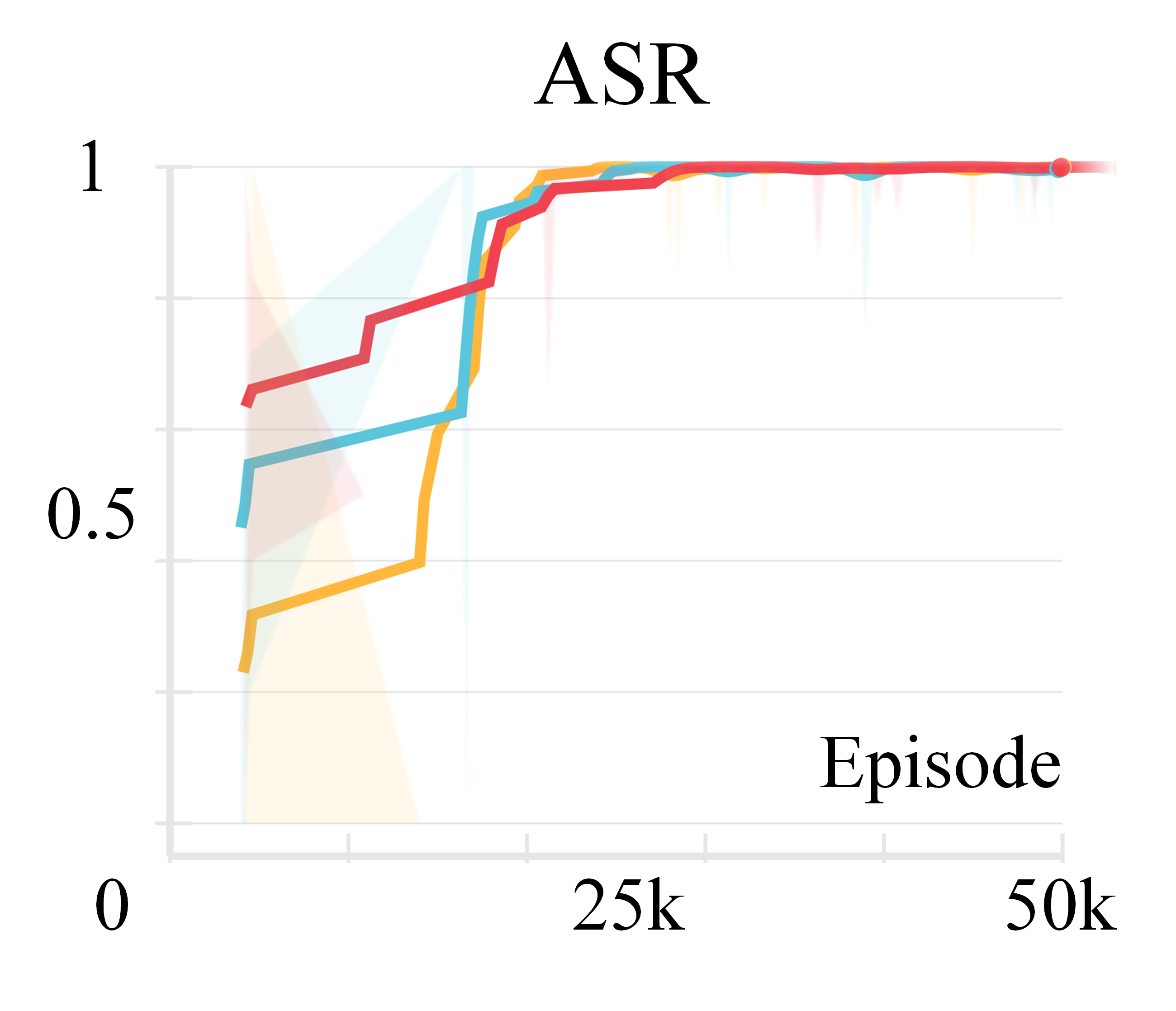}
    \includegraphics[width=0.29\textwidth]{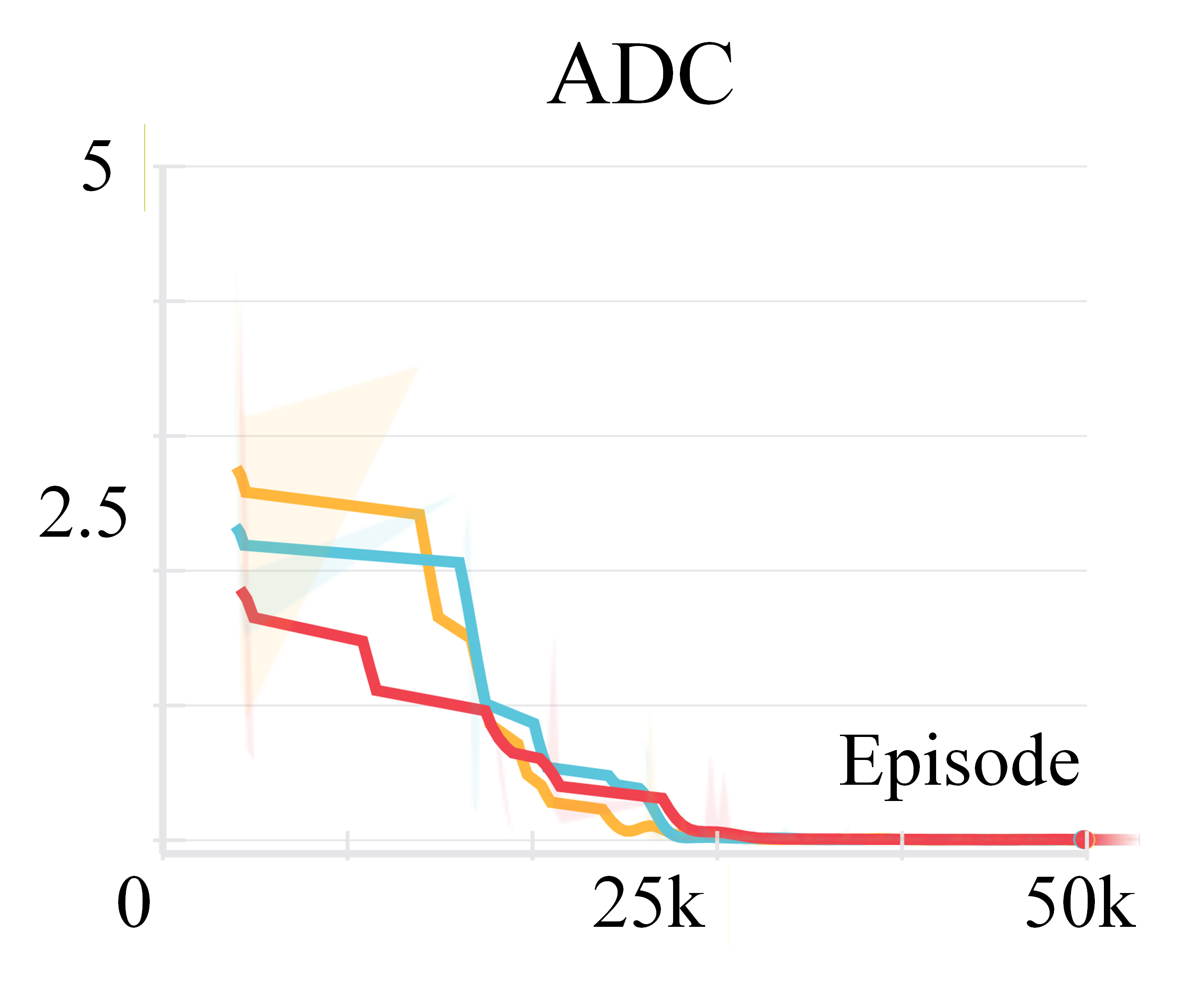}
    \includegraphics[width=0.15\textwidth]{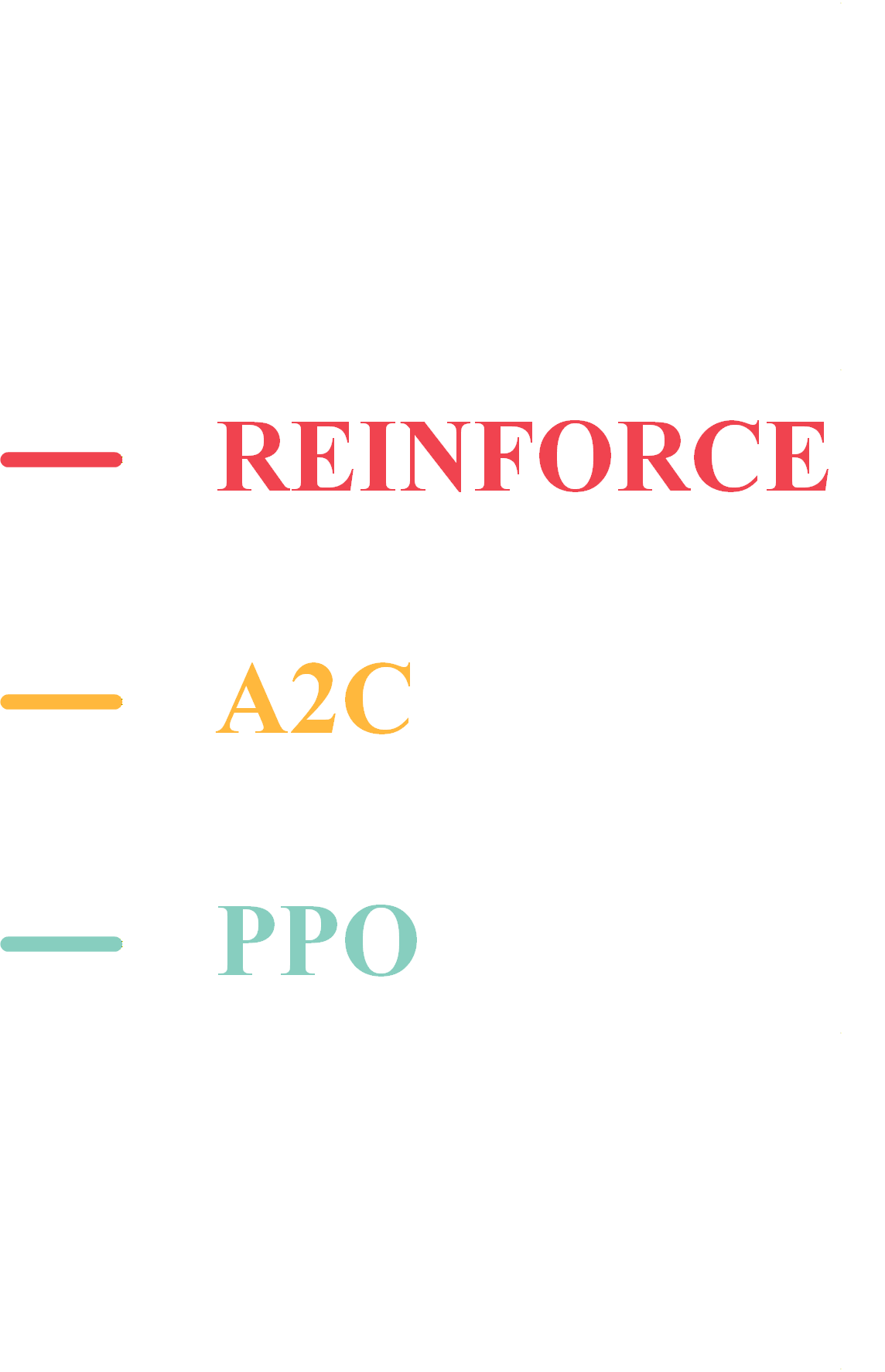}
    \caption{DECHRL's performance on the \textit{GetSilverore} ($\tau_{\max}=4$) tasks with varying gradient estimators (REINFORCE, PPO, A2C).}
    \label{fig:different_gradient}
\end{figure}

\textbf{\textit{Experimental Design.}}
We implement delay distribution modeling using three different gradient estimators (REINFORCE~\cite{DBLP:conf/iclr/BengioDRKLBGP20}, PPO~\cite{DBLP:journals/corr/SchulmanWDRK17}, and A2C~\cite{DBLP:journals/corr/MnihBMGLHSK16}), and then compare the performance of DECHRL when trained with the delay distributions obtained from each of these estimators. The results are shown in Figure~\ref{fig:different_gradient}.

\indent\textbf{\textit{Experimental Results.}}
The performance of all three gradient estimators is comparable, with differences primarily observed in the early stages of training. Specifically, in the early stages of training, REINFORCE quickly improved the success rate, demonstrating faster convergence due to its large gradient updates. As training progressed, both PPO and A2C slightly outperformed REINFORCE in convergence speed. This suggests that PPO and A2C offer better stability than REINFORCE, despite their slower initial convergence. Thereofre, REINFORCE achieves faster improvement in success rate during the early stages of training, whereas PPO and A2C exhibit more stable long-term performance and avoid premature convergence to suboptimal solutions. Each of these gradient estimators effectively models the delay distribution, consistently achieving a $100\%$ task success rate with only minor variations in convergence speed. Among these strong alternatives, we select REINFORCE for its simplicity and computational efficiency, which facilitate rapid initial exploration while maintaining low overhead.


\subsubsection{How does the inclusion of the delay-aware empowerment objective affect overall performance?} 

\indent\textbf{\textit{Experimental Design.}} 
We implement \textit{DECHRL/Emp}, a variant of \textit{DECHRL} that removes the delay-aware empowerment bonus. We evaluate the sub-goal policy training efficiency of \textit{DECHRL/Emp} and \textit{DECHRL} on the \textit{GetSilverore} task (\(\tau_{\max} = 4\), $\sigma_{delay} = 0.4$), with the results presented in Table~\ref{tab:w_wo_empowerment}.


\begin{table*}[htbp]
\centering
\resizebox{\textwidth}{!}{
\begin{tabular}{c|ccccc}
\toprule
\diagbox[width=4cm, height=0.75cm]{Variants}{Sub-goals} & Get Wood & Get Stone & Get Stick & Get Stone Axe & Get Silver Ore\\
\midrule
DECHRL & \textbf{1.00 ± 0.00} & \textbf{1.00 ± 0.00} & \textbf{0.99 ± 0.01} & \textbf{0.99 ± 0.01} & \textbf{0.99 ± 0.01}\\
DECHRL/Emp & 0.98 ± 0.02 & 0.99 ± 0.01 & 0.98 ± 0.11 & 0.83 ± 0.17 & 0.93 ± 0.26 \\
\bottomrule
\end{tabular}
}
\caption{With vs. Without Empowerment: Sub-goal Efficiency on the \textit{GetSilverore}  (\(\tau_{\max} = 4\), $\sigma_{delay} = 0.4$) task.}
\label{tab:w_wo_empowerment}
\end{table*}

\indent\textbf{\textit{Experimental Results.}}
\textit{DECHRL} outperforms its ablated variant in training efficiency across all subgoals. The strategies trained with the delay-aware empowerment objective are clearly more effective in ensuring thorough exploration of the subgoals in the environment, leading to faster and better performance.


\subsubsection{How does DECHRL perform under varying degrees of delay stochasticity?}\label{sec:exp_Generalization}

\begin{figure}[h]
    \centering
    \includegraphics[width=0.29\textwidth]{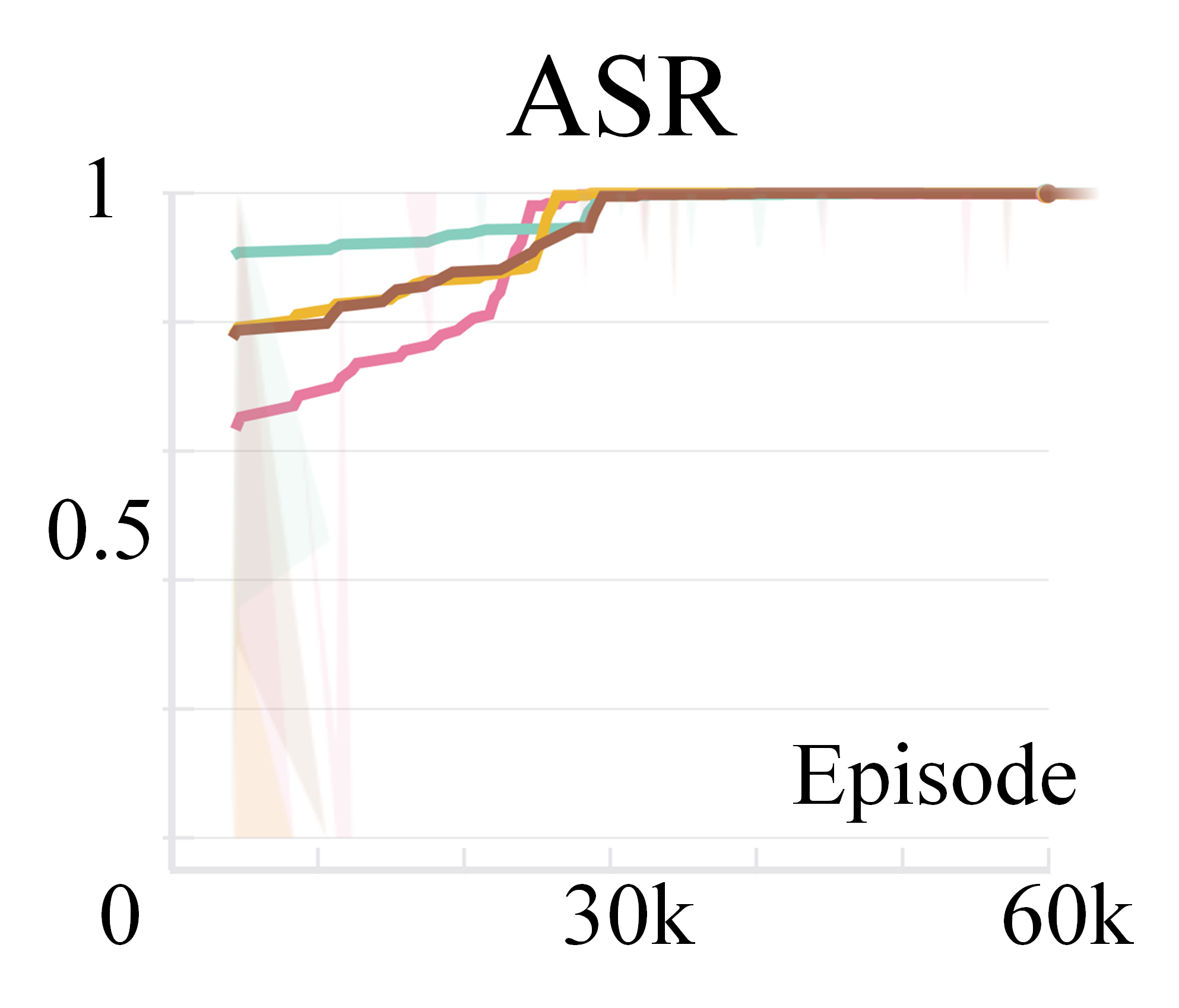}
    \includegraphics[width=0.29\textwidth]{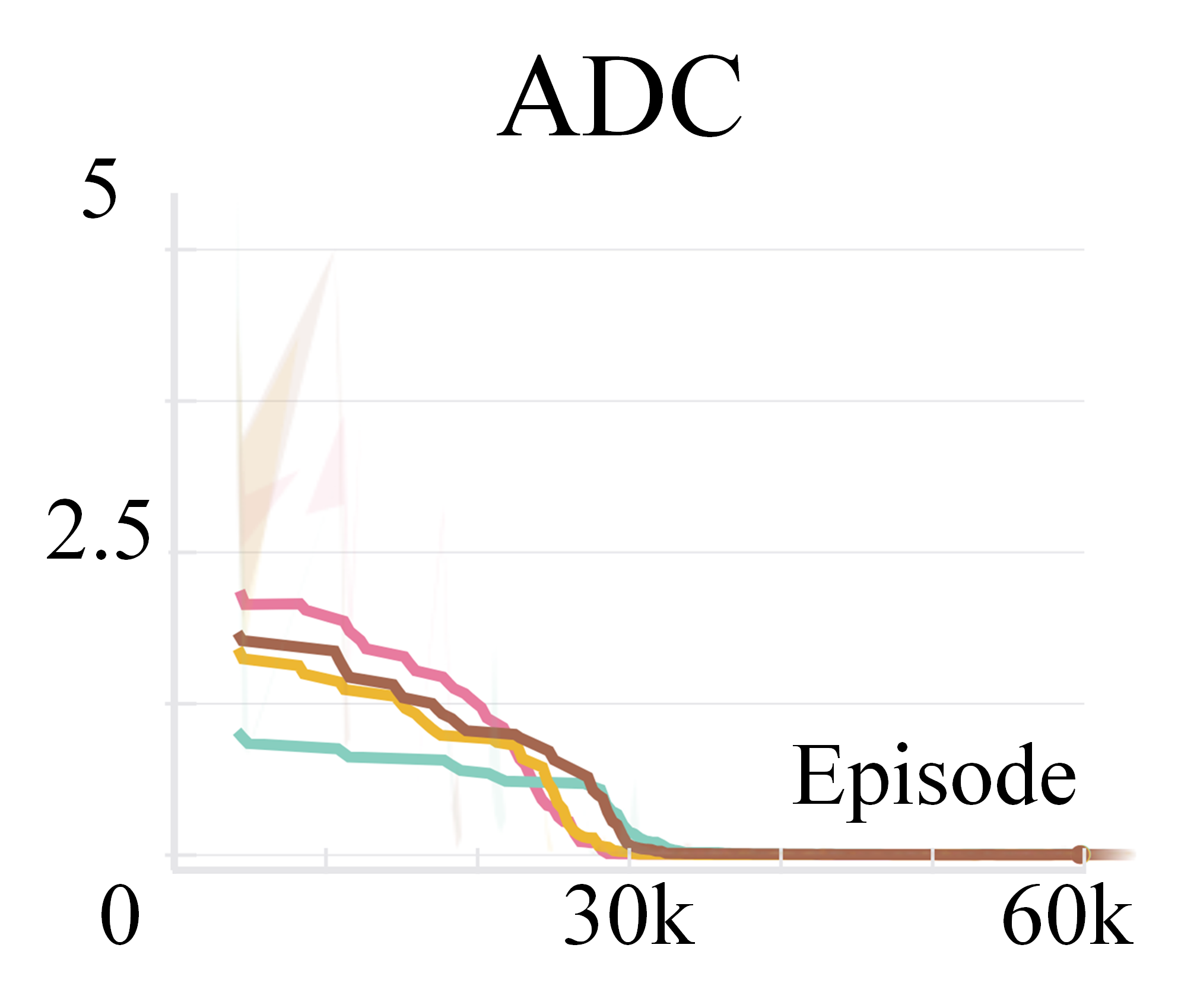}
    \includegraphics[width=0.07\textwidth]{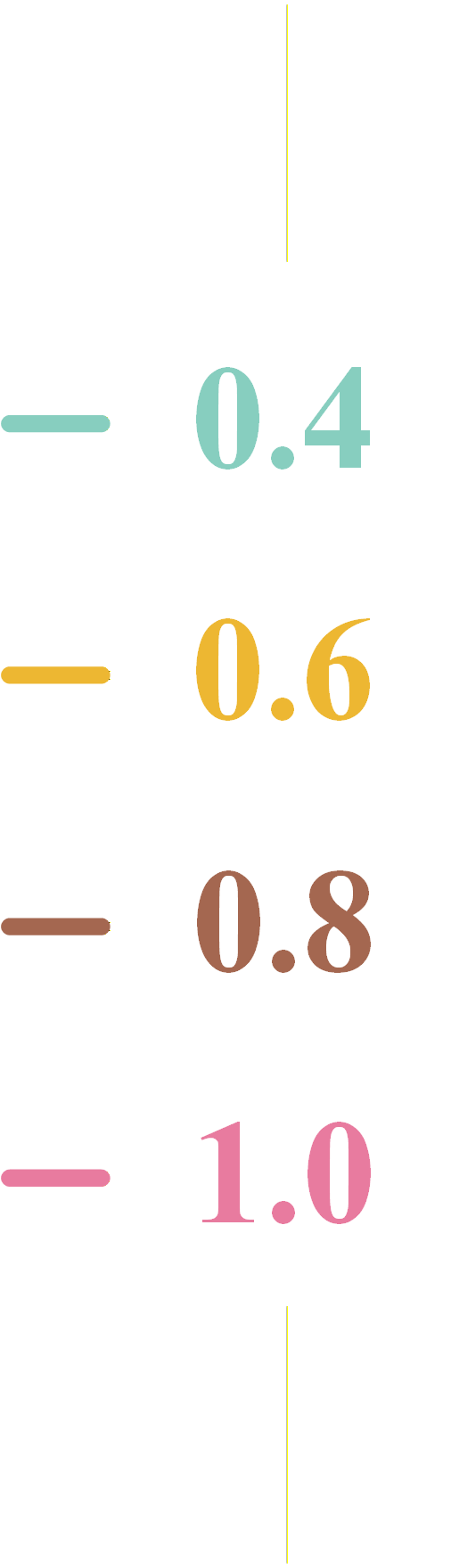}
    \caption{DECHRL's performance on the \textit{GetSilverore} ($\tau_{\max}=4$) tasks with varying $\sigma_{delay}=\{0.4,0.6,0.8,1.0\}$.}
    \label{fig:different_sigma}
\end{figure}

\textbf{\textit{Experimental Design.}}
We evaluated DECHRL on the \textit{GetSilverore} task under four levels of stochastic delay, with $\sigma_{delay} \in \{0.4, 0.6, 0.8, 1.0\}$. The corresponding delay distributions for these four levels of stochastic delay are illustrated in Figure~\ref{fig:minecraft_delay_dist}. The results are presented in Figure~\ref{fig:different_sigma}.

\indent\textbf{\textit{Experimental Results.}}
Increasing $\sigma_{delay}$ leads to greater uncertainty in state transition delays, posing challenges not only for delay detection but also for effective adaptation. Nevertheless, as shown in Figure~\ref{fig:different_sigma}, \textbf{DECHRL} maintains consistently strong performance across all delay settings, highlighting its robustness and generalization under varying levels of temporal uncertainty.

\subsubsection{How does Simplified-DECHRL perform under different delay modeling granularities and maximum delays?}\label{sec:exp_scalability}

\begin{figure*}[h]
    \centering
    \setlength{\tabcolsep}{3pt} 
    \includegraphics[width=0.75\textwidth]{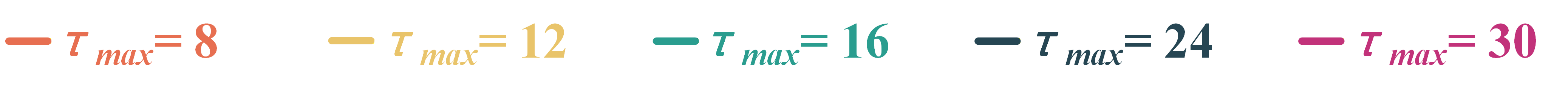}\\
    \begin{tabular}{cc}
        \subcaptionbox{$\kappa=2$}{
            \includegraphics[width=0.23\textwidth]{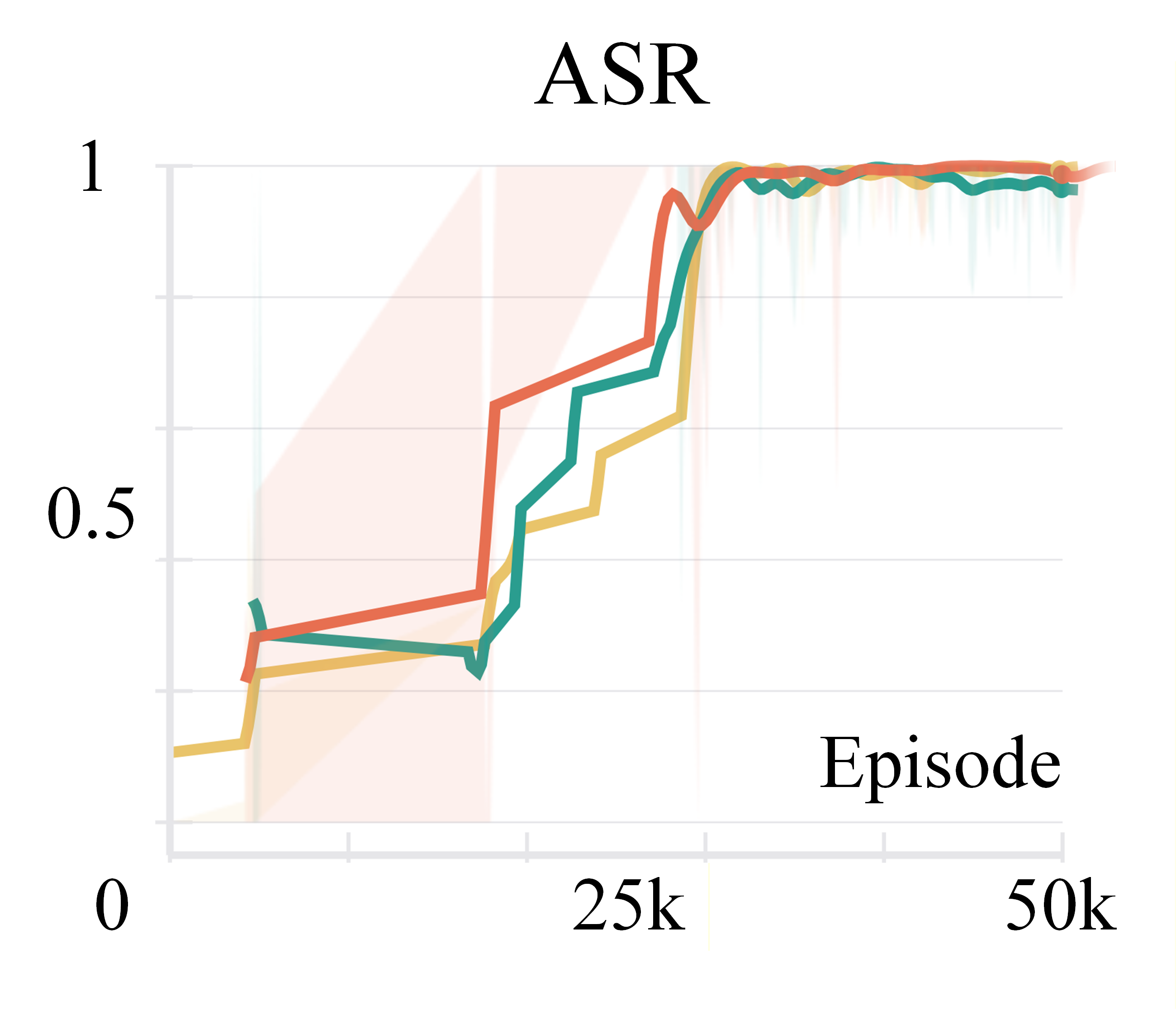}
            \hspace{-0.16cm}
            \includegraphics[width=0.23\textwidth]{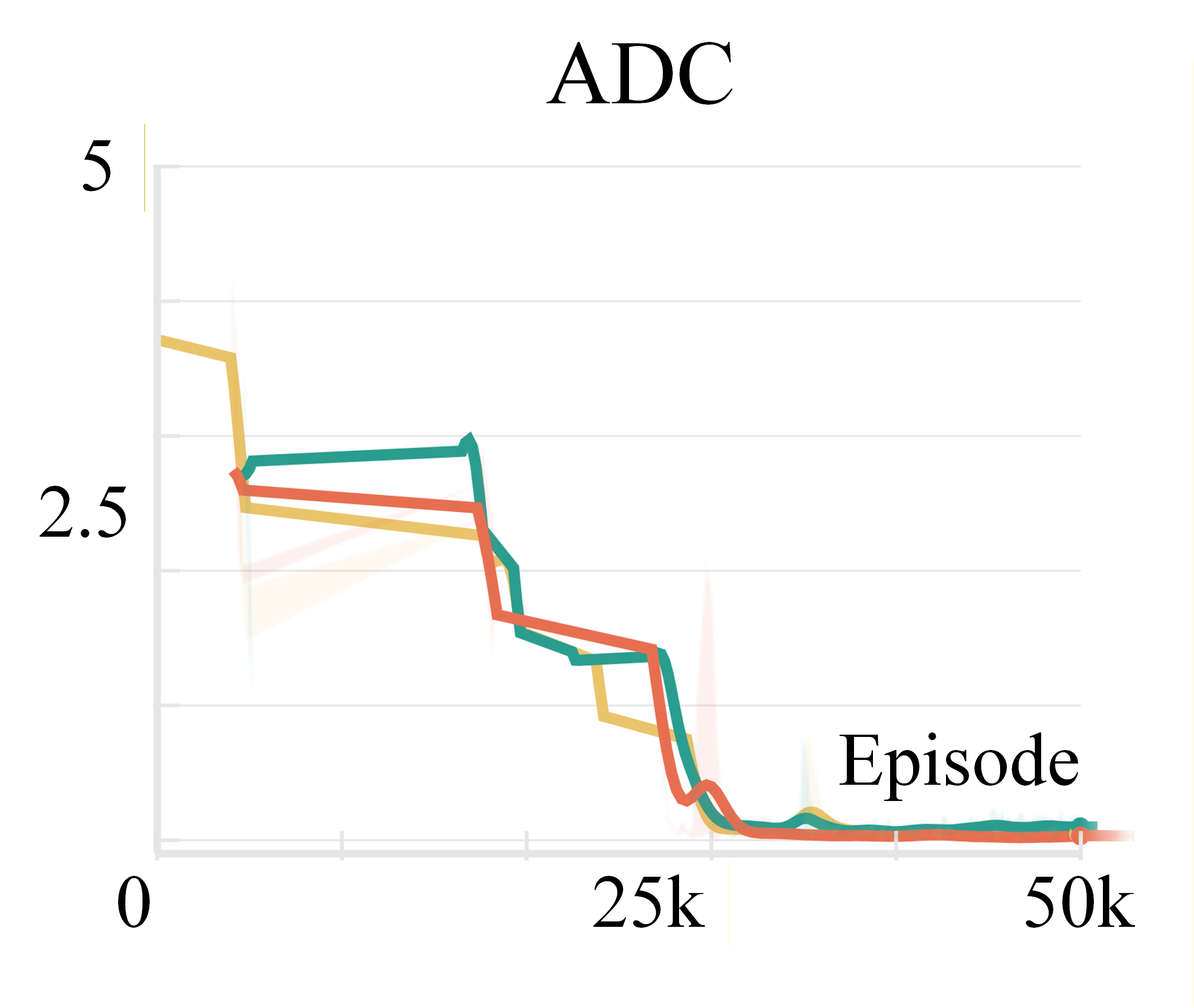}} &
        \subcaptionbox{$\kappa=4$}{
            \includegraphics[width=0.23\textwidth]{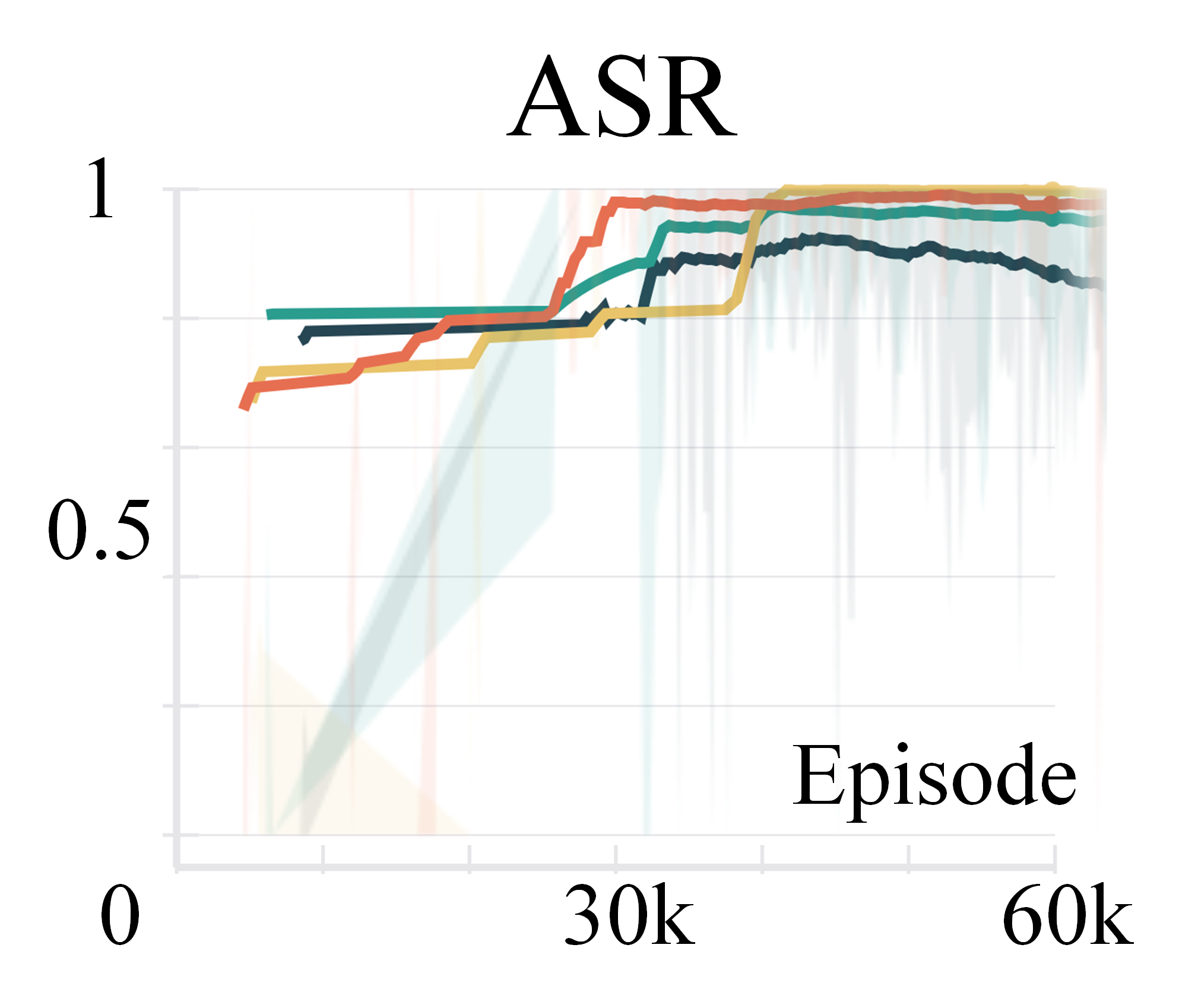}
            \hspace{-0.16cm}
            \includegraphics[width=0.23\textwidth]{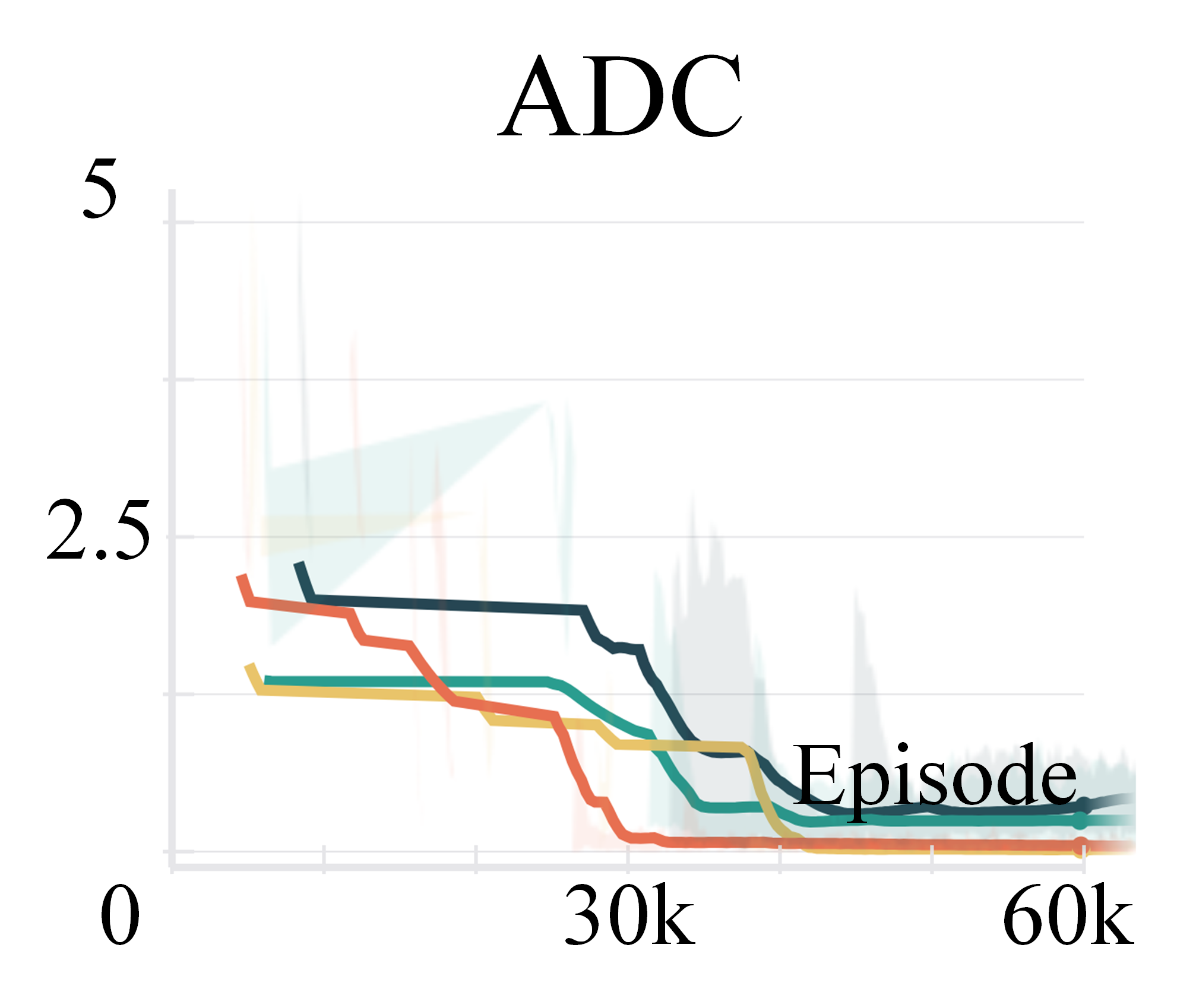}} \\
        \subcaptionbox{$\kappa=6$}{
            \includegraphics[width=0.23\textwidth]{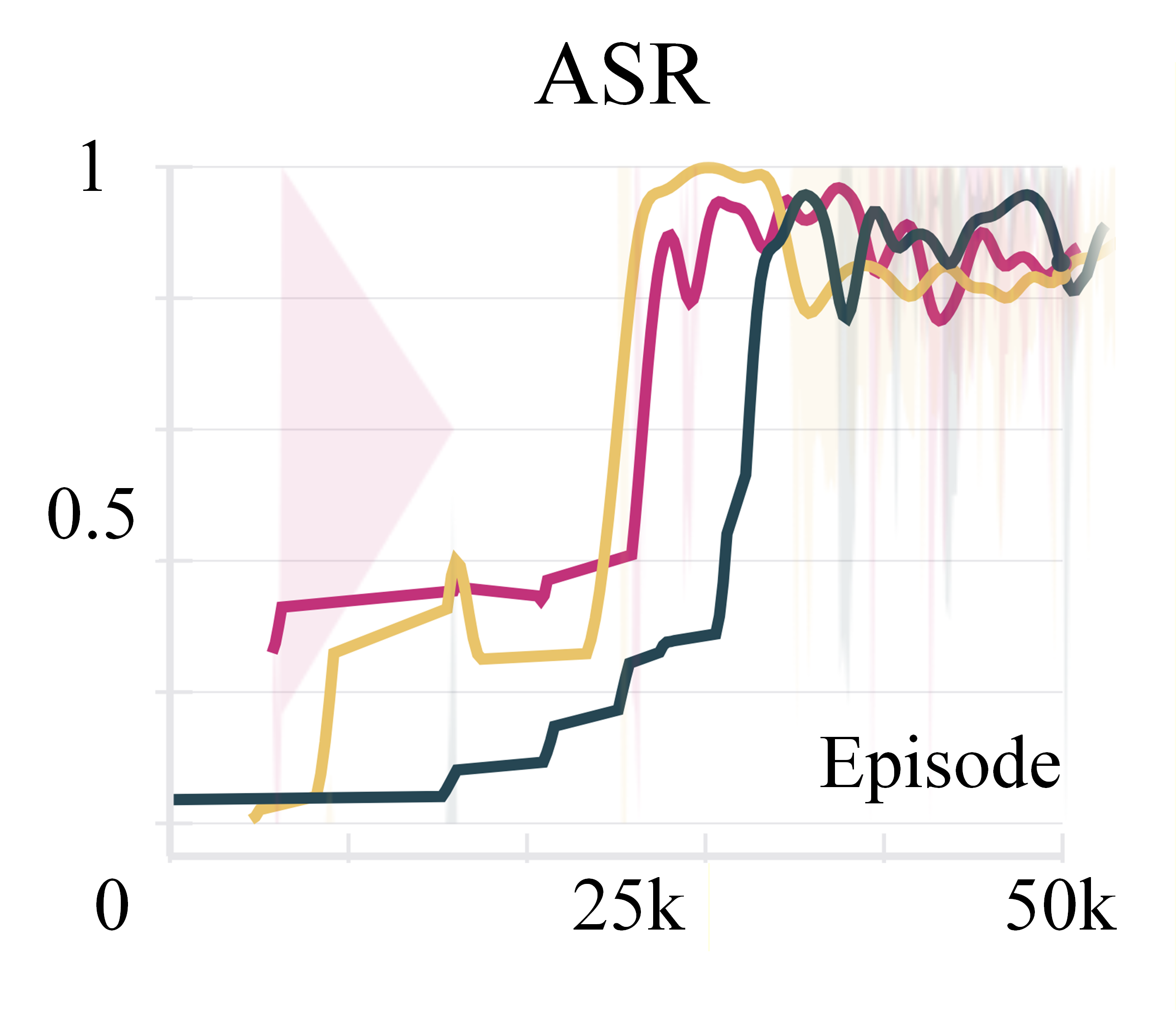}
            \hspace{-0.16cm}
            \includegraphics[width=0.23\textwidth]{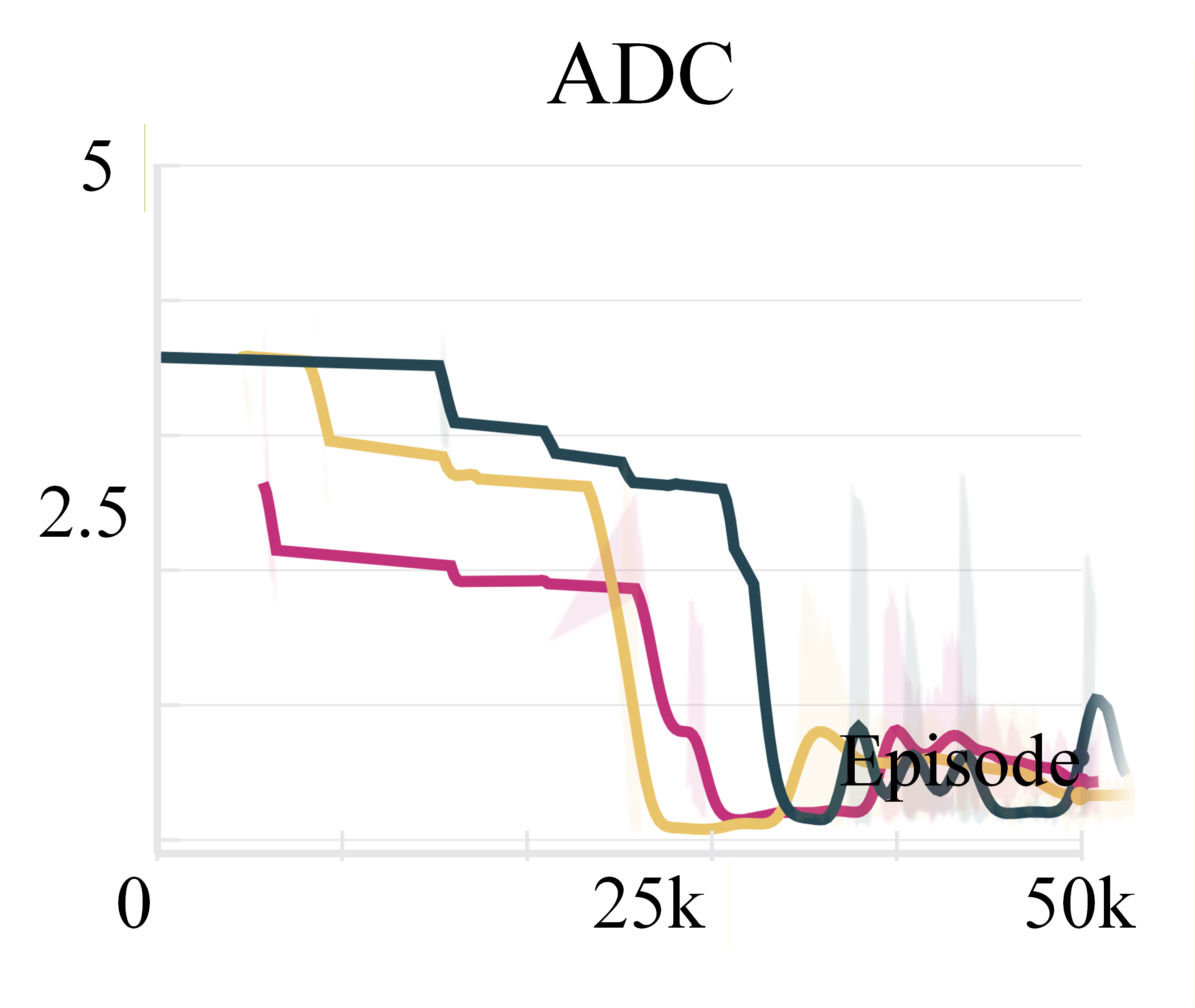}} &
        \subcaptionbox{$\kappa=8$}{
            \includegraphics[width=0.23\textwidth]{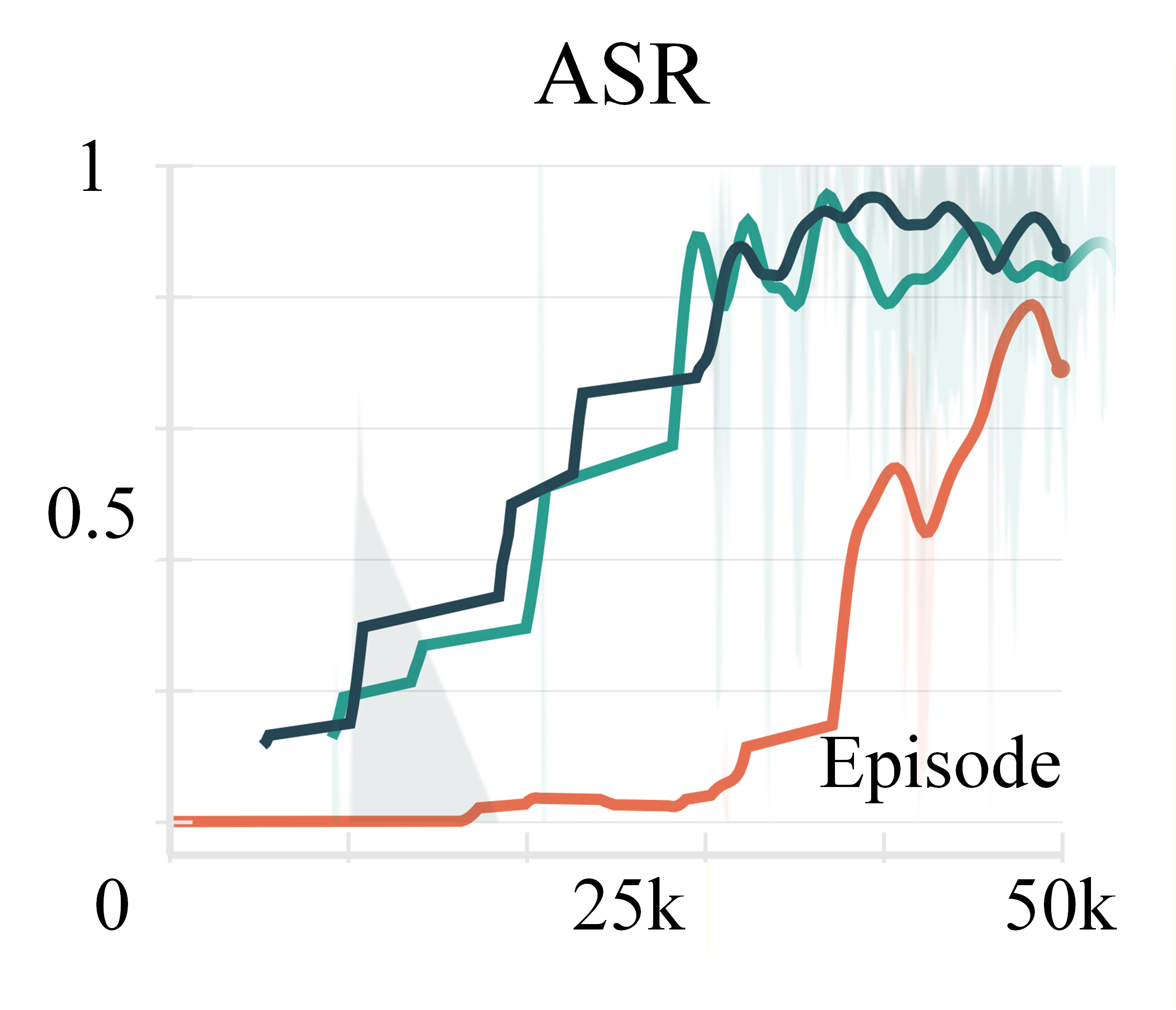}
            \hspace{-0.16cm}
            \includegraphics[width=0.23\textwidth]{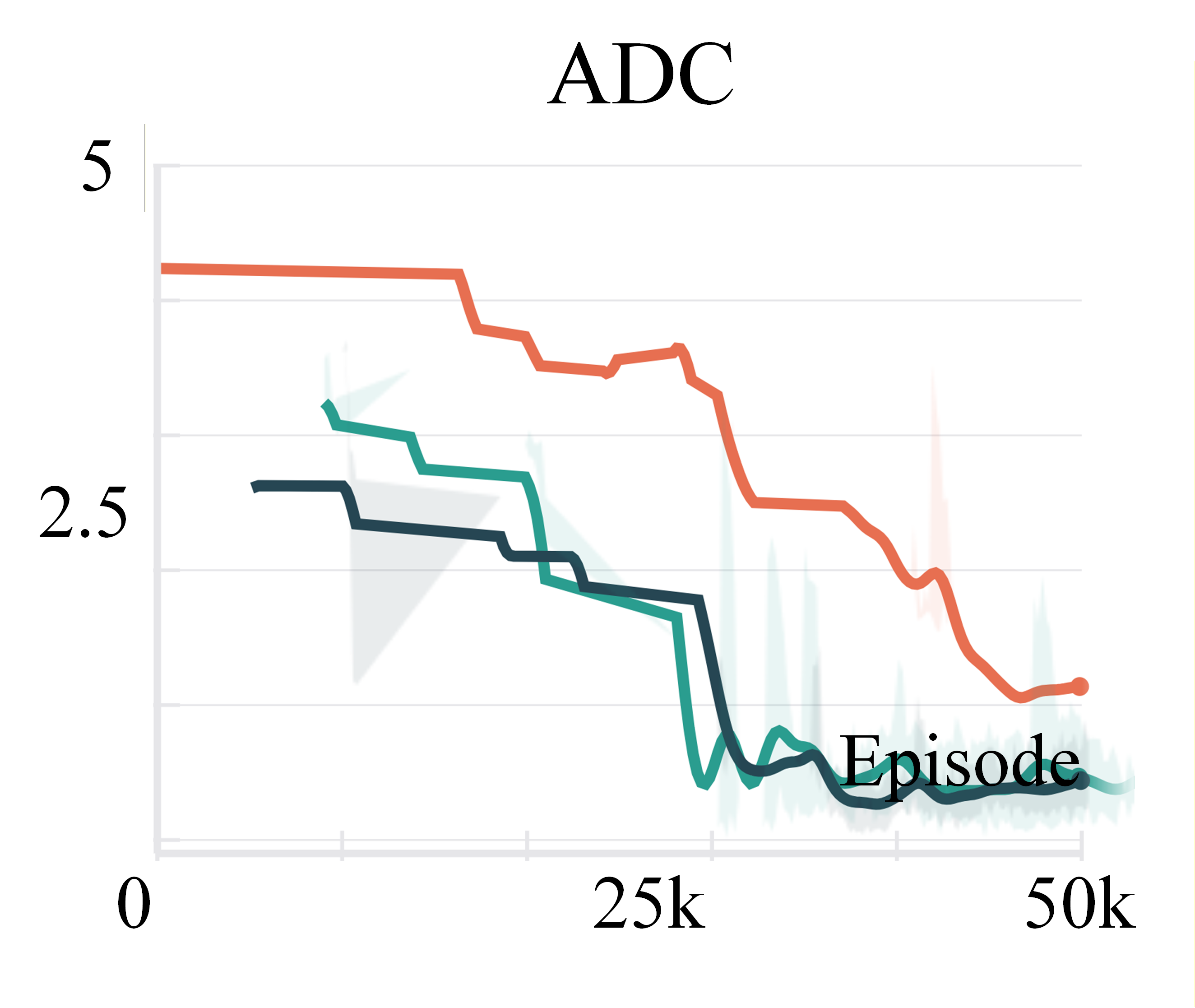}}
    \end{tabular}
    \caption{The ASR and ADC of Simplified-DECHRL on the \textit{GetSilverore} task under different maximum delays $\tau_{\max} \in \{8, 12, 16, 24, 30\}$ with modeling granularities $\kappa \in \{2, 4, 6, 8\}$.}
    \label{fig:different_tau_max}
\end{figure*}

\textbf{\textit{Experimental Design.}} 
We evaluate the overall performance of Simplified-DECHRL on the \textit{GetSilverore} task under different delay modeling granularities $\kappa \in \{2, 4, 6, 8\}$ and maximum environment delays $\tau_{\max} \in \{8, 12, 16, 24, 30\}$, with the results shown in Figure~\ref{fig:different_tau_max}.

\indent\textbf{\textit{Experimental Results.}} 
We first analyze how the choice of delay-modeling granularity $\kappa$ affects both performance and computational cost. A finer granularity ($\kappa = 2$) enables more accurate modeling of the delay distribution and thus better preserves the overall algorithmic performance. However, it incurs substantial process overhead; for instance, when the maximum delay is $\tau_{\max} = 24$, modeling delays with $\kappa = 2$ requires 12 parallel processes, whereas using a coarser granularity of $\kappa = 4$ reduces this number to only 6. At the other extreme, a coarser granularity ($\kappa = 8$) greatly reduces the process cost, but the resulting delay model is overly coarse and leads to a noticeable drop in performance. Intermediate granularities $\kappa \in \{4, 6\}$ offer a good trade-off between performance and computational overhead, with the finer granularity ($\kappa = 4$) achieving marginally superior results compared to $\kappa = 6$. In general, the choice of $\kappa$ should be tailored to the specific task; in the environments considered in this work, $\kappa= 4$ proved to be the most suitable.
Having fixed a reasonable delay-modeling granularity ($\kappa=4$), we next examine the impact of the maximum delay $\tau_{\max}$ on algorithm performance. Figure~\ref{fig:different_tau_max}(b) shows that as $\tau_{\max}$ increases, both the convergence speed and the final performance exhibit a slight decline; nevertheless, even at $\tau_{\max} = 24$, the algorithm still maintains strong performance. Although the simplified variant incurs a modest loss in performance, it substantially improves scalability by reducing computational cost. Given the considerable savings in computational resources, this trade-off is both acceptable and practically worthwhile.


\subsubsection{How sensitive is DECHRL to its hyperparameters $\lambda_1$ and $\lambda_2$ during delay distribution modeling?}\label{sec:sensitivity}

\begin{figure}[h]
    \centering
    \includegraphics[width=0.29\textwidth]{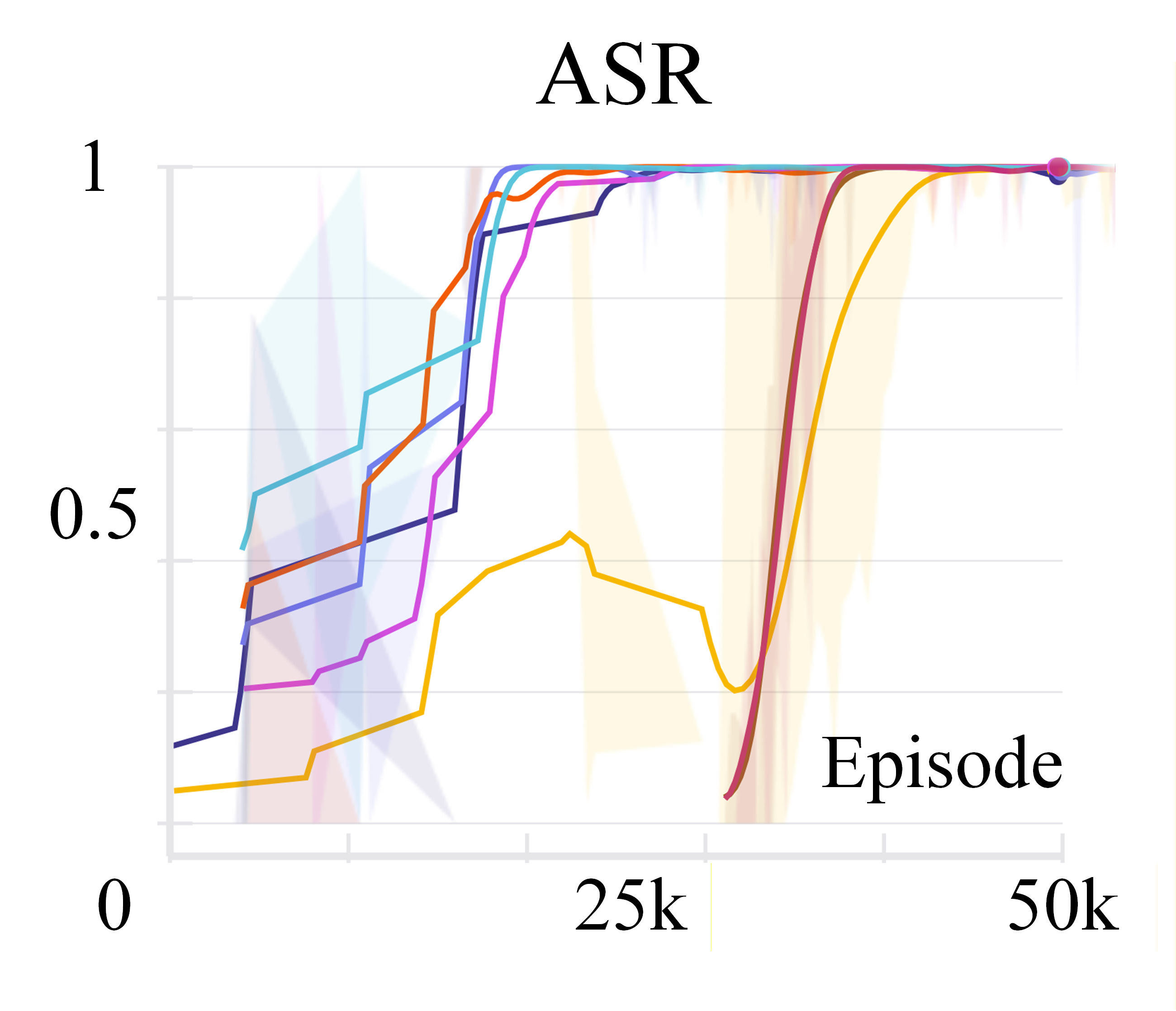}
    \includegraphics[width=0.29\textwidth]{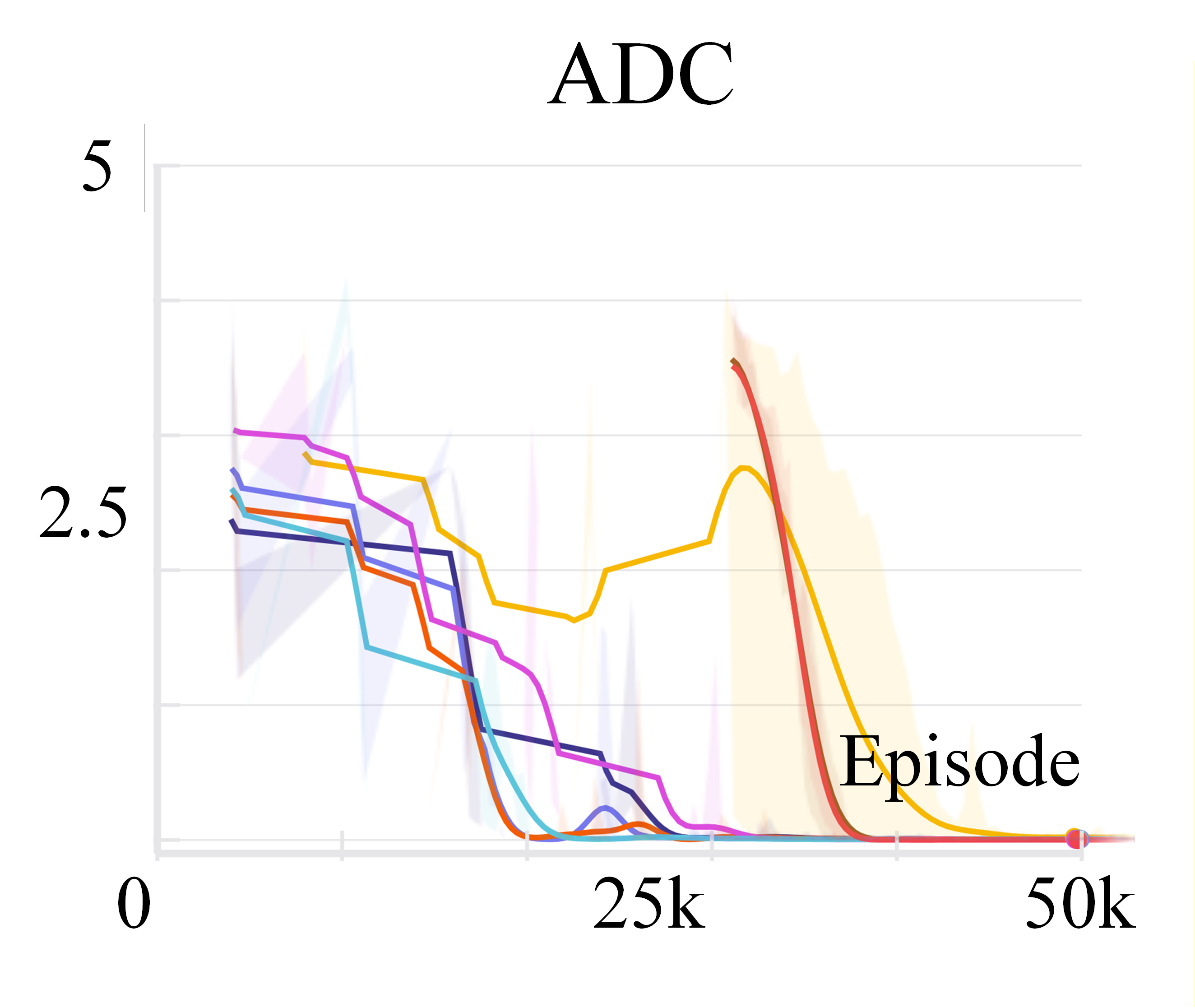}
    \includegraphics[width=0.102\textwidth]{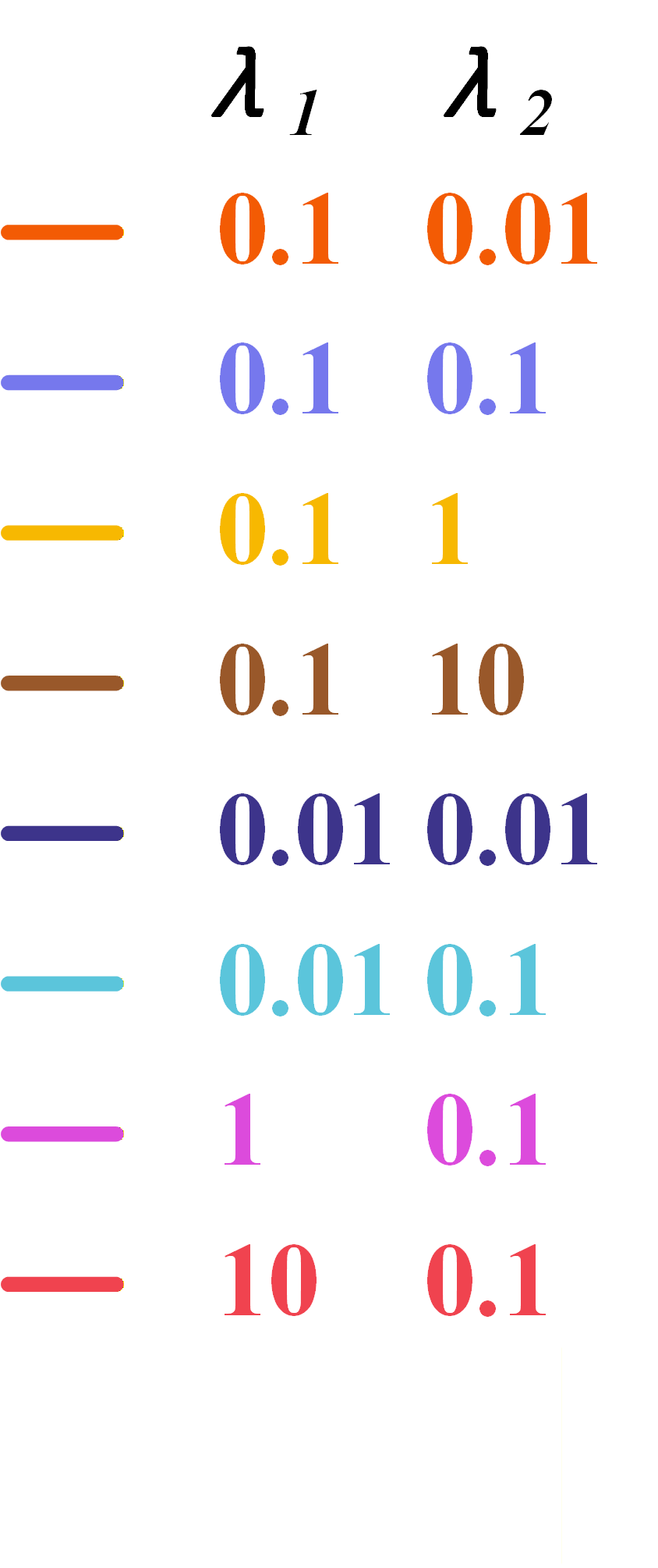}
    \caption{DECHRL's performance on the \textit{GetSilverore} ($\tau_{\max}=4, \sigma_{delay}=0.8$) task with varying values of hyperparameters $\lambda_1, \lambda_2$.}
    \label{fig:different_hyperparams}
\end{figure}

\textbf{\textit{Experimental Design.}}
To systematically assess the impact of hyperparameter variations on DECHRL's performance, we evaluated it on the \textit{GetSilverore} ($\tau_{\max}=4, \sigma_{delay}=0.8$) task across a range of values spanning multiple orders of magnitude: $\{0.01, 0.1, 1, 10\}$, with the results shown in Figure~\ref{fig:different_hyperparams}.


\indent\textit{\textbf{Experimental Results.}}
DECHRL exhibits robust performance and consistent, rapid convergence when both $\lambda_1$ and $\lambda_2$ are set between $0.01$ and $0.1$. Setting $\lambda_1$ to $1$ also yields rapid convergence, whereas increasing $\lambda_2$ to $1$ causes a substantial slowdown in convergence. When either $\lambda_1$ or $\lambda_2$ is set to $10$, the convergence rate drops drastically. Despite these variations in convergence speed, DECHRL consistently reaches a success rate close to $100\%$ across all hyperparameter settings. Based on these observations, we set both $\lambda_1$ and $\lambda_2$ to $0.05$, which lies near the midpoint of the effective range ($0.01\sim 0.1$). We would like to emphasize that the \textcolor[HTML]{99582a}{\textit{brown}} and \textcolor[HTML]{f0434f}{\textit{red}} curves only begin to produce meaningful results after approximately 25k training episodes. This delay is due to their highly imbalanced hyperparameter settings: $(\lambda_1, \lambda_2) = (\textcolor[HTML]{99582a}{0.1}, \textcolor[HTML]{99582a}{10})$ for the brown variant and $(\lambda_1, \lambda_2) = (\textcolor[HTML]{f0434f}{10}, \textcolor[HTML]{f0434f}{0.1})$ for the red variant. These extreme configurations impede the reliable estimation of causal structures and delay distributions, consequently delaying the onset of hierarchical policy training. As a result, these variants require substantially more interaction data before they can complete the causal discovery stage and transition into hierarchical policy training.


\subsubsection{How sensitive is DECHRL to its sub-goal success ratio threshold?}\label{sec:success_ratio_sentivity}

\begin{figure}[h]
    \centering
    \includegraphics[width=0.29\textwidth]{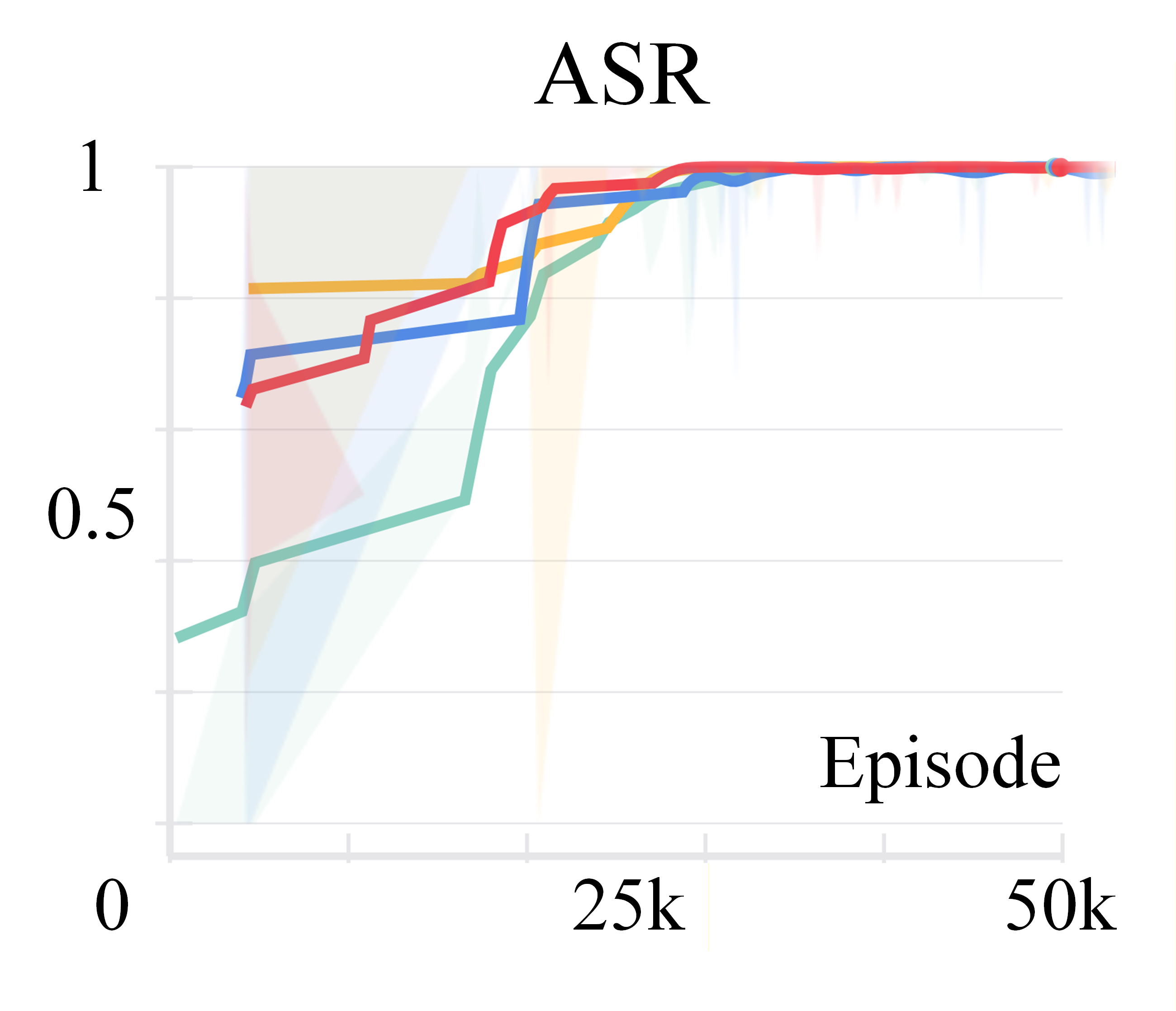}
    \includegraphics[width=0.29\textwidth]{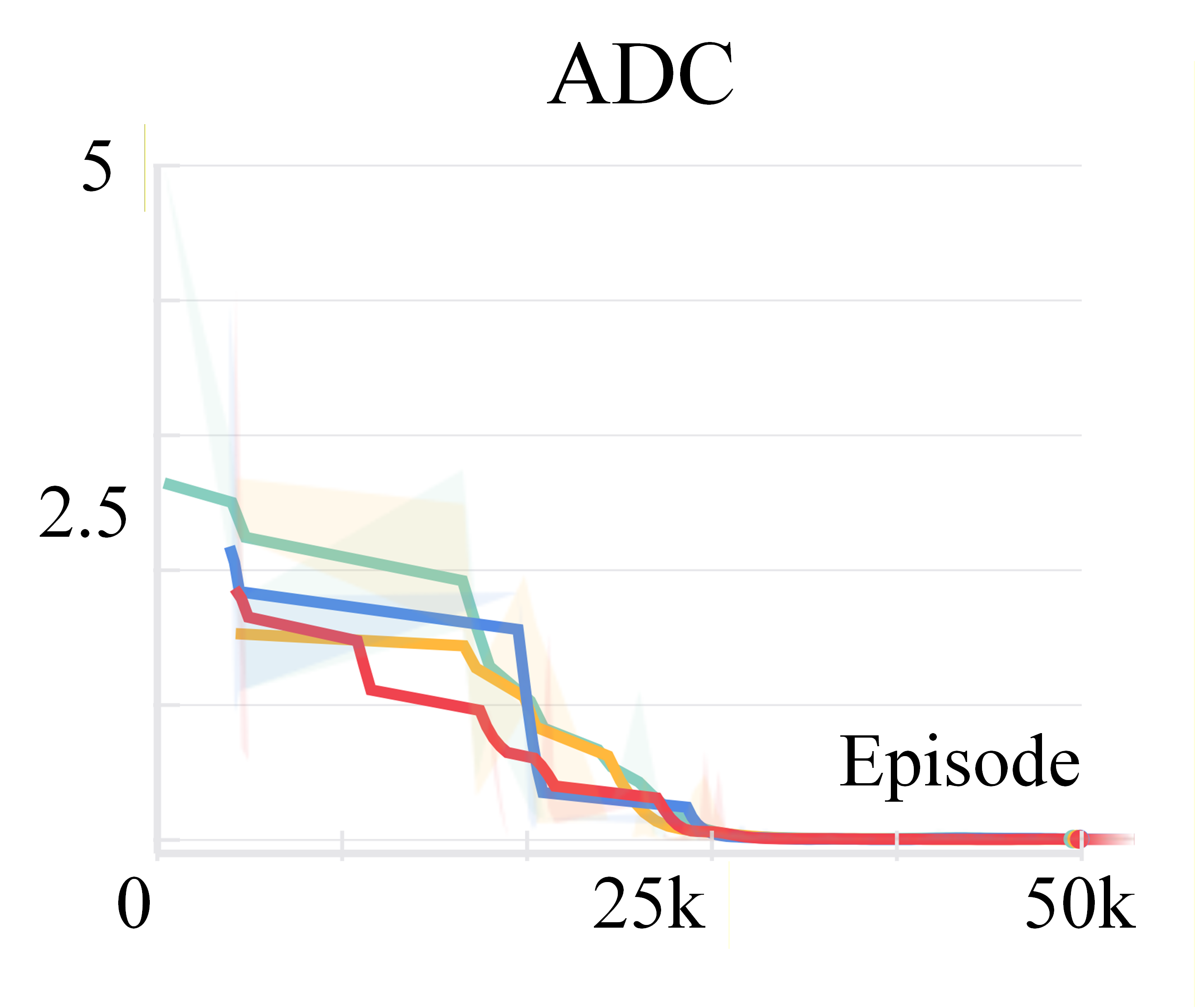}
    \includegraphics[width=0.163\textwidth]{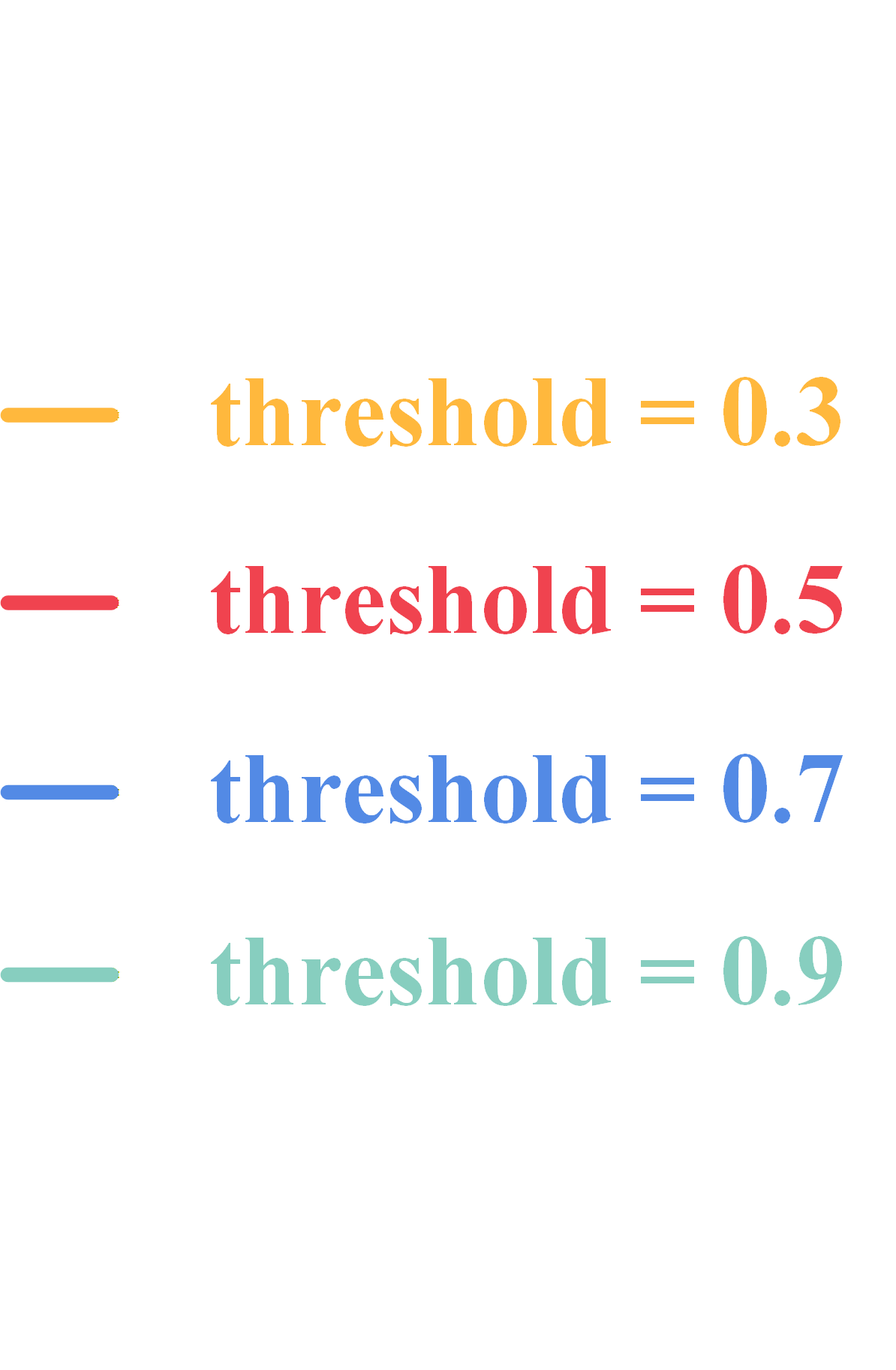}
    \caption{DECHRL's performance on the \textit{GetSilverore} ($\tau_{\max}=4, \sigma_{delay}=0.4$) task with varying values of subgoal success ratio threshold.}
    \label{fig:different_suc_threshold}
\end{figure}

\textit{\textbf{Experimental Design.}}
We evaluate the performance of our DECHRL on the \textit{GetSilverore} ($\tau_{\max}=4, \sigma_{delay}=0.4$) task under various subgoal success-rate thresholds to assess their influence, with the results presented in Figure~\ref{fig:different_suc_threshold}.

\indent\textbf{\textit{Experimental Results.}}
The achievement of higher-level subgoals relies on the reliable execution of lower-level subgoals. If the success-rate threshold is set too low, it can undermine the accomplishment of higher-level subgoals. On the other hand, setting the threshold too high can present challenges, as higher-level subgoals are inherently more difficult to train. Achieving a very high success rate within a limited timeframe can be somewhat challenging, and failure to meet the threshold within the allotted time can hinder subsequent learning, slowing the overall convergence of the algorithm. To strike a balance between robust subgoal execution and efficient convergence, we choose a success-rate threshold of $0.5$ for subgoal training.


\section{Conclusion}


Existing delay-aware RL methods typically address delays by augmenting the state space, incorporating prior knowledge of delay distributions, or relying on non-delayed data. These strategies often suffer from limited scalability and poor generalization.
While HRL is inherently suited to handling delays, current methods are restricted to fixed-delay settings.
To overcome these limitations, we propose DECHRL, which leverages multi-timescale causal discovery to model delay distributions and integrates them into a delay-aware empowerment objective. This allows agents to infer delay dynamics internally and explore more effectively, enabling efficient hierarchical reinforcement learning. We further propose a simplified variant to enhance scalability with minimal compromise in performance.
Currently, our approach is limited to tasks with discrete, semantically structured state spaces. Extending DECHRL to continuous or high-dimensional observations, such as pixel-based inputs, is a promising direction for future work toward broader applicability in complex, realistic environments.

\appendix

\section{Enhancement of Hierarchical Reinforcement Learning Baselines}\label{app:enhancement}

\subsection{Enhancement Methodology}

To ensure a fair comparison, we enhance the baseline hierarchical reinforcement learning (HRL) algorithms along two key dimensions:



\paragraph{\textbf{Enhancing Delay Handling Capability}}
Among the hierarchical reinforcement learning (HRL) baselines, only D3HRL explicitly models transition-specific fixed delays, and the option-based methods (Option-Critic and LESSON) implicitly handle variable delays through their termination functions. In contrast, CDHRL and HAC lack any intrinsic mechanism to cope with action-effect delays. To ensure a fair comparison, we equip these two methods with the ability to operate under a fixed delay horizon $\tau_{\max}$.
Specifically, for CDHRL, we modify both its causal discovery and policy training phases: during intervention data collection for causal structure learning, we enforce a temporal offset of exactly $\tau_{\max}$ steps between actions and their observed effects; the same offset is applied when gathering transition data for hierarchical policy updates. This allows CDHRL to correctly infer and learn from $\tau_{\max}$-step delayed dependencies. Similarly, for HAC, we align its experience collection process with the $\tau_{\max}$-step delay by sampling transitions at this fixed interval during policy training, enabling it to adapt its subgoal generation and low-level control to the delayed feedback.

\paragraph{\textbf{Enhancing Causal Awareness}}
D3HRL, CDHRL, and our DECHRL are capable of autonomously discovering causal relationships from interaction data—a key enabler for iterative environment exploration and dynamic construction of hierarchical policies. The remaining baselines (HAC, LESSON, and Option-Critic), however, have no built-in causal reasoning mechanism. Comparing them directly would therefore be unfair, as they cannot acquire the structural knowledge that informs effective subgoal decomposition.
To ensure a fair comparison, we provide all non-causal baselines with ground-truth causal information through a structured curriculum. Rather than discovering dependencies from scratch, these methods are guided to learn subgoals in a staged, bottom-up fashion: training begins with the most immediate, low-level subgoals, and only after mastery is demonstrated do we introduce higher-level objectives that depend on previously learned ones. This curriculum applies uniformly across architectures—whether multi-level (as in HAC) or flat option-based frameworks (as in LESSON and Option-Critic). For the latter, although they lack explicit hierarchical layers, we still sequence their training by gradually expanding the set of achievable subgoals, effectively simulating depth through curriculum progression. In this way, all baselines operate with equivalent causal priors.

\subsection{Empirical Analysis of Enhancements}

\paragraph{\textbf{Experimental Design}}

\begin{figure*}[htbp]
    \centering
    \subcaptionbox{ASR}{\includegraphics[width=0.29\textwidth]{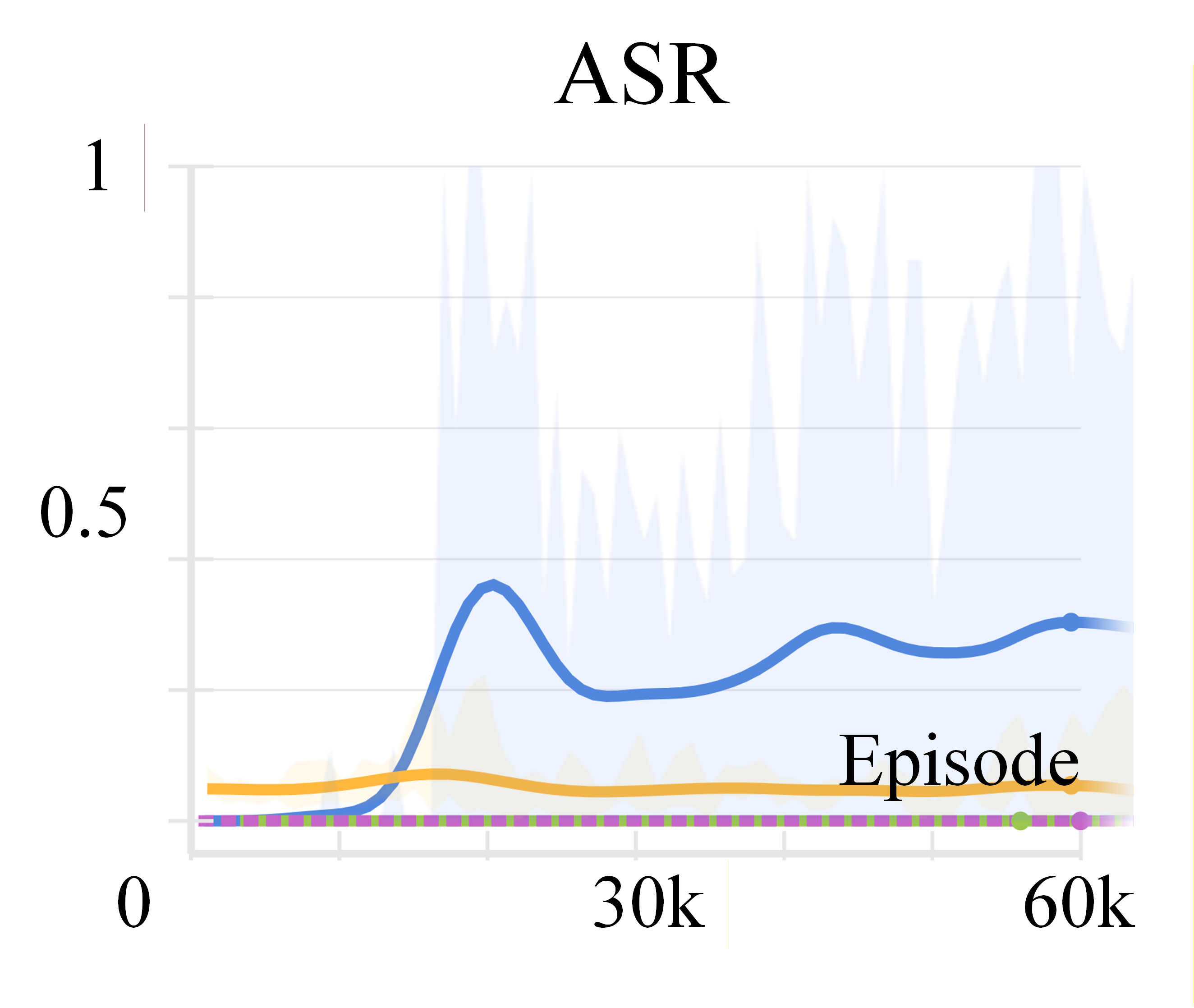}}
    \subcaptionbox{ADC}{\includegraphics[width=0.29\textwidth]{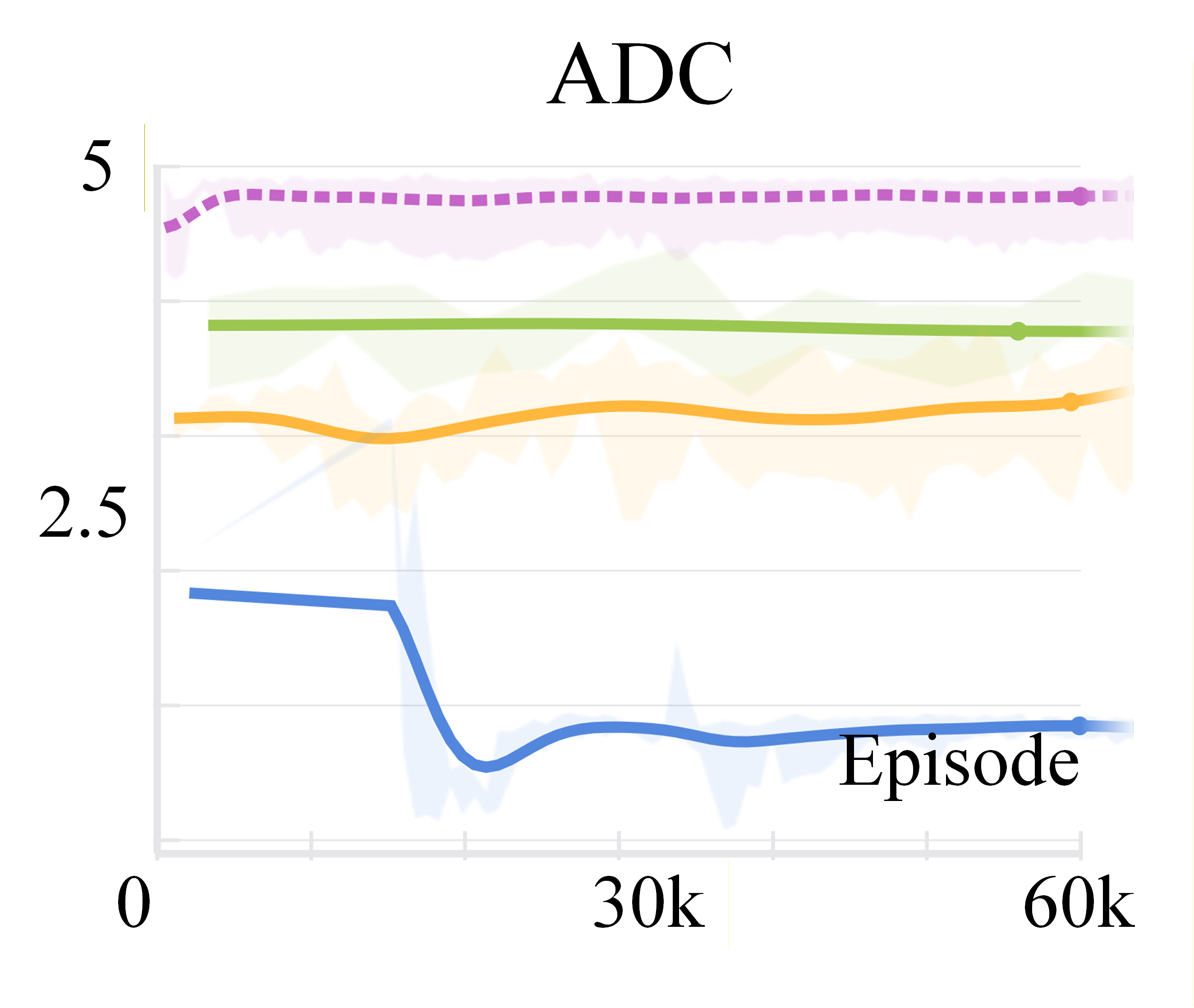}}
    \includegraphics[width=0.15\textwidth]{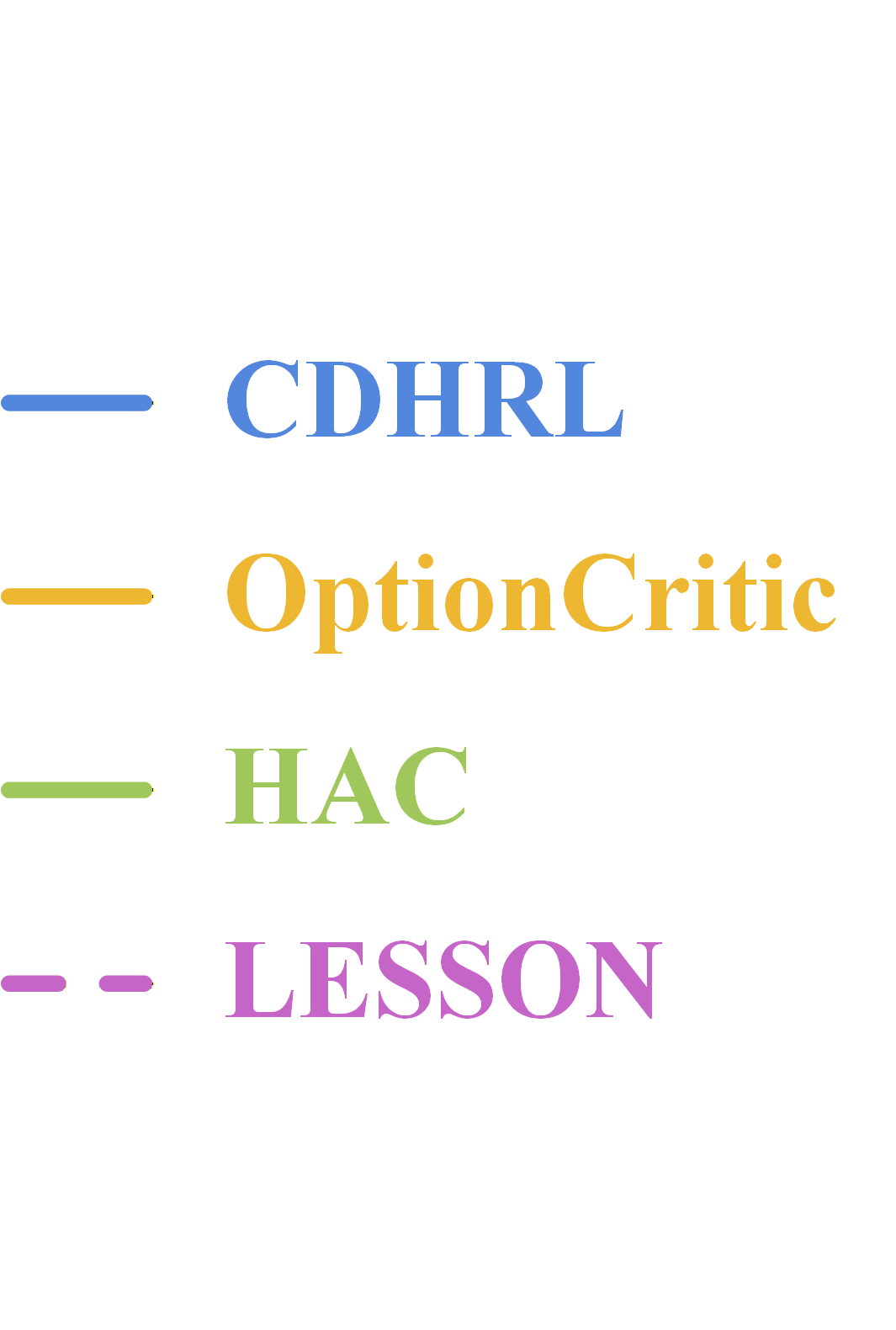}
    \caption{The performance of the unenhanced baselines on the \textit{GetSilverore} ($\tau_{\max} = 4, \sigma_{delay} = 0.4$) task.}\label{fig:com_baseline}
\end{figure*}

To assess the impact of the proposed enhancements, we conducted an evaluation of the unenhanced baseline algorithms on the \textit{GetSilverore} ($\tau_{\max} = 4, \sigma_{delay} = 0.4$) task, as illustrated in Figure~\ref{fig:com_baseline}.

\paragraph{\textbf{Experimental Results}}


A comparison between the pre-enhancement results in Figure~\ref{fig:com_baseline} and the post-enhancement results in Figure~\ref{fig:genernal_exp}(a) demonstrates the effectiveness of our enhancement approach, although its impact varies considerably across different algorithms. Among them, \textbf{CDHRL}, which is closely related to our method within the CHRL framework, exhibits the most pronounced improvement. Originally limited to handling only single-step state transitions or fixed single-step delays, CDHRL gains the ability to perceive delays up to $\tau_{\max}$ steps after our enhancement. This newfound capability naturally leads to a substantial performance boost. \textbf{HAC}, which employs a hierarchical multi-level architecture, also achieves notable gains following the integration of delay-awareness and causal-awareness mechanisms. However, because each policy in HAC is responsible for managing all subgoals at its corresponding level and these subgoals inherently exhibit varying delays, it becomes challenging to successfully train all subgoals simultaneously within the same hierarchy. This inconsistency hampers the learning of higher-level subgoals and ultimately constrains overall performance. In contrast, both \textbf{LESSON} and \textbf{Option-Critic} show virtually no improvement in terms of ASR, which remains close to zero even after enhancement. Nevertheless, their ADC metrics indicate that the enhanced versions are indeed capable of exploring deeper causal chains. The limited performance gain and minimal before-after difference can be attributed to their lack of a hierarchical structure, which prevents them from retaining previously learned subgoals during curriculum learning. This is a critical limitation that severely undermines their overall efficacy.

\section{Computational Complexity of Delay Distribution Modeling}\label{app:delay_complexity}


The delay distribution modeling component of DECHRL employs the REINFORCE algorithm to learn the delay matrix parameters $\mathbf{H}$, where each effect variable $S_i$ has a delay distribution over $\tau_{\max}$ possible time steps. We summarize all parameters appearing in the computational cost analysis, along with their definitions, in Table~\ref{tab:delay_complexity_symbols}.

\begin{table}[h]
\centering
\begin{tabular}{@{}ll@{}}
\toprule
Symbol & Description \\
\midrule
$M$ & Number of variables \\
$\tau_{\max}$ & Maximum action-effect delay horizon considered \\
$I$ & Number of outer training iterations \\
$K$ & Number of samples collected per iteration \\
$|List_{do}|$ & Number of do-intervention variables \\
\bottomrule
\end{tabular}
\caption{Notation used in computational cost analysis.}
\label{tab:delay_complexity_symbols}
\end{table}

\subsection{Computational Complexity Analysis}

The REINFORCE training loop consists of two main computation phases:
\begin{equation}
\mathcal{C}_{\text{delay}} = \underbrace{\sum_{i=1}^{I} \sum_{k=1}^{K} \sum_{j=1}^{|List_{do}|} \mathcal{C}_{\text{sampling\_accum}}(M, \tau_{\max})}_{\text{Configuration Sampling
\& Accumulation}} + \underbrace{\sum_{i=1}^{I} \mathcal{C}_{\text{gradient}}(K, |List_{do}|, M, \tau_{\max})}_{\text{REINFORCE Gradient Computation}}
\end{equation}
where $\mathcal{C}_{\text{sampling\_accum}}$ represents the per-sample computation within the triple nested loop, and $\mathcal{C}_{\text{gradient}}$ represents the batch gradient
computation performed at the end of each outer iteration. The computational complexity of each is given below:


(1) Configuration Sampling \& Accumulation (within triple loop):
\begin{equation}
\begin{aligned}
\mathcal{C}_{\text{sampling\_accum}}(M, \tau_{\max})
&= \mathcal{O}(\text{Configuration Sampling}) \\
& \quad + \mathcal{O}(\text{Probability Normalization}) \\
& \quad + \mathcal{O}(\text{Gradient Accumulation})
\end{aligned}
\end{equation}

Breaking down the accumulation:
\begin{align}
\mathcal{O}(\text{Configuration Sampling}) &= \mathcal{O}(M \times \tau_{\max}) \\
\mathcal{O}(\text{Probability Normalization}) &= \mathcal{O}(M \times \tau_{\max}) \\
\mathcal{O}(\text{Gradient Accumulation}) &= \mathcal{O}(M \times \tau_{\max}) \\
\mathcal{C}_{\text{sampling\_accum}} &= \mathcal{O}(M \times \tau_{\max})
\end{align}

(2) REINFORCE Gradient Computation (per outer iteration):
\begin{equation}
\mathcal{C}_{\text{gradient}}(K, |List_{do}|, M, \tau_{\max})
= \mathcal{O}(|List_{do}| \times K \times M \times \tau_{\max})
\end{equation}

Thus, the overall computational complexity is:
\begin{equation}
\begin{aligned}
\mathcal{C}_{\text{delay}}
&= \underbrace{I \times K \times |List_{do}| \times \mathcal{O}(M \times \tau_{\max})}_{\text{Sampling \& Accumulation Phase}} \\
& \quad + \underbrace{I \times \mathcal{O}(|List_{do}| \times K \times M \times \tau_{\max})}_{\text{Gradient Computation Phase}} \\
&= \mathcal{O}(I \times K \times |List_{do}| \times M \times \tau_{\max})
\end{aligned}
\end{equation}

\subsection{Space Complexity Analysis}
The space complexity of the delay distribution modeling consists of two main components with different memory lifetimes:
\begin{equation}
\mathcal{S}_{\text{delay}} = \underbrace{\sum_{i=1}^{I} \mathcal{S}_{\text{accumulation}}(K, |List_{do}|, M, \tau_{\max})}_{\text{Gradient Accumulation Phase}} +
\underbrace{\mathcal{S}_{\text{permanent}}(M, \tau_{\max})}_{\text{Permanent Storage Phase}},
\end{equation}
where $\mathcal{S}_{\text{accumulation}}$ represents temporary memory used during gradient accumulation and $\mathcal{S}_{\text{permanent}}$ represents persistent memory for network parameters. The space complexity of each is given below:

(1) Gradient Accumulation Storage:
\begin{equation}
\mathcal{S}_{\text{accumulation}}(K, |List_{do}|, M, \tau_{\max})
= \mathcal{O}(|List_{do}| \times K \times M \times \tau_{\max})
\end{equation}
This phase stores the accumulated gradient terms and logregret values during each outer iteration, where $K \times |List_{do}|$ samples of size $M \times \tau_{\max}$ are accumulated before gradient computation.

(2) Permanent Parameter Matrix Storage:
\begin{equation}
\mathcal{S}_{\text{permanent}}(M, \tau_{\max})
= \mathcal{O}(M^2 + M \times \tau_{\max})
\end{equation}
This phase includes the delay parameter matrix $\mathbf{H}$ of size $(M, \tau_{\max})$ and the causal relationship parameter matrix of size $(M, M)$, which persist throughout training.

Thus, the overall space complexity is:
\begin{equation}
\begin{aligned}
\mathcal{S}_{\text{delay}}
&= \mathcal{O}(|List_{do}| \times K \times M \times \tau_{\max}) + \mathcal{O}(M^2 + M \times \tau_{\max}) \\
&= \mathcal{O}(M^2 + |List_{do}| \times K \times M \times \tau_{\max})
\end{aligned}
\end{equation}


\paragraph{\textbf{Summary}}
\textbf{Overall, the delay distribution modeling component exhibits low computational and space complexity, consuming minimal resources and completing in negligible time during each training iteration.}

\section{Computational Complexity Analysis of Delay-Aware Empowerment-Driven Hierarchical Policy Training}\label{app:delay_aware_policy_training_complexity}

The delay-aware empowerment estimation component computes a Monte Carlo approximation of the conditional mutual information (CMI) 
$I(S^i_{t+\tau}; o \mid \mathbf{S}_t)$ for each effect variable $S^i$ and delay horizon $\tau \in \{1, \dots, \tau_{\max}\}$, as defined in Definition~\ref{def:emp}. 
Under the practical setting where the high-level policy $\pi_i(o \mid \mathbf{S}_t)$ is used to marginalize over sub-goals $o \in \Omega_i$, and the structural causal model (SCM) provides the surrogate transition function $f^i_\theta(S^i_{t+\tau} \mid \mathbf{S}_t, o)$, the empowerment estimator proceeds by:
(i) enumerating all valid sub-goals to compute the marginal distribution $p(S^i_{t+\tau} \mid \mathbf{S}_t) = \sum_{o} \pi_i(o \mid \mathbf{S}_t) f^i_\theta(S^i_{t+\tau} \mid \mathbf{S}_t, o)$ for entropy $H(S^i_{t+\tau} \mid \mathbf{S}_t)$,
and (ii) using the executed sub-goal $o$ to evaluate the conditional entropy $H(S^i_{t+\tau} \mid \mathbf{S}_t, o)$ via $f^i_\theta$. 
We summarize all parameters appearing in the computational cost analysis in Table~\ref{tab:empowerment_complexity_symbols}.

\begin{table}[h]
\centering
\resizebox{\textwidth}{!}{
\begin{tabular}{@{}ll@{}}
\toprule
Symbol & Description \\
\midrule
$B$ & Batch size (number of full-state samples $\mathbf{S}_t$ processed) \\
$\tau_{\max}$ & Maximum action-effect delay horizon considered \\
$A$ & Cardinality of the sub-goal space $\Omega_i$ (typically 2, up to 7 for top-level policy in our setup) \\
$V$ & Cardinality of the discrete effect variable $S^i$ (typically 2 in our setup) \\
$N_\tau$ & Number of samples in the batch with delay step $\tau$, where $\sum_{\tau=1}^{\tau_{\max}} N_\tau = B$ \\
\bottomrule
\end{tabular}
}
\caption{Notation used in computational cost analysis of delay-aware empowerment.}
\label{tab:empowerment_complexity_symbols}
\end{table}

\subsection{Computational Complexity Analysis}

The empowerment computation consists of two main phases executed for each delay step $\tau$:
\begin{equation}
\mathcal{C}_{\text{emp}} = 
\underbrace{\sum_{\tau=1}^{\tau_{\max}} \mathcal{C}_{\text{marginal}}(N_\tau, A, V)}_{\text{Marginal Entropy Estimation}} +
\underbrace{\sum_{\tau=1}^{\tau_{\max}} \mathcal{C}_{\text{conditional}}(N_\tau, V)}_{\text{Conditional Entropy Estimation}}
\end{equation}
where $\mathcal{C}_{\text{marginal}}$ captures the cost of computing $H(S^i_{t+\tau} \mid \mathbf{S}_t)$ via sub-goal enumeration (Eq.~\eqref{eq:emp_1}), and $\mathcal{C}_{\text{conditional}}$ captures the cost of computing $H(S^i_{t+\tau} \mid \mathbf{S}_t, o)$ using the sampled sub-goal $o$ (Eq.~\eqref{eq:emp_2}). The computational complexity of each phase is detailed below.

(1) Marginal Entropy Estimation (per $\tau$):
\begin{equation}
\begin{aligned}
\mathcal{C}_{\text{marginal}}(N_\tau, A, V)
&= \mathcal{O}(\text{Policy Evaluation}) \\
&\quad + \mathcal{O}(\text{SCM Transition Enumeration}) \\
&\quad + \mathcal{O}(\text{Marginal Log-Probability Aggregation})
\end{aligned}
\end{equation}

Breaking down the components:
\begin{align}
\mathcal{O}(\text{Policy Evaluation}) &= \mathcal{O}(N_\tau A)\\
\mathcal{O}(\text{SCM Transition Enumeration}) &= \mathcal{O}(A \cdot N_\tau V)\\
\mathcal{O}(\text{Marginal Log-Probability Aggregation}) &= \mathcal{O}(N_\tau A V)\\
\mathcal{C}_{\text{marginal}}(N_\tau, A, V) &= \mathcal{O}(A N_\tau V)
\end{align}

(2) Conditional Entropy Estimation (per $\tau$):
\begin{equation}
\mathcal{C}_{\text{conditional}}(N_\tau, V) = \mathcal{O}(N_\tau V)
\end{equation}
This corresponds to evaluating $f^i_\theta(S^i_{t+\tau} \mid \mathbf{S}_t, o)$ for the executed sub-goal $o$ and computing the log-probability for entropy estimation (Eq.~\eqref{eq:emp_2}).

Summing over all $\tau$ and using $\sum_{\tau=1}^{\tau_{\max}} N_\tau = B$, the total computational complexity becomes:
\begin{equation}
\begin{aligned}
\mathcal{C}_{\text{emp}}
&= \sum_{\tau=1}^{\tau_{\max}} \left[ \mathcal{O}(A N_\tau V) + \mathcal{O}(N_\tau V) \right] + \underbrace{\mathcal{O}(\tau_{\max} B)}_{\text{group filtering overhead}} \\
&= \mathcal{O}(A V B) + \mathcal{O}(V B) + \mathcal{O}(\tau_{\max} B) \\
&= \mathcal{O}(A V B + \tau_{\max} B)
\end{aligned}
\end{equation}


\subsection{Space Complexity Analysis}

The space complexity arises primarily from storing intermediate SCM outputs during marginal entropy computation:

\begin{equation}
\mathcal{S}_{\text{emp}} = 
\underbrace{\max_{\tau} \mathcal{S}_{\text{temp}}(N_\tau, A, V)}_{\text{Peak Temporary Storage}} +
\underbrace{\mathcal{S}_{\text{output}}(B)}_{\text{Empowerment Estimates}}
\end{equation}

(1) Peak Temporary Storage (for a given $\tau$):
\begin{equation}
\mathcal{S}_{\text{temp}}(N_\tau, A, V) = \mathcal{O}(A N_\tau V)
\end{equation}
This is required to hold all $A$ SCM output distributions $f^i_\theta(S^i_{t+\tau} \mid \mathbf{S}_t, o)$ of shape $(N_\tau, V)$ simultaneously before computing the marginal $p(S^i_{t+\tau} \mid \mathbf{S}_t)$. In the worst case ($N_\tau = B$), this yields $\mathcal{O}(A B V)$ memory.

(2) Output Storage:
\begin{equation}
\mathcal{S}_{\text{output}}(B) = \mathcal{O}(B)
\end{equation}
for the final empowerment estimates $\hat{\mathcal{E}}(S_t^i)$ used in the REINFORCE gradient.

Thus, the overall space complexity is:
\begin{equation}
\begin{aligned}
\mathcal{S}_{\text{emp}} 
&= \mathcal{O}(A B V) + \mathcal{O}(B) \\
&= \mathcal{O}(A B V)
\end{aligned}
\end{equation}

\paragraph{\textbf{Summary}}
\textbf{Given the small sub-goal space and binary effect variables in our setup, the delay-aware empowerment estimation incurs only linear computational and memory overhead with respect to the batch size, making it highly efficient in practice.}

\section{A Comparative Analysis of Implementation, Complexity in DECHRL, D3HRL, and CDHRL}\label{app:comparison_chrl_complexity}


Considering that the implementation frameworks of HRL and DARL baseline algorithms differ significantly from that of CHRL algorithms, we focus our comparison on CHRL-based baseline algorithms that share a similar framework with the work presented in this paper. This includes a comparison of implementation details as well as an analysis of computational complexity.

\subsection{Comparison of Implementation Details}\label{app:implementation_comarison}

We begin by outlining the implementation details of the three algorithms (CDHRL, D3HRL and our DECHRL). Our analysis focuses on the two main phases of the CHRL framework, namely causal discovery and hierarchical policy training, as described in Section~\ref{sec:chrl}.




\subsubsection{CDHRL: Baseline Framework}
CDHRL serves as the foundational framework that implements single-step causal relationship learning and hierarchical policy construction. It learns one-step causal
transitions using Structural Causal Models (SCM) and builds hierarchical strategies based on the discovered causal structure. For each hierarchical level, CDHRL establishes
a unified policy that handles all subgoals at the same level, with each policy relying on single-step state transitions learned through causal intervention experiments.
The computational complexity remains relatively low since it only processes single-step transitions without additional overhead from delay modeling or spurious correlation
detection.

\subsubsection{D3HRL: Multi-Temporal Causal Discovery with Spurious Correlation Detection}
D3HRL extends CDHRL's single-step causal learning to multi-temporal causal relationship discovery by utilizing $\tau_{\max}$ parallel processes, each focusing on different temporal horizons. This extension inherently increases the computational complexity of causal discovery by approximately a factor of $\tau_{\max}$ compared to CDHRL. Furthermore, D3HRL incorporates Conditional Mutual Information (CMI) networks to detect and filter spurious correlations, representing a substantial additional computational burden. The CMI-based detection involves multiple rounds of neural network training and is relatively complex, incurring additional computational cost.
In the hierarchical policy training phase, D3HRL establishes independent policies for each subgoal and collects transition data based on the learned delay lengths, using standard HRL objectives without empowerment mechanisms.

\subsubsection{DECHRL: Delay Distribution Modeling with Empowerment-Driven Policy Training}
DECHRL builds upon both frameworks by implementing delay distribution modeling through REINFORCE-based learning of probability distributions over possible temporal delays. Unlike D3HRL's computationally expensive CMI networks, DECHRL's delay modeling employs a lightweight REINFORCE approach (please refer to \ref{app:delay_complexity} for details).
This efficiency stems from the fact that delay modeling operates on small parameter matrices and requires no complex neural network forward-backward passes.
In the hierarchical policy training phase, DECHRL also establishes independent policies for each subgoal but enhances exploration through delay-aware empowerment. This intrinsic motivation provides more efficient exploration and goal generation compared to both CDHRL's single-step approach and D3HRL's discrete multi-step framework.

\paragraph{\textbf{Summary}} 
\textbf{The implementation details of the three algorithms in these two phases are summarized in Table~\ref{tab:chrl_conn_diff}.}

\begin{table}[htbp]
\centering
\resizebox{\textwidth}{!}{
\begin{tabular}{p{0.10\textwidth} p{0.52\textwidth} p{0.48\textwidth}}
\toprule
& \textbf{\textit{Causal Discovery}} & \textbf{\textit{Hierarchical Policy Training}} \\
\midrule
CDHRL 
& Uses the current hierarchical policy to collect intervention data for learning single-step causal relationships. 
& Learns hierarchical policies using single-step state transitions, with each policy responsible for training all sub-goals at the same depth. \\
\midrule
D3HRL 
& Uses the current hierarchical policy to collect intervention data for learning multi-timescale causal relationships, and employs spurious correlation detection to identify transition-specific/causal relationship-specific fixed delays.
& Uses the learned transition-specific delays to collect transition data for hierarchical policy training, with a separate policy for each sub-goal. \\
\midrule
DECHRL 
& Uses the current hierarchical policy to collect intervention data for learning multi-timescale causal relationships, and models delay distributions corresponding to each state transition/causal relationship.
& Uses the learned transition-specific delay distributions to collect transition data for hierarchical policy training, with a separate policy for each sub-goal. \\
\bottomrule
\end{tabular}
}
\caption{A Comparative Analysis of CDHRL, D3HRL, and DECHRL Implementations.}
\label{tab:chrl_conn_diff}
\end{table}

\subsection{Computational Complexity Analysis}
Based on the detailed implementation analysis of each algorithm in Section~\ref{app:implementation_comarison}, the computational complexity of the three methods across the two training stages can be summarized as follows:
\begin{itemize}
    \item \textbf{Causal discovery stage.} CDHRL learns only single-step causal relationships. In contrast, both D3HRL and DECHRL adopt a multi-timescale approach to learn causal dependencies across delays from 1 to $\tau_{\max}$ steps. Additionally, D3HRL incorporates an extra module for spurious correlation detection, while DECHRL introduces a dedicated delay distribution modeling component.
    \item \textbf{Hierarchical policy training stage.} As illustrated in Figure~\ref{fig:task_state_transitions}(a), the \textit{GetSilverOre} task involves a causal chain of depth 4 and 5 intermediate subgoals. Building on our analysis in the previous section, CDHRL employs a number of hierarchical policies equal to the causal chain depth (4), with each policy jointly trained to handle all subgoals at its corresponding level. In contrast, both D3HRL and DECHRL allocate one policy per intermediate subgoal, resulting in a total of 5 hierarchical policies, each trained independently for a specific subgoal. Furthermore, the hierarchical policies in DECHRL are trained with a delay-aware empowerment objective.
\end{itemize}


To enable a clear and consistent comparison, we take CDHRL, the foundational framework shared by all three methods, as the reference. Specifically, we express the computational complexity of D3HRL and DECHRL at each stage as a multiple of CDHRL’s baseline cost, plus the overhead incurred by their respective additional modules (e.g., spurious correlation detection in D3HRL and delay distribution modeling in DECHRL). These normalized complexity estimates, derived from the analysis above, are consolidated in Table~\ref{tab:complexity_analysis} below.

\begin{table}[htbp]
\centering
\resizebox{\textwidth}{!}{
\begin{tabular}{ccc}
\toprule
 & \textbf{\textit{Causal Discovery}} & \textbf{\textit{Hierarchical Policy Training}} \\
 \midrule
CDHRL 
& $\mathcal{O}(\text{Single Step SCM Training})$ 
& $\text{4 level} \times \mathcal{O(\text{Policy Training})}$\\
\midrule
D3HRL 
& 
$\begin{aligned}
&\tau_{\max} \times \mathcal{O}(\text{Single Step SCM Training}) \\
&+ \mathcal{O}(\text{Spurious Correlation Detection})
\end{aligned}$
& $\text{5 level} \times \mathcal{O(\text{Policy Training})}$\\
\midrule
DECHRL 
& 
$\begin{aligned}
&\tau_{\max} \times \mathcal{O}(\text{Single Step SCM Training}) \\
&+ \mathcal{O}(\text{Delay Distribution Modeling})
\end{aligned}$ 
& $\text{5 level} \times \mathcal{O(\text{Delay-Aware Policy Training})}$\\
\bottomrule
\end{tabular}
}
\caption{Computational Complexity Analysis of three CHRL algorithms on the \textit{GetSilverore} task.}
\label{tab:complexity_analysis}
\end{table}



It is evident that CDHRL achieves the lowest computational complexity, as it is limited to modeling single-step, fixed-delay state transitions and causal relationships. The key comparison, however, lies between the two delay-aware frameworks, D3HRL and DECHRL, whose computational trade-offs warrant closer examination. 
As shown in Table~\ref{tab:complexity_analysis}, each method introduces distinct algorithmic components. Specifically, DECHRL employs delay distribution modeling and delay-aware empowerment computation, both of which are relatively lightweight (see detailed analyses in~\ref{app:delay_complexity} and~\ref{app:delay_aware_policy_training_complexity}). D3HRL relies on spurious correlation detection via conditional mutual information (CMI) estimation, which involves iterative neural network training and consequently incurs higher computational overhead—particularly during the causal discovery phase. Based on this distinction, we can reasonably infer that: \textit{(i) the causal discovery phase in D3HRL is more computationally demanding than in DECHRL, whereas (ii) the hierarchical policy training phase in DECHRL is slightly more expensive due to the additional empowerment-based reward calculation}. Overall, their total computational costs are comparable, satisfying:
\begin{equation}
\mathcal{T}_{\text{CDHRL}} < \mathcal{T}_{\text{D3HRL}} \approx \mathcal{T}_{\text{DECHRL}}.
\end{equation}
While the computational difference between D3HRL and DECHRL is marginal, DECHRL delivers significant improvements in temporal reasoning and exploration efficiency.
\textbf{Therefore, DECHRL strikes an optimal balance between computational efficiency and advanced temporal modeling capabilities, making it the preferred choice for environments with complex, multi-step temporal dynamics.}

\section*{Declaration of Generative AI and AI-assisted technologies in the writing process}
During the preparation of this work, the authors used Qwen~\cite{DBLP:journals/corr/abs-2505-09388} to check grammatical errors and improve the clarity of the text. After using them, the authors reviewed and edited the content as needed and take full responsibility for the content of the publication.

\section*{Acknowledgments}
This work was supported by the National Natural Science Foundation of China (No. 62372459).

\bibliographystyle{elsarticle-num} 
\bibliography{nn_emp}



\end{document}